\documentclass[letterpaper]{article} 
\usepackage{aaai23}  
\usepackage{times}  
\usepackage{helvet}  
\usepackage{courier}  
\usepackage[hyphens]{url}  
\usepackage{graphicx} 
\usepackage{natbib}  
\usepackage{caption} 
\usepackage{algorithm}
\usepackage{algorithmic}

\usepackage{newfloat}
\usepackage{listings}
\DeclareCaptionStyle{ruled}{labelfont=normalfont,labelsep=colon,strut=off} 
\floatstyle{ruled}
\newfloat{listing}{tb}{lst}{}
\floatname{listing}{Listing}

\nocopyright

\usepackage{amssymb}
\usepackage{amsmath}
\usepackage{amsthm}
\usepackage{bm}
\usepackage{caption}
\usepackage{subcaption}
\newtheorem{proposition}{Proposition}
\title{Goal-Conditioned Q-Learning as Knowledge Distillation }
\author{
Alexander Levine\textsuperscript{\rm 1}\\
Soheil Feizi\textsuperscript{\rm 1}
}
\affiliations{
    \textsuperscript{\rm 1}University of Maryland\\
    College Park, Maryland, USA\\
    alevine0@cs.umd.edu
}

\begin{document}

\maketitle

\begin{abstract}
Many applications of reinforcement learning can be formalized as goal-conditioned environments, where, in each episode, there is a ``goal'' that affects the rewards obtained during that episode but does not affect the dynamics. Various techniques have been proposed to improve performance in goal-conditioned environments, such as automatic curriculum generation and goal relabeling. In this work, we explore a connection between off-policy reinforcement learning in goal-conditioned settings and knowledge distillation. In particular: the current Q-value function and the target Q-value estimate are both functions of the goal, and we would like to train the Q-value function to match its target \textit{for all goals}. We therefore apply Gradient-Based Attention Transfer \cite{zagoruyko2017paying}, a knowledge distillation technique, to the Q-function update. We empirically show that this can improve the performance of goal-conditioned off-policy reinforcement learning when the space of goals is high-dimensional. We also show that this technique can be adapted to allow for efficient learning in the case of \textit{multiple simultaneous sparse goals}, where the agent can attain a reward by achieving any one of a large set of objectives, all specified at test time. Finally, to provide theoretical support, we give examples of classes of environments where (under some assumptions) standard off-policy algorithms such as DDPG require at least $O(d^2)$ replay buffer transitions to learn an optimal policy, while our proposed technique requires only $O(d)$ transitions, where $d$ is the dimensionality of the goal and state space. Code is available at \url{https://github.com/alevine0/ReenGAGE}.
\end{abstract}

\section{Introduction}
In recent years, many works have focused on applying deep reinforcement learning to goal-conditioned tasks, through approaches such as goal relabeling \cite{andrychowicz2017hindsight, nair2018visual,yang2021mher,fang2019curriculum} and automatic curriculum generation \cite{florensa2018automatic,sukhbaatar2018intrinsic,zhang2020automatic}. In this work, we focus on model-free off-policy goal-conditioned RL, and present a novel technique for improving performance in this setting. Our approach relies on a connection between the standard Bellman update used in off-policy reinforcement learning in a goal-conditioned setting, and \textit{knowledge distillation}, the task of training a student network to model the same function as a (generally more complex) teacher network.\footnote{Some works use the term ``knowledge distillation'' to refer to the particular method for this task proposed by \cite{Hinton2015DistillingTK}, while others, such as \cite{gou2021knowledge} use it to refer to the task in general; we use the latter definition. } In brief, the Bellman update can be viewed as an instance of (conditional, stochastic) knowledge distillation, where the current Q-value estimate is the student, the target Q-value network (averaged over transitions) is the teacher, and the independent variable is the \textit{goal} that the agent is attempting to reach. We use this insight to develop an novel off-policy algorithm that in some instances has improved performance over baselines for goal-conditioned tasks. Our main contributions are as follows:
\begin{enumerate}
    \item We propose \textbf{ReenGAGE}, a novel technique for goal-conditioned off-policy reinforcement learning, and evaluate its performance.
    \item We propose \textbf{Multi-ReenGAGE}, a variant of ReenGAGE  well-suited for goal-conditioned environments with \textit{many simultaneous sparse goals.}
    \item We provide theoretical justification for ReenGAGE by showing that it is in some cases asymptotically more efficient, in terms of the total number of replay buffer transitions required to learn an optimal policy, than standard off-policy algorithms such as DDPG. 
\end{enumerate}
In most of this work, we focus on continuous action control problems; we extend our method to discrete action spaces in the appendix. Note that while we mostly focus on using ReenGAGE on top of HER \cite{andrychowicz2017hindsight} and DDPG \cite{DBLP:journals/corr/LillicrapHPHETS15} in this work, it can be easily applied alongside any goal-relabeling scheme or automated curriculum, and can be adapted for other off-policy algorithms such as SAC \cite{haarnoja2018soft} or TD3 \cite{fujimoto2018addressing}. In particular, we include an application to SAC in the appendix.

\section{Preliminaries and Notation}
We consider control problems defined by goal-conditioned MDPs $(S,A,G,\mathcal{T},R)$, where $S$, $A$, and $G$ denote sets of states, actions, and goals, respectively, $G$ and $A$ are assumed to be continuous spaces, $\mathcal{T} \in S\times A \rightarrow \mathcal{P}(S)$ is a stochastic transition function, and $R \in S \times G \rightarrow \mathbb{R}$ is a reward function. At every step, starting at state $s \in S$, an agent chooses an action $a \in A$. The system then transitions to $s' \sim \mathcal{T}(s,a)$, and the agent receives the reward $R(s',g)$.

For now, we assume that the reward function $R(s',g)$ is \textit{known a priori} to the learning algorithm (while the transition function is not): this just means that we know how to interpret the objective which we are trying to achieve. Note that existing goal relabeling techniques, such as HER \cite{andrychowicz2017hindsight}, implicitly make this assumption, as it is necessary to compute the rewards of relabeled transitions. We will discuss cases where this assumption can be relaxed in later sections.

We consider both \textit{shaped} rewards, in which $R(s',g)$ is continuous and differentiable everywhere, as well as \textit{sparse} rewards, where $R(s',g)$ is not assumed to be differentiable, but maintains some constant value $c_\text{low}$ (e.g., $c_\text{low} =$ -1 or 0) with $\nabla_g R(s',g) = \bm{0}$ for a substantial fraction of inputs. (There may be some boundary points where $R(s',g) = c_\text{low}$ but $\nabla_g R(s',g)$ is not defined or $\nabla_g R(s',g) \neq \bm{0}$, but in theoretical discussion we will assume these are of measure zero).

The objective of goal-conditioned RL is to find a policy $\pi\in S \times G \rightarrow A$ such that the discounted future reward:
\begin{equation}
    r = \sum^\infty_{t=0} \gamma^t R(s_{t+1},g)
\end{equation}
is maximized in expectation. One common approach is to find the policy $\pi$ and Q-function $Q \in S\times A\times G \rightarrow \mathbb{R}$ that solve the Bellman equation for Q-learning \cite{watkins1992q}, conditioned on a goal $g$:
\begin{equation} \label{eq:standard_bellman}
\begin{split}
            &\forall s,a,g,\\  &Q(s,a,g) = \mathop \mathbb{E}_{s' \sim \mathcal{T}(s,a)} [R(s',g) + \gamma Q (s',\pi(s', g), g) ]  
\end{split}
\end{equation}
\begin{equation}
\forall s,g, \,\,\,
           \pi(s, g) = \arg\max_{a}  Q(s,a,g). 
\end{equation}
If functions $\pi$ and $Q$ satisfy these, then $\pi$ is guaranteed to be an optimal policy. In practice, off-policy RL techniques, notably DDPG \cite{DBLP:journals/corr/LillicrapHPHETS15} can be used to solve for these functions iteratively by drawing tuples $(s,a,s',g)$ from a replay buffer:

\begin{equation}
\begin{split}
       \mathcal{L}_{\text{critic}} =  \mathop\mathbb{E}_{(s,a,s',g)\sim \text{Buffer}} \Bigg[& \mathcal{L_{\text{mse}}} \Big[Q_\theta(s,a,g), \\
        R(s',g)&+ \gamma Q_{\theta'} (s',\pi_{\phi'}(s', g), g)\Big]\Bigg] \label{eq:standard_critic_loss}
\end{split}
\end{equation}
\begin{equation}
    \mathcal{L}_{\text{actor}} = \mathop\mathbb{E}_{(s,g)\sim \text{Buffer}}  - Q_{\theta} (s,\pi_{\phi}(s,g),g),
\end{equation}
where $\theta$ and $\phi$ are the \textit{current} critic and actor parameters, and $\theta'$ and $\phi'$ are \textit{target} parameters, which are periodically updated to more closely match the current estimates.
Note that Equation \ref{eq:standard_bellman} should ideally hold for \textit{all} $(s,a,g)$: therefore the distribution of $(s,a,g)$ in the replay buffer does not need to precisely follow any particular distribution, assuming sufficient visitation of possible tuples.\footnote{In practice, replay buffers which better match the behavioral distribution result in better training, due to sources of ``extrapolation error'', including incomplete visitation and model inductive bias; see \cite{fujimoto2019off}. } The only necessary constraint on the buffer distribution is that the marginal distribution of $s'$ matches the transition function:
\begin{equation}
   \forall s, a,g,\,\,\, \Pr_{\text{Buffer}} [s'|s,a,g] \approx \Pr_{\mathcal{T}} [s'|s,a] \label{eq:relabeling}, 
\end{equation}
so that the relation in Equation 2 is respected. This means that the goal which is included in the buffer need not necessarily reflect a ``true'' historical experience of the agent during training, but can instead be relabeled to enhance training. \cite{andrychowicz2017hindsight,nair2018visual,yang2021mher,fang2019curriculum}.
Interestingly, \cite{schroecker2021universal} shows that HER \cite{andrychowicz2017hindsight}, a popular relabeling technique, actually \textit{does not} respect Equation \ref{eq:relabeling} when the transition function is nondeterministic, and therefore may exhibit ``hindsight bias.''
\section{Proposed Method}
From Equation \ref{eq:standard_bellman}, we can take the gradient with respect to $g$:

\begin{equation}
\begin{split}
           &\nabla_g Q(s,a,g) = \\
           &\nabla_g \mathop \mathbb{E}_{s' \sim \mathcal{T}(s,a)}   [R(s',g) + \gamma Q (s',\pi(s', g), g) ] = \\ 
           &  \mathop \mathbb{E}_{s' \sim \mathcal{T}(s,a)}  [\nabla_g R(s',g) + \gamma \nabla_g Q (s',\pi(s', g), g) ]. \\
           \end{split} \label{eq:grad_trick}
\end{equation}
Because the gradient of the Q-value function is equal to the \textit{expectation of the gradient} of the sum of the reward and the next-step Q-value, this suggests that we can augment the standard DDPG critic loss with a gradient term, which estimates this expected gradient using the replay buffer samples:
\begin{equation}
\begin{split}
        \mathcal{L_\textbf{ReenGAGE}} =  \mathop\mathbb{E}_{(s,a,s',g)\sim \text{Buffer}} \Bigg[& \mathcal{L_{\text{mse}}} \Big[Q_\theta(s,a,g), \\
        R(s',g)+ \gamma Q_{\theta'}& (s',\pi_{\phi'}(s', g), g)\Big] \\
        + \alpha  \mathcal{L_{\text{mse}}} \Big[\nabla_g Q_\theta(s,a,g)&, \\
         \nabla_g R(s',g)+ \gamma \nabla_g &Q_{\theta'}  (s',\pi_{\phi'}(s', g), g)\Big]\Bigg]  \label{eq:method_dense}
\end{split}
\end{equation}

where $\alpha$ is a constant hyperparameter. Note that the second MSE term is applied to a vector: thus we are fitting $\nabla_g Q_\theta(s_0,a_0,g)$ in all $\text{dim}(g)$ dimensions. This allows more information to flow from the target function to the current Q-function (a $\text{dim}(g)-$vector instead of a scalar), and may therefore improve training. We call our method \textbf{Re}inforcem\textbf{en}t learning with \textbf{G}radient \textbf{A}ttention for \textbf{G}oal-seeking \textbf{E}fficiently, or \textbf{ReenGAGE}.

In the case of shaped rewards, we can use this loss function directly. In the case of sparse rewards, $\nabla_g R(s',g)$ is not necessarily defined or available. However, it is also zero for a substantial fraction of inputs, and, if $R(s',g) =c_\text{low}$, then $\nabla_g R(s',g) = \bm{0}$ with high probability. Therefore, we use the gradient loss term only when training on tuples where $R(s',g) =c_\text{low}$, and assume $\nabla_g R(s',g) = \bm{0}$:
\begin{equation}
\begin{split}
        \mathcal{L^{\text{(sparse)}}_\text{\textbf{ReenGAGE}}} =  \mathop\mathbb{E}_{(s,a,s')} &\Bigg[ \mathcal{L_{\text{mse}}} \Big[Q_\theta(s,a,g), \\
        R(s',g)&+ \gamma Q_{\theta'} (s',\pi_{\phi'}(s', g), g)\Big] \\
        +  \bm{1}_{R(s',g) =c_\text{low}}  \alpha  &\mathcal{L_{\text{mse}}} \Big[\nabla_g Q_\theta(s,a,g), \\
    & \gamma \nabla_g Q_{\theta'}  (s',\pi_{\phi'}(s', g), g)\Big]\Bigg]. \label{eq:sparse_method}
\end{split}
\end{equation}
In this sparse case, if ReenGAGE is used alone (i.e., without goal relabeling), then the reward function $R$ does not need to be known explicitly \textit{a priori}. Instead, the observed values of the rewards $R(s',g)$ from the training rollouts may be used.  

Note that ReenGAGE can only be used in goal-conditioned reinforcement learning problems: in particular, we cannot use gradients with respect to states or actions in a way similar to Equation \ref{eq:grad_trick}, because, unlike in  Equation \ref{eq:grad_trick}: 
\begin{equation}
          \nabla_{s,a} \mathop \mathbb{E}_{s' \sim \mathcal{T}(s,a)}   [ (\cdot) ] \neq 
             \mathop \mathbb{E}_{s' \sim \mathcal{T}(s,a)}  [\nabla_{s,a} (\cdot) ]
\end{equation}
because the sampling distribution depends on $s$ and $a$.
\subsection{Connection to Knowledge Distillation}
\begin{figure}
    \centering
    \includegraphics[width=0.48\textwidth]{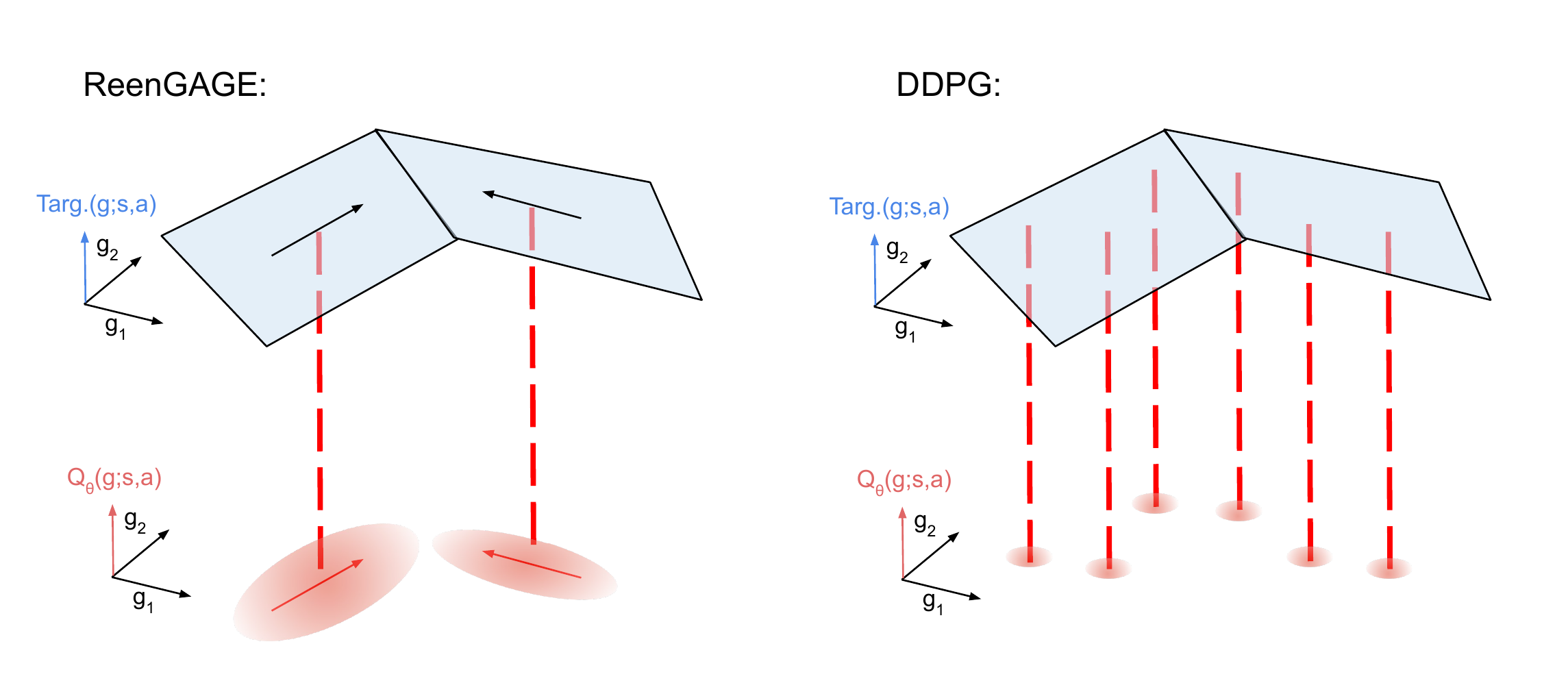}
    \caption{Illustration comparing the information flow from the Q-value target function to the current Q-function in ReenGAGE, compared to standard DDPG. In ReenGAGE, for each $(s,a,s',g)$ tuple, the gradient with respect to the goal is used as supervision, while in standard DDPG, only point values are used. Note that in stochastic environments, each tuple only provides a stochastic estimate of the target gradient (in ReenGAGE) or target point value (in DDPG).}
    \label{fig:reengage_diagram}
\end{figure}

We can view Equation \ref{eq:standard_critic_loss} as the loss function of a regression problem, fitting $Q_\theta$ to the target. We treat $g$ as the independent variable, and $s$ and $a$ as parameters:
\begin{equation}
\begin{split}
            &\forall g, \,\,\,\, Q_\theta(g; s,a) := \text{Targ.}(g;s,a) \text{   where}\\
         \text{Targ.}&(g;s,a) =\hspace{-12pt}\mathop\mathbb{E}_{s'\sim \mathcal{T} (s,a)} \hspace{-12pt} R(s',g)+ \gamma Q_{\theta'} (s',\pi_{\phi'}(s', g), g) 
\end{split}
\end{equation}
This can be framed as a knowledge distillation problem: we can view $\text{Targ.}(g;s,a)$ as a (difficult to compute) \textit{teacher} function, and we are trying to fit the network $Q_\theta(g; s,a)$ to represent the same function of $g$. Note however that conventional ``knowledge distillation'' \cite{Hinton2015DistillingTK}, which matches the logits of one network to another for classification problems in order to provide richer supervision than simply matching the class label output, cannot be applied here because the output is scalar. However, \cite{zagoruyko2017paying} proposes \textit{Gradient-based Attention Transfer} which instead matches the \textit{gradients} of the student to the teacher using a regularization term. Applied to our Q function and target, this is:
\begin{equation}
    \begin{split}
        \mathcal{L_{\text{GAT}}} =&  \mathcal{L_{\text{MSE}}} (Q_\theta(g; s,a), \text{Targ.}(g;s,a))\\ +&\alpha \|\nabla_g Q_\theta(g; s,a)- \nabla_g  \text{Targ.}(g;s,a)\|_2^2,
    \end{split}
\end{equation}
which is in fact the ReenGAGE loss function. Therefore we can think of ReenGAGE as applying knowledge distillation (specifically Gradient-based Attention Transfer) to the Q-value update.\footnote{ \cite{zagoruyko2017paying} actually uses the $\ell_2$ distance between the gradients as the regularization term, rather than its square. However, because our gradient estimate is stochastic (in particular, we are using samples of $s'\sim \mathcal{T}(s,a)$ rather than the expectation), we instead use the mean squared error, so that the current $Q$ gradients will converge to the population mean. \cite{zagoruyko2017paying} notes that their particular choice of the $\ell_2$ norm is arbitrary: other metrics should work. } See Figure \ref{fig:reengage_diagram} for an illustration. 
While the computation of the loss function gradient is somewhat more complex here than in standard training, involving mixed partial derivatives, \cite{zagoruyko2017paying} notes that it can still be performed efficiently using modern automatic differentiation packages; in fact, this ``double backpropagation'' should only scale the computation time by a constant factor \cite{Etmann2019ACL}. Further discussion of this and empirical runtime comparisons are provided in the appendix.

\cite{zagoruyko2017paying} also propose Activation-based Attention Transfer, which transfers intermediate layer activations from teacher to student network rather than gradients; in fact, they report better performance using this method than the gradient method. However, this is not applicable in our case. Firstly, in the dense reward case, we cannot model the reward function in this way. Secondly, unlike the gradient operator, activations are nonlinear: so, even in the sparse case, we cannot assume that the activations of a ``converged'' Q-network perfectly modeling the expected target will be the equal to the expected activations of the target network (i.e., there is no activation equivalent to Equation \ref{eq:grad_trick}.) See \cite{hsu2022closer} for a review of sources of auxiliary network information that can be used for knowledge distillation.

\begin{figure*}[t]
     \centering
     \begin{subfigure}[b]{0.33\textwidth}
         \centering
         \includegraphics[width=\textwidth]{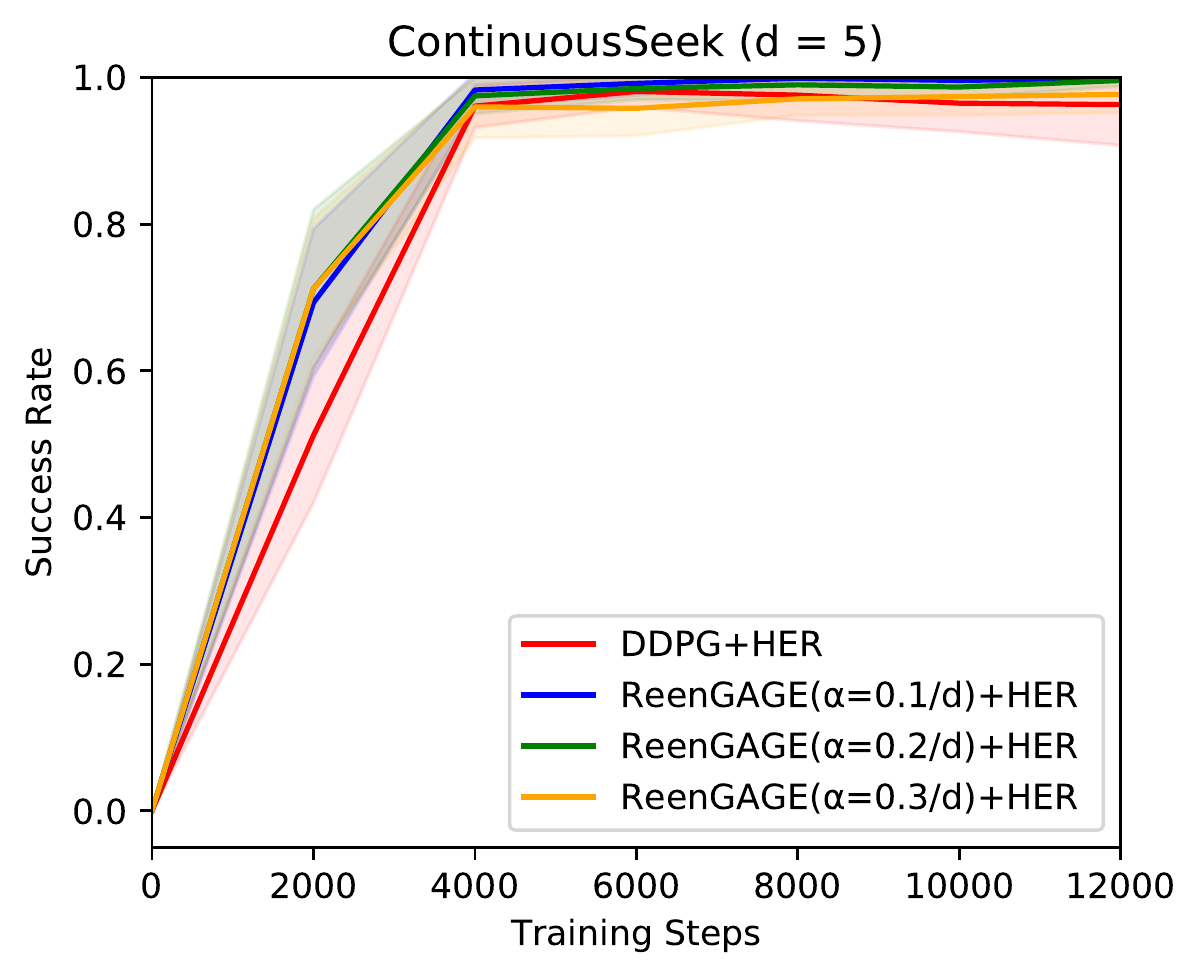}
     \end{subfigure}
     \hfill
     \begin{subfigure}[b]{0.33\textwidth}
         \centering
         \includegraphics[width=\textwidth]{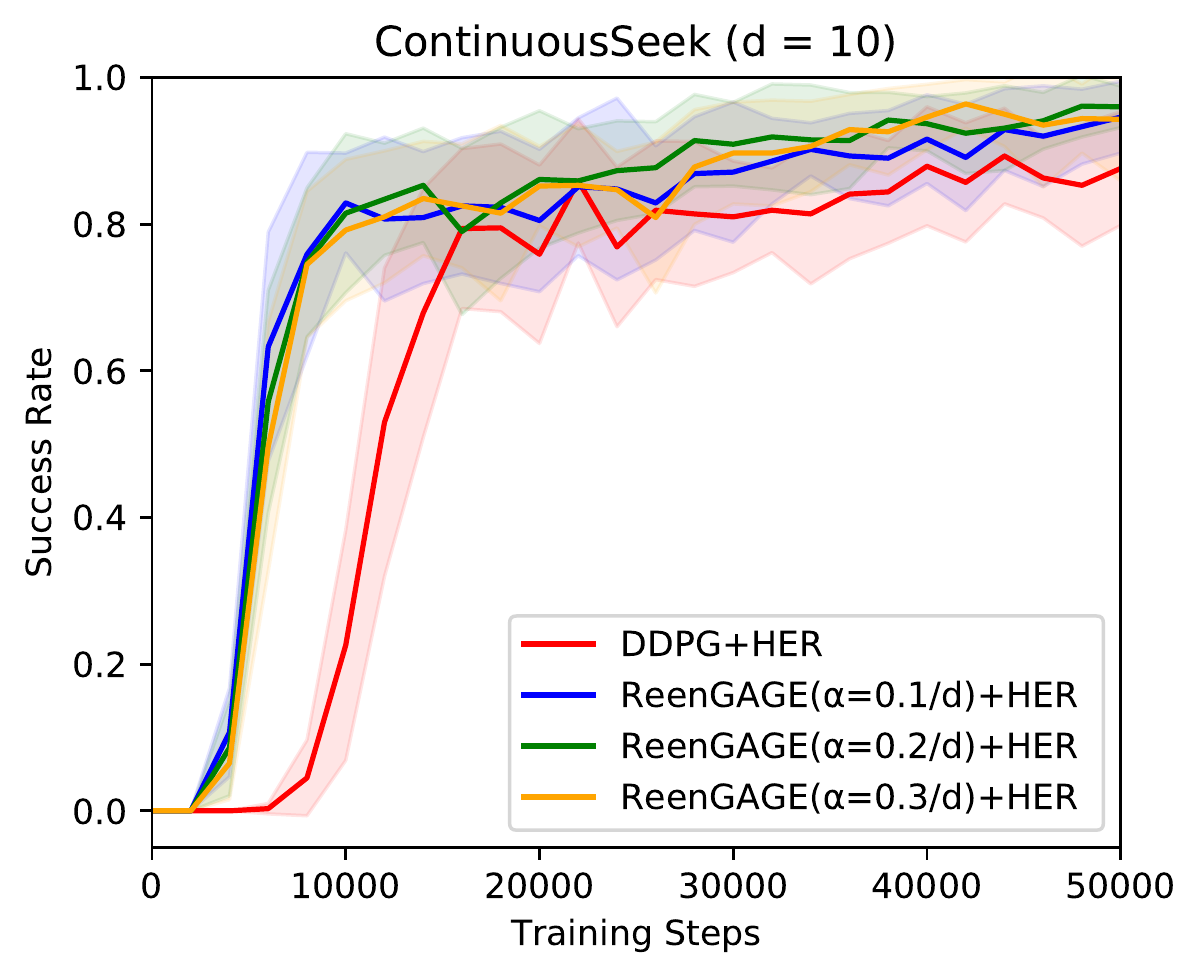}
     \end{subfigure}
     \hfill
     \begin{subfigure}[b]{0.33\textwidth}
         \centering
         \includegraphics[width=\textwidth]{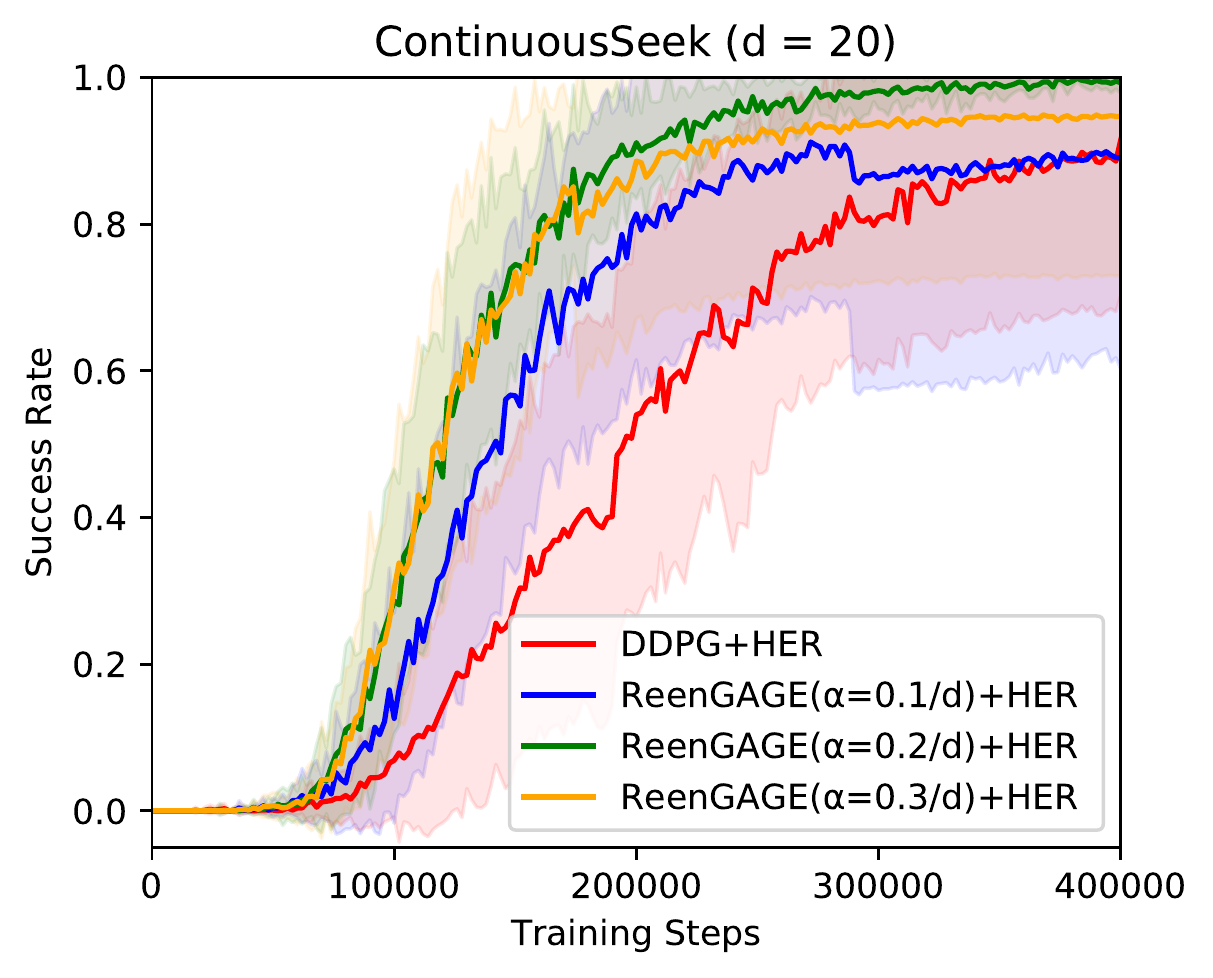}
     \end{subfigure}
        \caption{ContinuousSeek results. Lines show the mean and standard deviation over 20 random seeds (kept the same for all experiments.) The Y-axis represents the success rate, defined as the fraction of test episodes for which the goal is ever reached.}
        \label{fig:continuous_seek_results}
\end{figure*}
\section{Toy Example Experiments}
We first apply ReenGAGE to a simple sparse-reward environment, which we call \textbf{ContinuousSeek}. This task is a continuous variant of the discrete ``Bit-Flipping'' environment proposed in \cite{andrychowicz2017hindsight}. In our proposed task, the objective is to navigate from an initial state in d-dimensional space to a desired goal state, by, at each step, adding an $\ell_\infty$-bounded vector to the current state. Formally:
\begin{itemize}
    \item $s,g \in [-D,D]^d$
    \item $a \in [-1,1]^d$
    \item $\mathcal{T}(s,a) = s+a$ (clipped into $[-D,D]^d$)
    \item $R(s,g) = -1 + \bm{1}_{\|s-g\|_\infty \leq \epsilon}$
    \item initial state $s_0 = \bm{0}$.
\end{itemize}

In our experiments, we use $D=5$, $\epsilon=0.1$, and we run for 10 steps per episode. We test with $d =5,10$ and $20$. The chance that a random state achieves the goal is approximately $\frac{1}{50^d}$, so this is an extremely sparse reward problem (as sparse as ``Bit-Flipping'' with $5.6\times d$ bits). As a baseline, we use DDPG with HER.

See Figure \ref{fig:continuous_seek_results} for results. For the baseline and each value of $\alpha$, we performed a grid search over learning rates \{0.00025,\,0.0005,\,0.001,\,0.0015\} and batch sizes \{128,\,256,\,512\}; the curves shown represent the ``best'' hyperparameter settings for each $\alpha$, defined as maximizing the area under the curves. See appendix for results for all hyperparameter settings. We studied the learning rate specifically to ensure that the ReenGAGE regularization term is not simply ``scaling up'' the loss function with similar gradient updates. Other hyperparameters were kept fixed and are listed in the appendix.
 We see that ReenGAGE clearly improves over the baseline for larger-dimensionality goals ($d=10$ and $d=20$): this shows that ReenGAGE can improve the performance of DDPG in such high-dimensional goal settings. See the appendix for a similar experiment with SAC as the base off-policy learning algorithm instead of DDPG.
 \section{Robotics Experiments}
 \begin{figure*}
     \centering
     \begin{subfigure}[b]{0.33\textwidth}
         \centering
         \includegraphics[width=0.9\textwidth]{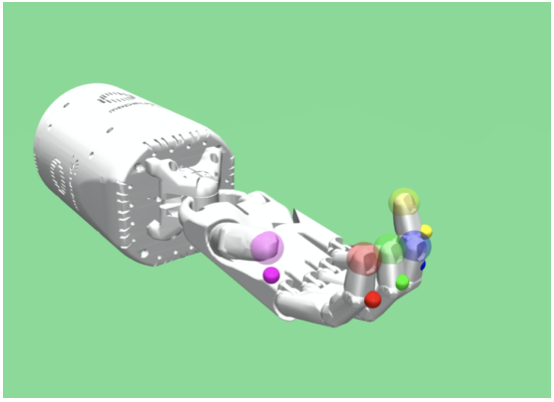}\vspace{0.6cm}
         \caption{} 
     \end{subfigure}
     \hfill
     \begin{subfigure}[b]{0.33\textwidth}
         \centering
         \includegraphics[width=\textwidth]{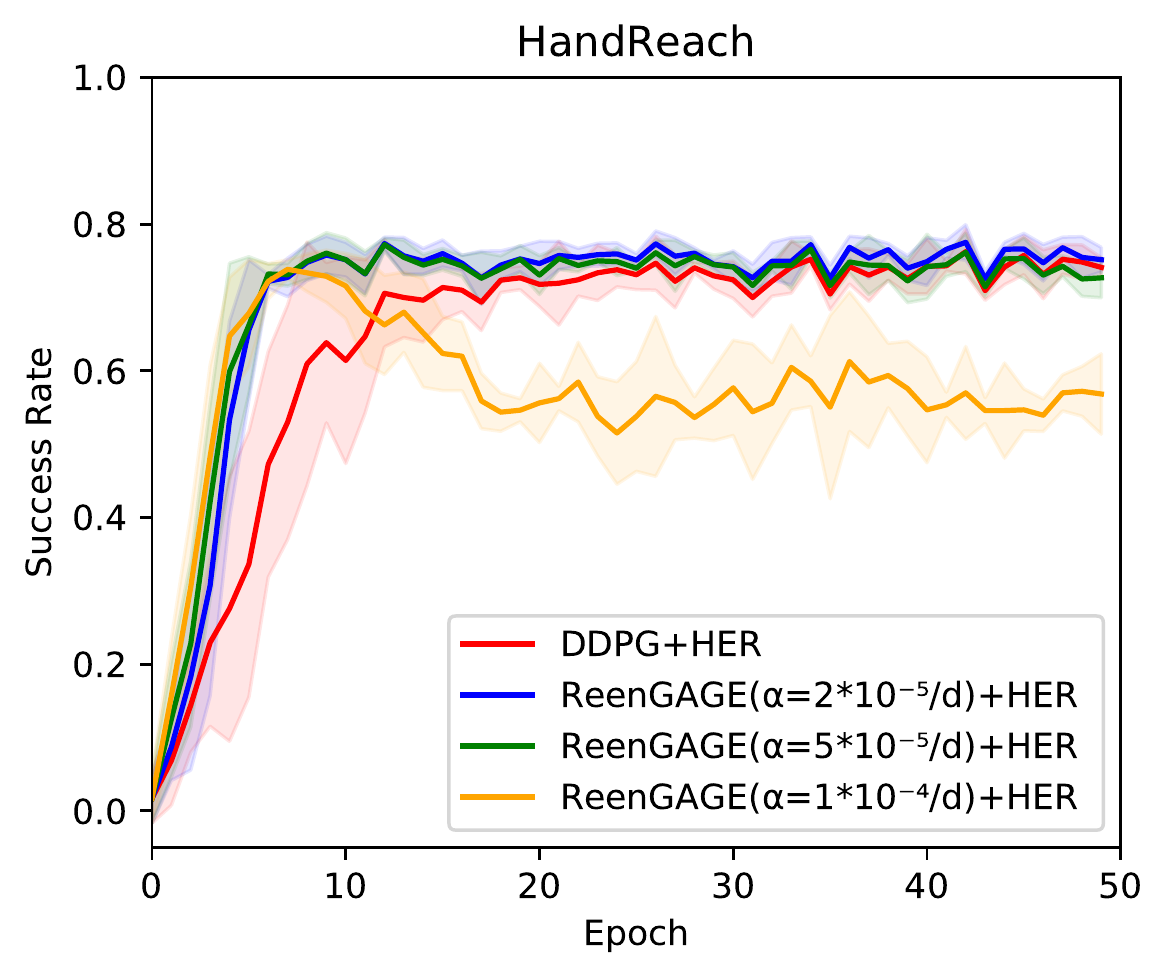}
      \caption{}
     \end{subfigure}
     \hfill
     \begin{subfigure}[b]{0.33\textwidth}
         \centering
         \includegraphics[width=\textwidth]{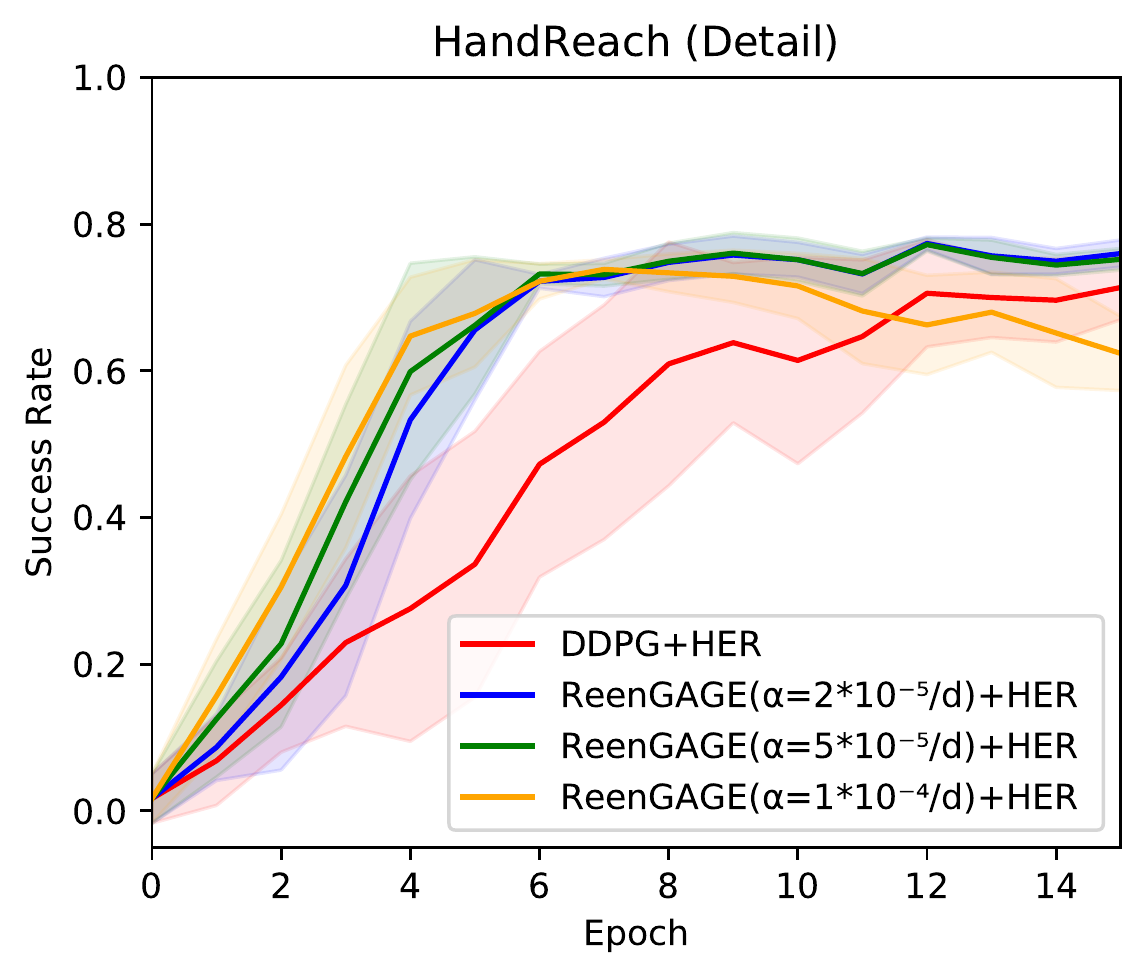}
         \caption{}
     \end{subfigure}
        \caption{HandReach results. (a) Rendering of the HandReach simulation environment. Figure taken from \cite{plappert2018}. (b) Performance of ReenGAGE on HandReach, compared to the baseline from \cite{plappert2018}. Lines represent mean and standard deviation for the same set of 5 seeds. The X-axis is the number of training epochs, as defined in \cite{plappert2018}, while the Y-axis is the success rate, defined by \cite{plappert2018} as the fraction of test episodes where the \textit{final} state satisfies the goal. (c) Detailed view of (b), showing the epochs before convergence, where the advantage of ReenGAGE is most clear.}
        \label{fig:hand_reach_results}
\end{figure*}
We tested our method on \textbf{HandReach}, the environment from the OpenAI Gym Robotics suite \cite{plappert2018} with the highest-dimensional goal space ($d=15$). In this sparse-reward environment, the agent controls a simulated robotic hand with 20-dimensional actions controlling the hand's joints; the goal is to move all of the fingertips to the specified 3-dimensional positions. As a baseline, we use the released DDPG+HER code from \cite{plappert2018}, with all hyperparameters as originally presented, and only modify the critic loss term. Results are presented in Figure \ref{fig:hand_reach_results}. In this environment, we see that ReenGAGE greatly speeds up convergence compared to the baseline. However, at a high value of $\alpha$, the success rate declines after first converging. This shows that ReenGAGE may cause some instability if the gradient loss term is too large, and that tuning the coefficient $\alpha$ is necessary (see also the Limitations section below).

We also tried our method on the \textbf{HandManipulateBlock} environment from the same paper; however, in this lower-dimensional goal environment ($d=7$) ReenGAGE was not shown to improve performance. This is compatible with our observation from the ContinuousSeek environment that ReenGAGE leads to greater improvements for higher-dimensional tasks, as the dimensionality of the additional goal-gradient information that ReenGAGE propagates increases. Results are provided in the appendix.
\section{Multi-ReenGAGE: ReenGAGE for Multiple Simultaneous Goals}
\begin{figure*}[ht!]
     \centering
     \begin{subfigure}[b]{0.20\textwidth}
         \centering
         \includegraphics[width=\textwidth]{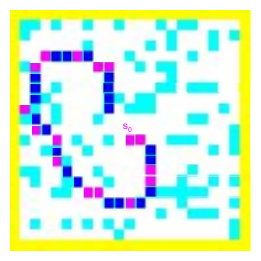}
         \caption{}
         \vspace{0.5cm}
     \end{subfigure}
     \hfill
     \begin{subfigure}[b]{0.29\textwidth}
         \centering
         \includegraphics[width=\textwidth]{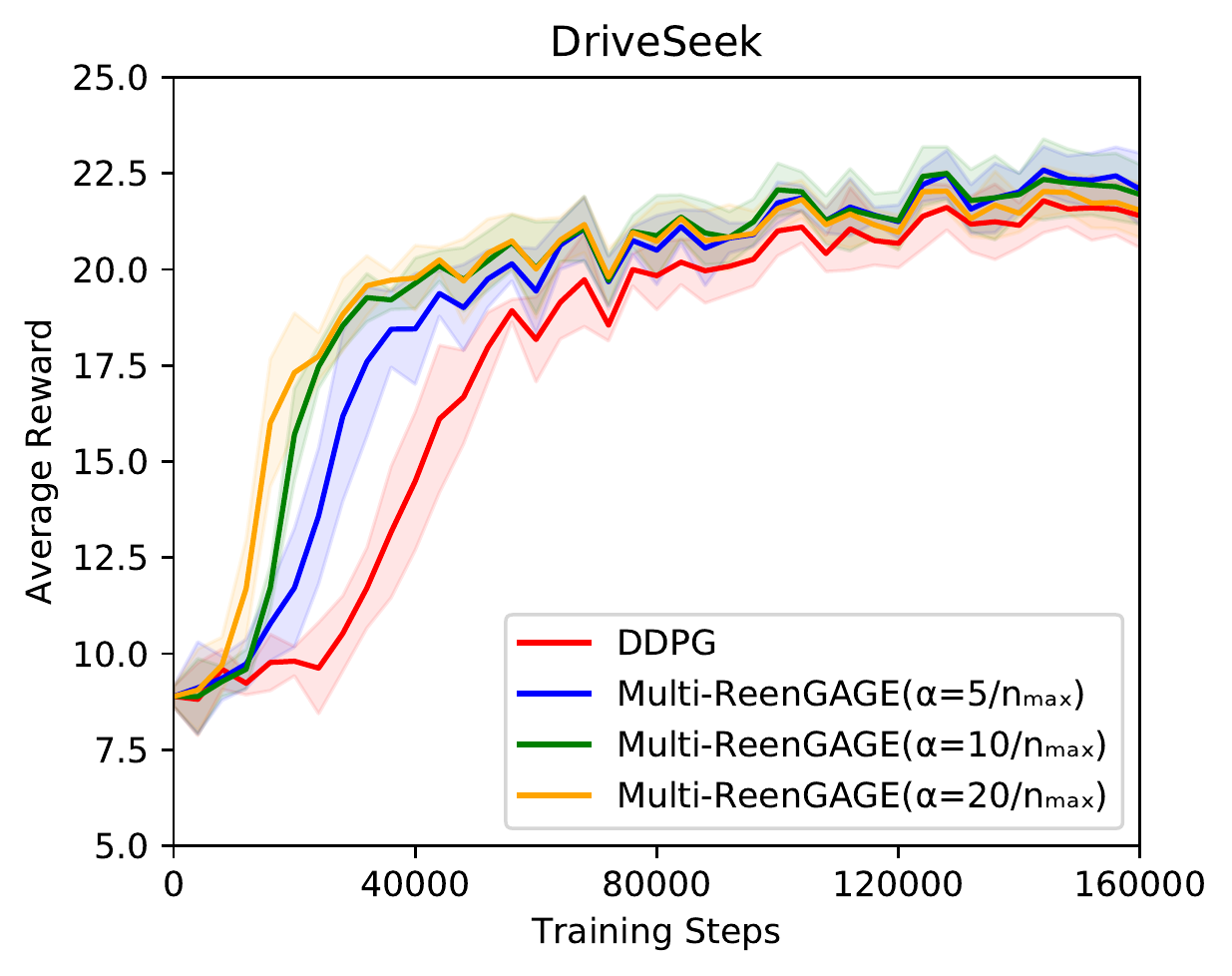}
         \caption{}
     \end{subfigure}
     \hfill
     \begin{subfigure}[b]{0.20\textwidth}
         \centering
         \includegraphics[width=\textwidth]{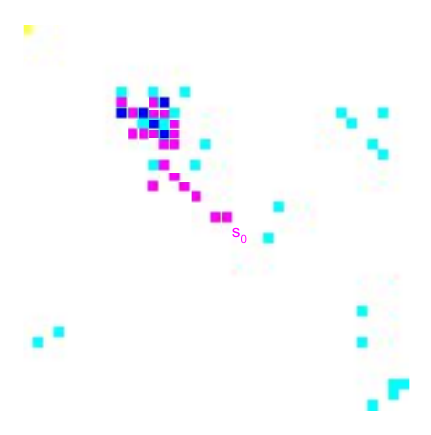}
         \caption{}
         \vspace{0.5cm}
     \end{subfigure}
          \hfill
     \begin{subfigure}[b]{0.29\textwidth}
         \centering
         \includegraphics[width=\textwidth]{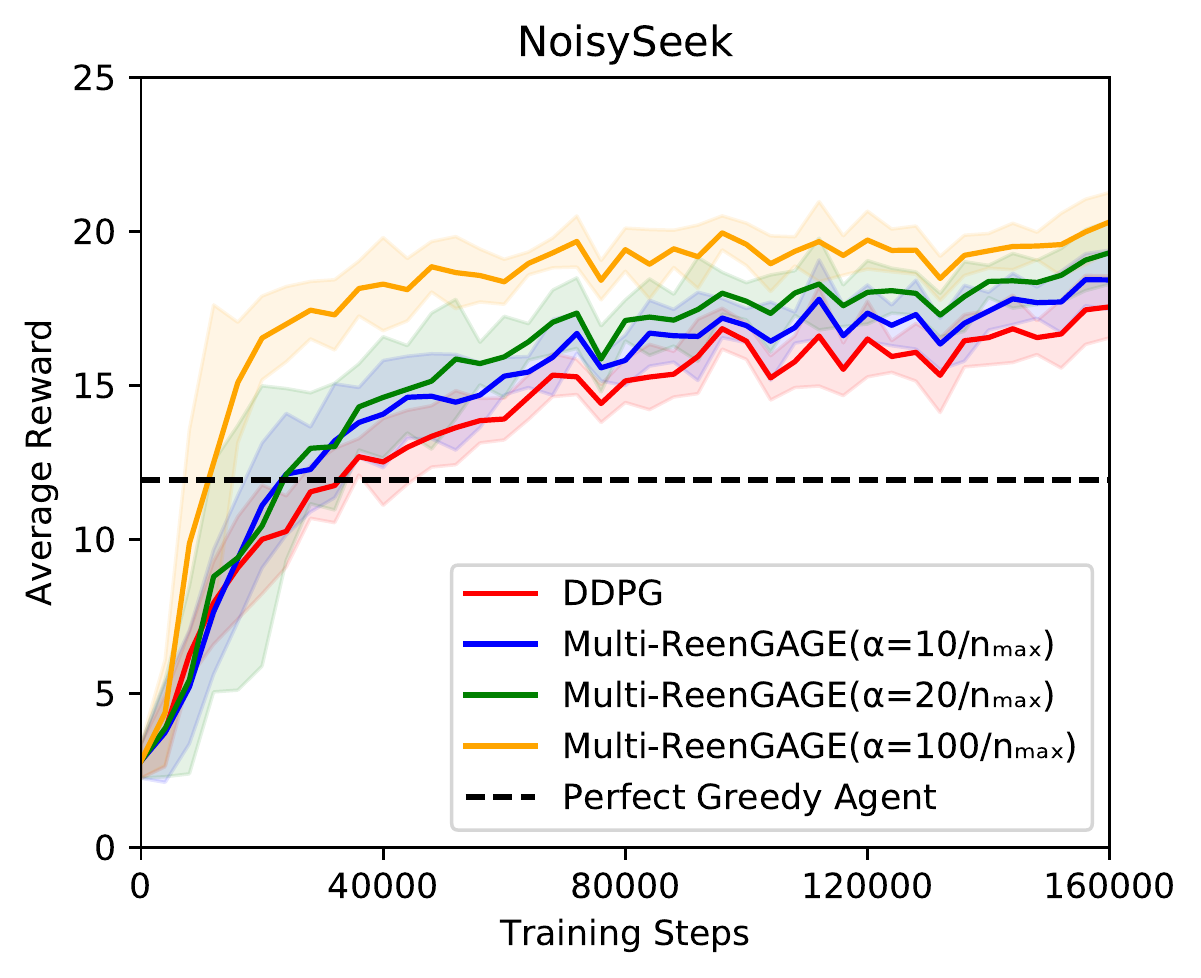}
         \caption{}
     \end{subfigure}
        \caption{Multi-ReenGAGE results. (a) and (c): Illustrations of DriveSeek and NoisySeek environments, respectively: cyan points show goals that were never reached, blue points show goals that were reached, and magenta points show (rounded) non-goal states that were reached. In (c), we see a NoisyReach agent (correctly) avoiding the trap of going to the nearby isolated points in favor of seeking the larger cluster. (b) and (d): Results for DriveSeek and NoisySeek, respectively. We see that Multi-ReenGAGE substantially improves over standard DDPG for both tasks. Lines are an average of (the same) 5 random seeds. For NoisySeek, we also show the performance of a perfect ``greedy'' agent, which simply goes towards the nearest individual goal. For NoisySeek, evaluations with more values of $\alpha$ are included in the appendix.  Note that for both experiments, the agent takes as input a list of goal coordinates, rather than an image: the agents do not use convolutional layers to interpret the goals. (On DriveSeek, where the coordinates are bounded, we attempted learning from images as well; ReenGAGE still outperformed DDPG, but overall performance was worse for both -- this experiment is presented in the appendix.) }
        \label{fig:multireengate_results}
\end{figure*}
In this section, we propose a variant of ReenGAGE for a specific class of RL environments: environments where the agent is rewarded for achieving any goal in a large set of arbitrary sparse goals, all of which are specified at test time. 
Formally, we consider goals in the form $g= \{g_1,...g_n\}$, where $n$ may vary but $n  \leq n_{\max}$. We consider $\{0,1\}$ rewards, where the reward function takes the form:
\begin{equation}
    R(s',g)  = \begin{cases}
    1  & \text{ if }  \underset{g_i\in g}{\lor} (R_\text{item}(s',g_i) = 1)\\
    0 & \text{ otherwise }
    \end{cases}.
\end{equation}
In our experiments, we only consider cases where the goals are mutually exclusive, so this is equivalent to:
\begin{equation}
    R(s',g)   = \sum_{g_i\in g} R_\text{item}(s',g_i).
\end{equation}
We assume that either: (i) the function $ R_\text{item}$ is known \textit{a priori} to the agent, or (ii) the item rewards $R_\text{item}(s',g_i)$ are observed separately for each goal $g_i$ at each time step during training.
This scenario presents several challenges. Firstly, many goal relabeling strategies cannot be directly applied here: strategies such as HER \cite{andrychowicz2017hindsight} assume that achieved states can be \textit{projected down} into the space of goals. In this case, the space of goals is \textit{much larger} than the space of possible states, so this assumption is broken. Secondly, we suggest that standard Q-learning is somewhat unsuited to this kind of problem, because it loses information about \textit{which} goal led to a reward. For instance, if there are 100 goals $g_i$, and a reward is received for a certain state $s'$, there is no direct indication of which goal was satisfied. This means that a very large number of episodes may need to be run in order to learn the effect of \textit{each individual} goal on the reward.

We now describe our approach. For concreteness, we will assume that the agent uses an architecture based on DeepSets \cite{NIPS2017_f22e4747} to process the goal set input (although we believe our technique can likely be adapted to using more complex neural set architectures, such as Set Transformer \cite{pmlr-v97-lee19d}). Concretely, this means that our Q-function takes the form:
\begin{equation}
Q_\theta(s,a,g) := Q^{\text{head}}_{\theta^\text{h.}}(s,a,\sum_{g_i\in g}[Q_{\theta^\text{e.}}^{\text{encoder}}(s,g_i)])  \label{eq:multi_q_func}
\end{equation}
and the policy has a similar architecture. Note that $Q_{\theta^\text{e.}}^{\text{encoder}}$ outputs a vector-valued embedding for a given goal $g_i$. From this baseline, introduce a set of scalar \textit{gate variables} $b_i$:
\begin{equation}
Q_\theta(s,a,g) := Q^{\text{head}}_{\theta^\text{h.}}(s,a,\sum_{i= 1}^n[b_i Q_{\theta^\text{e.}}^{\text{encoder}}(s,g_i)]). 
\end{equation}
Each gate $b_i$ is set to 1. However, if $b_i$ were zero, this would be equivalent to the goal $g_i$ being absent from the set $g$. We then treat the gate variables as \textit{differentiable}. If a certain goal $g_i$ contributes to the Q function (i.e., if it is likely to be satisfied), then we expect $Q_\theta(s,a,g)$ to be \textit{highly sensitive} to $b_i$; in other words, we expect $\frac{\partial Q_\theta}{\partial b_i}$ to be large. Then $\nabla_b Q$ represents the importance of each goal to the Q-value function. Our key insight is that we can use a ReenGAGE-style loss to transfer $\nabla_b Q$ from target to current Q-value estimate, therefore preserving attention on the relevant goals. 

However, this requires us to have a value for $\nabla_b R(s',g)$. Note that the reward can be written as:
\begin{equation}
    R(s',g)   = \sum_{g_i\in g} b_i R_\text{item}(s',g_i). 
\end{equation}
Setting $b_i=0$ is again like $g_i$ being absent. Interpolating: 
\begin{equation}
    \frac{\partial R}{\partial b_i}   := R_\text{item}(s',g_i)
\end{equation}
which gives us a ``ground-truth'' reward gradient we can compute. This yields the following loss function:
\begin{equation}
\begin{split}
        &\mathcal{L_\textbf{Multi-ReenGAGE}} =  
         \mathcal{L_\textbf{DDPG-Critic}} + \alpha  \mathcal{L_{\text{mse}}} \Big[\nabla_{b} Q_\theta(s,a,g), \\
        & R_\text{item}(s',g)  + \gamma \nabla_b Q_{\theta'}  (s',\pi_{\phi'}(s', g), g)\Big] 
\end{split}
\end{equation}
where $[R_\text{item}(s',g)]_i := R_\text{item}(s',g_i)$. In practice, we make two modifications to this algorithm. First, we use $b_i^2$ as the gate rather than $b_i$.\footnote{And use $2R_\text{item}(s',g)$ instead of $R_\text{item}(s',g)$. } While algebraically this should do nothing but multiply the gradient loss term by 4, it is important for vectorized implementation; see the appendix for details.  

Second, we share the encoder $Q^{\text{encoder}}$ between the Q-function and the policy $\pi$. This is so the policy does not have to learn to interpret the goal set ``from scratch'' and is empirically important (see ablation study in the appendix). We train the encoder only during critic training.
\subsection{Experiments}
We test Multi-ReenGAGE on two environments: \textbf{DriveSeek} and \textbf{NoisySeek}. Both environments are constructed such that a ``greedy'' strategy of simply going to the closest individual goal is not optimal, so the entire goal set must be considered. We describe the environments informally here and provide additional detail in the appendix.

\textbf{DriveSeek} is a deterministic environment, where the continuous position $s_{\text{pos.}} \in [-10,10]^2$ always moves with constant $\ell_2$ speed $1$, in a direction determined by a velocity vector $s_{\text{vel.}}$ on the unit circle. At each step, the agent takes a 1-dimensional action $a \in [-0.5,0.5]$, which represents \textit{angular acceleration}: it specifies an angle in radians which is added to the angle of the velocity vector. At the edges of the space, the state position wraps around to the opposite edge.

The objective is to reach any of up to $n_\text{max} = 200$ goals. The goals all lie on integer coordinates in $[-10,10]^2$, and the agent receives a reward if its coordinates round to a goal. In addition to $s_{\text{pos.}}$ and $s_{\text{vel.}}$, the agent receives an observation of its current rounded position. The agent also receives as input the current list of goal coordinates. Note that the agent cannot simply stay at a single goal, or take an arbitrary path between goals: it is constrained to making wide turns. Therefore all goals must be considered in planning an optimal trajectory. 

\textbf{NoisySeek} is a randomized environment. In it, $s\in \mathbb{R}^2$, $a \in \mathbb{R}^2$, with $\|a\|_2 \leq 1$, and the transition function is defined as $\mathcal{T}(s,a) \sim \mathcal{N}( s + a , I)$. In other words, the agent moves through space at a capped speed, and noise is constantly added to the position. The goals are defined as integer coordinates in a similar manner to in DriveSeek, but without a box constraint. Additionally, the goal distribution is such that goals tend to be clustered together. Note that a ``greedy'' agent that simply goes to the nearest goal is suboptimal, because the probability of consistently reaching that one goal is low: it is better to seek clusters.

Results are presented in Figure \ref{fig:multireengate_results}. We see that Multi-ReenGAGE substantially outperforms the baseline of DDPG on both environments.
\section{Theoretical Properties}
\subsection{Bias}
In the Preliminaries section, we discuss that goal relabeling strategies can exhibit bias if Equation \ref{eq:relabeling} is not respected. In the dense reward case, our method does not cause bias of this sort (although such bias may be present if our method is combined with a relabeling strategy.) However, in the sparse reward case, if the transitions are nondeterministic, our method may cause a similar bias. In particular, note that, in the sparse case, Equation \ref{eq:sparse_method} effectively trains the gradient of the Q-value function to match the following target:
\begin{equation}
\begin{split}
        &\nabla_g Q_\theta(s,a,g) := \\ &\mathop\mathbb{E}_{s'\sim \mathcal{T}(s,a)} \hspace{-0.2cm} [ \gamma  \nabla_g Q_{\theta'}  (s',\pi_{\phi'}(s', g), g) | R(s',g) =c_\text{low}]  \hspace{-0.05cm} \approx\\
         & \nabla_g \mathop\mathbb{E}_{s'\sim \mathcal{T}(s,a)} \Big[  R(s',g) +\\
         &\qquad \qquad  \quad \gamma  Q_{\theta'}  (s',\pi_{\phi'}(s', g), g) | R(s',g) =c_\text{low}\Big] \\ \label{eq:bias}
\end{split}
\end{equation}
where the last line holds exactly if the ``boundary'' points where $R(s',g) =c_\text{low}$ but $\nabla_g R(s',g) \neq \bm{0}$ are of measure zero (and the derivative is defined at all such points).

Equation \ref{eq:bias} shows us that, in the sparse case, our method trains the gradient of the Q-function to match an expected target gradient where the expectation is taken over a biased distribution: if \begin{equation}
  \Pr_{\mathcal{T}} [s'|s,a; R(s',g) = c_\text{low}] \neq \Pr_{\mathcal{T}} [s'|s,a],
\end{equation}
then this will cause bias in our target gradient estimate, in a similar manner to the hindsight bias of HER described by \cite{schroecker2021universal}. 

Note that this is only an issue in nondeterministic environments: in deterministic environments, for a given $(s,a,g)$, either $R(s',g)$ is always $\neq c_\text{low}$, in which case the gradient term is never involved in training, or $R(s',g)$ is always $ c_\text{low}$, in which case 
\begin{equation}
  \Pr_{\mathcal{T}} [s'|s,a; R(s',g) = c_\text{low}] = \Pr_{\mathcal{T}} [s'|s,a] = 1.
\end{equation}
\subsection{Learning Efficiency}
In this section, we provide examples of classes of environments for which our method will result in provably more efficient learning than standard DDPG-style updates. We treat both the dense reward case (in which we have access to the gradient of the reward function) and the sparse reward case (in which we do not).
\subsubsection{Dense Reward Example}
\begin{figure}
    \centering
    \includegraphics[width=0.25\textwidth]{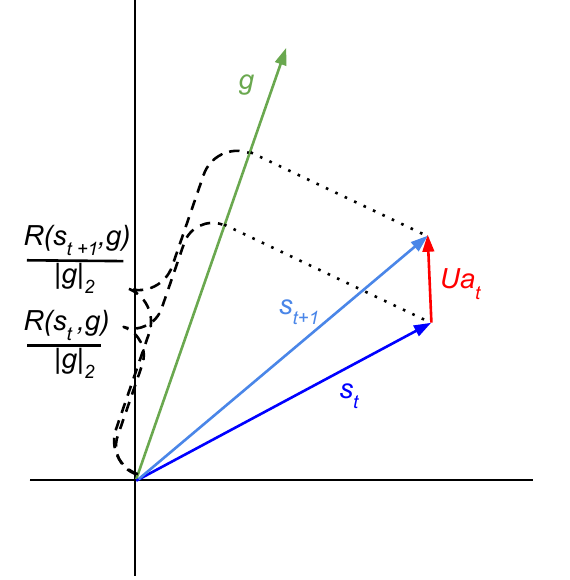}
    \caption{Example dense reward environment class. The agent receives a reward proportional to the projection of the state vector onto the goal vector at each step, and can change the state vector by adding an action vector with $\ell_2$ distance up to $1$ at each step. However, at each step, this action is distorted by an unknown rotation $U$: the agent must learn to compensate for this distortion. }
    \label{fig:dense_theory_example}
\end{figure}
Consider the class of simple, deterministic environments described as follows:
\begin{itemize}
    \item $g,s,a\in \mathbb{R}^d$; $\|a\|_2 \leq 1$
    \item  $R(s,g) = g^Ts$
    \item $\mathcal{T}(s,a) = s + Ua$, where $U\in \mathbb{R}^{d \times d}$ is an unknown orthogonal (rotation) matrix.
\end{itemize}
This environment class is illustrated in Figure \ref{fig:dense_theory_example}. Environments of this class are parameterized by $U$, so the learning task is to estimate $U$. We make the following assumptions:
\begin{itemize}
    \item The ``hypothesis class'' consists of all environments with dynamics of the type described above. We therefore take as an inductive bias that each model in the considered model class consists of a $Q$-function $Q_{\tilde{U}}$ and policy $\pi_{\tilde{U}}$ which are in the form of the optimal $Q$-function and policy for an estimate of $U$, notated as $\tilde{U} \in \mathbb{R}^{d\times d}$ (constrained to be orthogonal).
    \item We assume that $Q_{\tilde{U}}$ and $\pi_{\tilde{U}}$ share the same estimated parameter $\tilde{U}$. (This is analogous to -- although admittedly stronger than -- the parameter sharing we used for Multi-ReenGAGE.) Taken with the above assumption, this implies that $a = \pi_{\tilde{U}}(s,g)$ maximizes $Q_{\tilde{U}}(s,a,g)$, so we do not need to train $\pi$ separately. Similar parameter sharing occurs between the target policy and Q-function.
    \item We are comparing our method, minimizing the loss in Equation \ref{eq:method_dense}, with minimizing the ``vanilla'' DDPG loss (Equation \ref{eq:standard_critic_loss}). 
    \item States and actions in the replay buffer are in general position.
\end{itemize}
In this case, the following proposition holds:
\begin{proposition}
Under the above assumptions, minimizing the ReenGAGE loss can learn $U$ (and therefore learn the optimal policy) using $O(d)$ unique replay buffer transitions. However, minimizing the standard DDPG loss requires at least $O(d^2)$ unique transitions to successfully learn $U$. \label{prop:dense}
\end{proposition}
Proofs are provided in the appendix. This result shows that, in some cases, ReenGAGE requires asymptotically less replay data to successfully learn to perform a task than standard DDPG.

\subsubsection{Sparse Reward Case}
\begin{figure}
    \centering
    \includegraphics[width=0.47\textwidth]{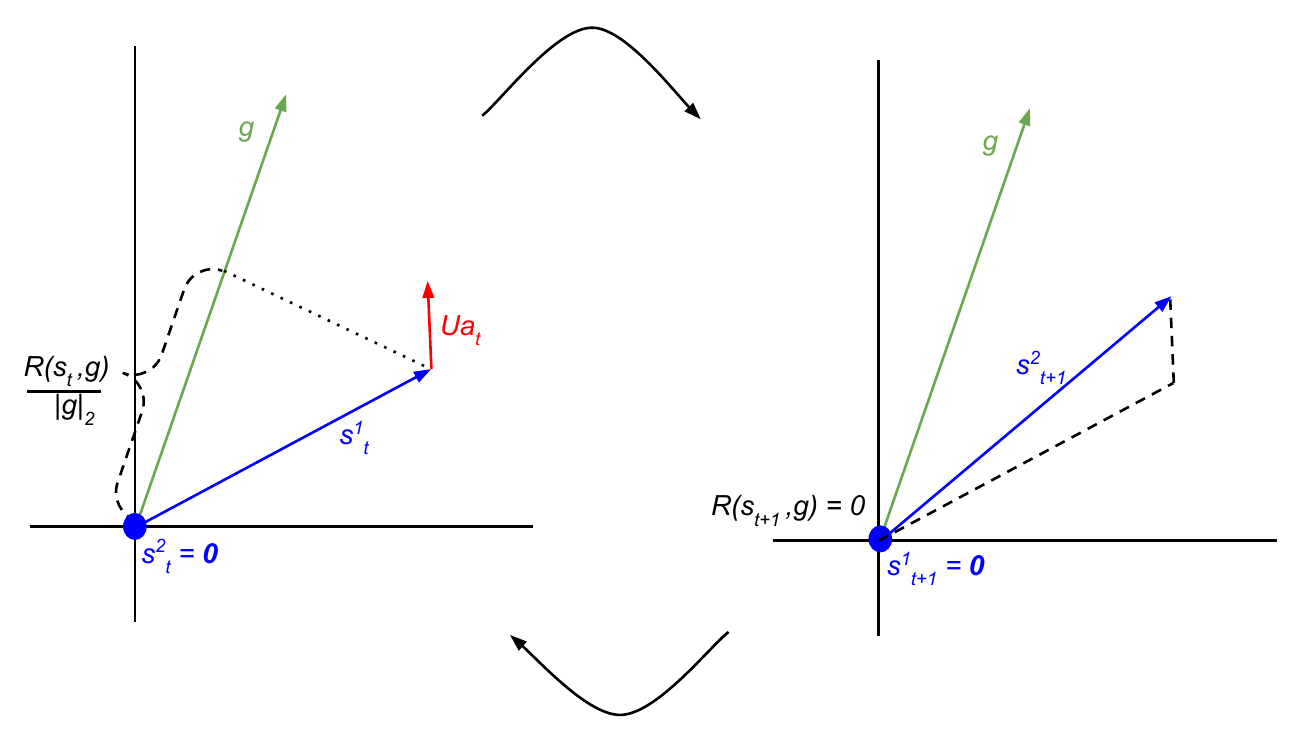}
    \caption{Example sparse reward environment class. If $s^1$ is initially nonzero, then the (rotated) action $Ua$ is added to it, as in the dense case. However, the resulting vector is immediately ``stored'' in $s^2$, and $s^1$ is zeroed: this means that no immediate reward is obtained. In the next step, with $s^1$ zero, the action is ignored and $s^2$ is ``reloaded'' into $s^1$, resulting in a reward that depends on the \textit{previous} action.  }
    \label{fig:sparse_theory_example}
\end{figure}

The result shown above might be unsurprising to many readers. Specifically, because the gradient $\nabla_g R(s',g)$ is used by our method and not by standard DDPG, in the dense-reward case, our method is utilizing more information from the environment (to the extent that $R$, which we assume that agents know \textit{a priori}, is part of the ``environment'') than the standard algorithm. However, here we show a class of \textit{sparse reward} environments for which the same result holds, despite $\nabla_g R(s',g)$ being unavailable. The environments are constructed as follows:
\begin{itemize}
    \item $g,a\in \mathbb{R}^d$; $\|a\|_2 \leq 1$; $s\in \mathbb{R}^{2d}$; the state vector consists of two halves, denoted $s^1,s^2$; we write $s$ as $(s^1;s^2)$.
    \item  $R(s,g) = g^Ts^1$
    \item $\mathcal{T}(s,a) =\begin{cases}
    (\bm{0};s^1+Ua)&\,\,\,\,\text{ if } s^1 \neq \bm{0}\\
    (s^2;\bm{0})&\,\,\,\,\text{ if } s^1 = \bm{0}\\
    \end{cases}$
    \item $U\in \mathbb{R}^{d \times d}$ is an unknown orthogonal (rotation) matrix.
    \item We define $c_\text{low} = 0$.
\end{itemize}
See Figure \ref{fig:sparse_theory_example} for an illustration. Note that this satisfies sparseness properties: namely, $R(s',g) =0 = c_\text{low}$ at least every other step; and, when $R(s',g) =0$, then $\nabla_g R(s',g) = \bm{0}$ (assuming general position). It is also deterministic, so we do not need to worry about the bias discussed in the previous section. We can therefore apply the sparse version of our method (Equation \ref{eq:sparse_method}), which does \textit{not} use gradient feedback from the reward:
\begin{proposition}
Under the same assumptions as Proposition \ref{prop:dense} (replacing Equation \ref{eq:method_dense} with Equation \ref{eq:sparse_method}), minimizing the ReenGAGE loss can learn $U$ in the sparse environment class using $O(d)$ unique replay buffer transitions. However, minimizing the standard DDPG loss requires at least $O(d^2)$ unique transitions to successfully learn $U$. \label{prop:sparse}
\end{proposition}
This example is admittedly a bit contrived: the single-step reward can always be computed without knowledge of the parameter $U$. However, it may still give insight about real-world scenarios in which predicting immediate reward is much easier than understanding long-term dynamics.

Note that these two scaling results apply to the number of \textit{replay buffer transitions}. In particular, if a goal relabeling algorithm is used on top of DDPG, then $O(d^2)$ replay buffer transitions may be able to be constructed from $O(d)$ observed training rollout transitions, so standard DDPG \textit{combined with goal relabeling} might only require $O(d)$ training rollout transitions. However, this would be computationally expensive, and may not work in practice for particular goal relabeling algorithms. (HER, for instance, only relabels using achieved states from the same episode: if the episode length is $O(1)$ in $d$, then $O(d^2)$ observed training rollout transitions would still be required for DDPG+HER.) Also, goal relabeling techniques require \textit{a priori} knowledge of the function $R$, while in the sparse example, ReenGAGE does not (although in the case of this example, we assume that we are using the correct ``hypothesis class'', i.e., the functional form of $Q_{\tilde{U}}$: constructing this in practice would likely require knowing $R$).

\section{Related Works}
Many prior approaches have been taken to the goal-conditioned reinforcement learning problem \cite{pmlr-v37-schaul15}. See \cite{ijcai2022-770} for a recent survey of this area. One line of work for this problem involves \textit{automated curriculum generation}: here, the idea is to select goals during training that that are dynamically chosen to be the most informative \cite{florensa2018automatic,sukhbaatar2018intrinsic,zhang2020automatic}. In the off-policy reinforcement learning setting, a related technique becomes a possibility: one can re-label past experiences with counter-factual goals. This allows a single experienced transition to be used to train for multiple goals, and the re-labeled goals can be chosen using various heuristics to improve training \cite{andrychowicz2017hindsight,nair2018visual,yang2021mher,fang2019curriculum}. Note that our proposed method can be combined with any of these off-policy techniques. \cite{schroecker2021universal} discusses bias that can result from some goal relabeling techniques. \cite{eysenbach2021clearning} proposes a method based on recursive classification which is in practice similar to hindsight relabeling, but requires less parameter tuning.

In alternative approaches to goal-conditioned RL, \cite{eysenbach2022contrastive} has proposed using an on-policy goal-conditioned reinforcement learning technique, using contrastive learning, while \cite{janner2022planning} and \cite{janner2021offline} propose model-based techniques. 

Note that our proposed method is distinct from \textit{policy distillation} \cite{rusu2016policy}: the goal of policy distillation is to consolidate one or more \textit{already trained} policy networks into a smaller network; whereas our method is intended to improve initial training.
Some prior \cite{manchin2019reinforcement,choi2017multi,Wu_Khetarpal_Precup_2021} and concurrent \cite{bertoin2022look} works have focused on using attention-based mechanisms to improve either the performance or interpretability of reinforcement learning algorithms. However, to our knowledge, ours is the first to apply gradient-based attention transfer to the critic update to enhance goal-conditioned off-policy reinforcement learning.

Some prior works have been proposed for goal-seeking with structured, complex goals made up of sub-goals, similar to (and in some cases more general than) the multi-goal setting that Multi-ReenGAGE is designed for. Some of these works \cite{oh2017zero,sohn2018hierarchical} use a hierarchical policy; however, such a structure may be unable to represent the true optimal policy \cite{vaezipoor2021ltl2action}. \cite{vaezipoor2021ltl2action} proposes a method without this limitation; although the setting considered (Linear Temporal Logic) is different from the multi-goal setting considered here, in that a reward is achieved at most once per episode. \cite{touati2021learning} proposes a method for arbitrary reward functions specified at test time, under discrete action spaces; in concurrent work \cite{touati2023does}, this is generalized to continuous actions. \cite{janner2022planning}, a model-based technique mentioned above, can use reward function gradient information to adapt to an arbitrary \textit{shaped} (i.e., non-sparse) reward function at test time.

\section{Limitations and Conclusion}
ReenGAGE has some important limitations. For example, we have seen that the hyper-parameter $\alpha$ requires tuning and can vary greatly (likely due to diverse scales in goal coordinates and rewards); and the benefits of ReenGAGE seem limited to tasks with high goal dimension.

Still, ReenGAGE represents a novel approach to goal-conditioned RL, with benefits demonstrated both empirically and theoretically. In future work, we are particularly interested in exploring the use of Multi-ReenGAGE in safety and robustness applications. In particular, the ability to encode many simultaneous goals at test time could allow the agent to consider many ``backup'' goals, all of which are acceptable, rather than forcing the agent to focus only a single goal (resulting in total failure if that goal is unreachable.)
\section{Acknowledgements}
This project was supported in part by NSF CAREER AWARD 1942230 and ONR YIP award N00014-22-1-2271.
\bibliography{aaai23}
\newpage
\onecolumn 
\appendix
\section{Proofs}

\subsection{Proposition \ref{prop:dense} (Gradient reward feedback example)}
\subsubsection{Problem setting}
\begin{itemize}
    \item $g,s,a\in \mathbb{R}^n$; $\|a\|_2 \leq 1$
    \item Single step reward $R(s,g) = g^Ts$
    \item Transition function: $s_{t+1} := s_t + Ua_t$, where $U\in \mathbb{R}^{n\times n}$ is an unknown orthogonal (rotation) matrix
    \item The ``hypothesis class'' consists of all environments with dynamics of this form: the learning task is to learn the unknown rotation matrix $U$. We therefore take as an ``inductive bias'' that the model class consists of a $Q$-function $Q_{\tilde{U}}$ and policy $\pi_{\tilde{U}}$ which are in the form of the optimal $Q$-function and policy for an estimate $\tilde{U} \in \mathbb{R}^{n\times n}$ of $U$ (constrained to be orthogonal). The task is to learn this parameter. We assume that $Q_{\tilde{U}}$ and $\pi_{\tilde{U}}$ share the parameter estimate.
    \item Let $\tilde{U}$ be the ``current'' parameter estimate and   $\bar{U}$ be the ``target'' parameter estimate for TD updates.
    \item We assume actions in the replay buffer are in general position.
\end{itemize}
\subsubsection{Analysis}
The optimal Q-value function is of the form:
\begin{equation}
\begin{split}
    Q^*(s,a,g)& = \sum^{\infty}_{n=0} \gamma^n \left(g^T(s + Ua)+n\|g\|_2\right)\\
    & = \frac{g^T(s+Ua)}{1-\gamma} + \frac{\gamma\|g\|_2}{(1-\gamma)^2}
\end{split}
\end{equation}
The optimal policy $\pi$ then takes the form:
\begin{equation}
    a^* = \frac{U^Tg}{\|g\|_2}
\end{equation}
(Note that, because we are sharing the parameter $\tilde{U}$ between $Q_{\tilde{U}}$ and $\pi_{\tilde{U}}$, and because the optimal action for $Q_{\tilde{U}}$ can be written in closed form, we do not require a training step for $\pi_{\tilde{U}}$.)

The ``standard'' MSE Bellman error for a tuple $(s,a,g,s')$ is 
\begin{align*}
[ Q_{\tilde{U}}(s,a,g)&\,\,\,\boldsymbol{-}\,\,\,R(s',g) + \gamma  Q_{\bar{U}}(s',\pi_{\bar{U}}(s,g),g) ]^2=\\
\Biggl[\frac{g^T(s+\tilde{U}a)}{1-\gamma} + \frac{\gamma\|g\|_2}{(1-\gamma)^2}&\,\,\,\boldsymbol{-}\,\,\,g^Ts' +\gamma\left(\frac{g^T(s'+\bar{U}\frac{\bar{U}^Tg}{\|g\|_2})}{1-\gamma} + \frac{\gamma\|g\|_2}{(1-\gamma)^2} \right)\Biggr]^2=\\
\Biggl[\frac{g^T(s+\tilde{U}a)}{1-\gamma} + \frac{\gamma\|g\|_2}{(1-\gamma)^2}&\,\,\,\boldsymbol{-}\,\,\,g^Ts' +\frac{\gamma}{1-\gamma}g^Ts' + \frac{\gamma}{1-\gamma} \|g\|_2 + \frac{\gamma^2\|g\|_2}{(1-\gamma)^2}\Biggr]^2= \text{(using }\bar{U}\bar{U}^T = I\text{)}\\
\Biggl[\frac{g^T(s+\tilde{U}a)}{1-\gamma} + \frac{\gamma\|g\|_2}{(1-\gamma)^2}&\,\,\,\boldsymbol{-}\,\,\, \frac{g^Ts'}{1-\gamma} + \frac{\gamma\|g\|_2}{(1-\gamma)^2}\Biggr]^2=\\
\Biggl[\frac{g^T(s+\tilde{U}a)}{1-\gamma}  &\,\,\,\boldsymbol{-}\,\,\, \frac{g^Ts'}{1-\gamma}  \Biggr]^2=\\
\Biggl[g^T(s+\tilde{U}a)  &\,\,\,\boldsymbol{-}\,\,\, g^Ts'  \Biggr]^2= \text{ (dropping constant factor)}\\
\Biggl[g^T\tilde{U}a  &\,\,\,\boldsymbol{-}\,\,\, g^T(s'-s)  \Biggr]^2= \\
\Biggl[g^T\tilde{U}a  &\,\,\,\boldsymbol{-}\,\,\, g^TUa  \Biggr]^2\\
\end{align*}
This is equivalent to fitting the bilinear form $g^T\tilde{U}a$ using the observed scalar $g^TUa \,\,\,(= g^T(s'-s))$. Noting that orthogonal matrices have $\sim n^2/2$ degrees of freedom, 
\textbf{this will require at least $O(n^2)$ samples to learn}.\\
Now consider our gradient-based Bellman error:\\
\begin{align*}
\|\nabla_g Q_{\tilde{U}}(s,a,g)&\,\,\,\boldsymbol{-}\,\,\,\nabla_g[R(s',g) + \gamma  Q_{\bar{U}}(s',\pi_{\bar{U}}(s,g),g)] \|_2^2=\\
\Biggl\|\nabla_g\left[\frac{g^T(s+\tilde{U}a)}{1-\gamma} + \frac{\gamma\|g\|_2}{(1-\gamma)^2}\right]&\,\,\,\boldsymbol{-}\,\,\,\nabla_g\left[g^Ts' +\gamma\left(\frac{g^T(s'+\bar{U}\frac{\bar{U}^Tg}{\|g\|_2})}{1-\gamma} + \frac{\gamma\|g\|_2}{(1-\gamma)^2} \right)\Biggr]\right\|_2^2=\\
\Biggl\|\nabla_g\left[\frac{g^T(s+\tilde{U}a)}{1-\gamma}\right]  &\,\,\,\boldsymbol{-}\,\,\, \nabla_g\left[\frac{g^Ts'}{1-\gamma}\right]  \Biggr\|_2^2=\\
\Biggl\|\frac{(s+\tilde{U}a)}{1-\gamma}  &\,\,\,\boldsymbol{-}\,\,\, \frac{s'}{1-\gamma} \Biggr\|_2^2=\\
\Biggl\|(s+\tilde{U}a)  &\,\,\,\boldsymbol{-}\,\,\, s'  \Biggr\|_2^2= \text{ (dropping constant factor)}\\
\Biggl\|\tilde{U}a  &\,\,\,\boldsymbol{-}\,\,\, (s'-s)  \Biggr\|_2^2= \\
\Biggl\|\tilde{U}a  &\,\,\,\boldsymbol{-}\,\,\, Ua  \Biggr\|_2^2 \\
\end{align*}
Here, we are fitting the linear form $\tilde{U}a $ to the  observed vector $(s'-s)$. Assuming general position, \textbf{this can be solved with $O(n)$ samples!}
\subsection{{Proposition \ref{prop:sparse} (No reward gradient case example)}}
\subsubsection{Problem setting}

\begin{itemize}
    \item $g,a \in \mathbb{R}^n$;$\|a\|_2 \leq 1$; $s\in \mathbb{R}^{2n}$; the state vector consists of two halves, which we denote $s^1$ and $s^2$.
    \item Single step reward $R(s,g) = g^Ts^1$
    \item Transition function:
    \begin{itemize}
        \item If $s_t^1 \neq \bm{0}$, then $s^1_{t+1} :=\bm{0}$, $s^2_{t+1} := s^1_t + Ua_t$. (Note that the reward is always zero here.)
        \item If $s_t^1 = \bm{0}$, then $s^1_{t+1} :=s^2_t$, $s^2_{t+1} := \bm{0}$, 
    \end{itemize}
 where $U\in \mathbb{R}^{n\times n}$ is an unknown orthogonal (rotation) matrix
 \item We make the same assumptions about the hypothesis class and inductive bias as in the last example, and also assume general position.
\end{itemize}
\subsection{Analysis}
Optimal  Q-function
\begin{equation}
\begin{split}
    Q^*(s,a,g)& = 
    \begin{cases} \sum^{\infty}_{n\text{ odd}} \gamma^n \left(g^T(s^1 + Ua)+\frac{n-1}{2}\|g\|_2\right)
     &\text{ if } s^1  \neq \bm{0}\\\sum^{\infty}_{n\text{ even}} \gamma^n \left(g^Ts^2+\frac{n}{2}\|g\|_2\right)
     &\text{ if } s^1  = \bm{0}\\\end{cases}\\
    & =\begin{cases} \sum^{\infty}_{m=0} \gamma^{2m+1} \left(g^T(s^1 + Ua)+m\|g\|_2\right)
     &\text{ if } s^1  \neq \bm{0}\\\sum^{\infty}_{m=0} \gamma^{2m} \left(g^Ts^2+m\|g\|_2\right)
     &\text{ if } s^1  = \bm{0}\\\end{cases}\\
& =\begin{cases} \frac{\gamma g^T(s^1+Ua)}{1-\gamma^2} + \frac{\gamma^3\|g\|_2}{(1-\gamma^2)^2}
     &\text{ if } s^1  \neq \bm{0}\\ \frac{g^Ts^2}{1-\gamma^2} + \frac{\gamma^2\|g\|_2}{(1-\gamma^2)^2}
     &\text{ if } s^1  = \bm{0}\\\end{cases}\\
\end{split}
\end{equation}
The optimal policy $\pi$ then takes the form:
\begin{equation}
    a^* = \frac{U^Tg}{\|g\|_2}
\end{equation}
(Note that if  $ s^1  = \bm{0}$, then the action does not appear in the $Q$ function, so any action is optimal.)

We now consider the standard TD error:
\begin{equation}
    [ Q_{\tilde{U}}(s,a,g)\,\,\,\boldsymbol{-}\,\,\,R(s',g) + \gamma  Q_{\bar{U}}(s',\pi_{\bar{U}}(s,g),g) ]^2
\end{equation}
Note that if $ s^1  = \bm{0}$, then the trainable parameter $\tilde{U}$ does not appear in the expression of $ Q_{\tilde{U}}(s,a,g)$. Then no learning can occur from these tuples; we can instead only consider the case where $ s^1  \neq 0$. In this case, the immediate reward is always zero, and we also know that $s^{1\prime} = \bm{0}$ so the ``standard'' TD update is:

\begin{align*}
[ Q_{\tilde{U}}(s,a,g)&\,\,\,\boldsymbol{-}\,\,\,  \gamma  Q_{\bar{U}}(s',\pi_{\bar{U}}(s,g),g) ]^2=\\
\Biggl[ \frac{\gamma g^T(s^1+\tilde{U}a)}{1-\gamma^2} + \frac{\gamma^3\|g\|_2}{(1-\gamma^2)^2} &\,\,\,\boldsymbol{-}\,\,\,\gamma\left(  \frac{g^Ts^{2\prime}}{1-\gamma^2} + \frac{\gamma^2\|g\|_2}{(1-\gamma^2)^2} \right)\Biggr]^2=\\
\Biggl[ \frac{\gamma g^T(s^1+\tilde{U}a)}{1-\gamma^2}  &\,\,\,\boldsymbol{-}\,\,\,  \frac{\gamma g^Ts^{2\prime}}{1-\gamma^2} \Biggr]^2=\\
\Biggl[g^T(s^1+\tilde{U}a)  &\,\,\,\boldsymbol{-}\,\,\, g^Ts^{2\prime}  \Biggr]^2= \text{ (dropping constant factor)}\\
\Biggl[g^T\tilde{U}a  &\,\,\,\boldsymbol{-}\,\,\, g^T(s^{2\prime}-s^1)  \Biggr]^2= \\
\Biggl[g^T\tilde{U}a  &\,\,\,\boldsymbol{-}\,\,\, g^TUa  \Biggr]^2\\
\end{align*}
This is the same update as in the previous example, and again \textbf{we need $O(n^2)$ samples}. (Note that there is a constant factor of 2 increase in the number of needed samples, due to the wasted samples in which $ s^1 =  \bm{0}$. However, assuming general position, this will only account for half of the replay buffer.)

We now consider the case in which we use our gradient TD update. Again we only can use  the tuples where $ s^1  \neq 0$. For all of these, the immediate reward is zero, so we can use the gradient-based update for all of them.

\begin{align*}
\| \nabla_g Q_{\tilde{U}}(s,a,g)&\,\,\,\boldsymbol{-}\,\,\,  \gamma  \nabla_g Q_{\bar{U}}(s',\pi_{\bar{U}}(s,g),g) \|_2^2=\\
\Biggl\|\nabla_g\left( \frac{\gamma g^T(s^1+\tilde{U}a)}{1-\gamma^2} + \frac{\gamma^3\|g\|_2}{(1-\gamma^2)^2}\right) &\,\,\,\boldsymbol{-}\,\,\,\gamma\nabla_g\left(  \frac{g^Ts^{2\prime}}{1-\gamma^2} + \frac{\gamma^2\|g\|_2}{(1-\gamma^2)^2} \right)\Biggr\|_2^2=\\
\Biggl\| \frac{\gamma (s^1+\tilde{U}a)}{1-\gamma^2}  &\,\,\,\boldsymbol{-}\,\,\,  \frac{\gamma s^{2\prime}}{1-\gamma^2} \Biggr\|_2^2=\\
\Biggl\|(s^1+\tilde{U}a)  &\,\,\,\boldsymbol{-}\,\,\, s^{2\prime}  \Biggr\|_2^2= \text{ (dropping constant factor)}\\
\Biggl\|\tilde{U}a  &\,\,\,\boldsymbol{-}\,\,\, (s^{2\prime}-s^1)  \Biggr\|_2^2= \\
\Biggl\|\tilde{U}a  &\,\,\,\boldsymbol{-}\,\,\, Ua  \Biggr\|_2^2\\
\end{align*}
 As in the previous gradient feedback case, assuming general position, \textbf{this can be solved with $O(n)$ samples!}

\section{ReenGAGE for Discrete Actions}
\begin{figure*}
     \centering
     \begin{subfigure}[b]{0.44\textwidth}
         \centering
         \includegraphics[width=\textwidth]{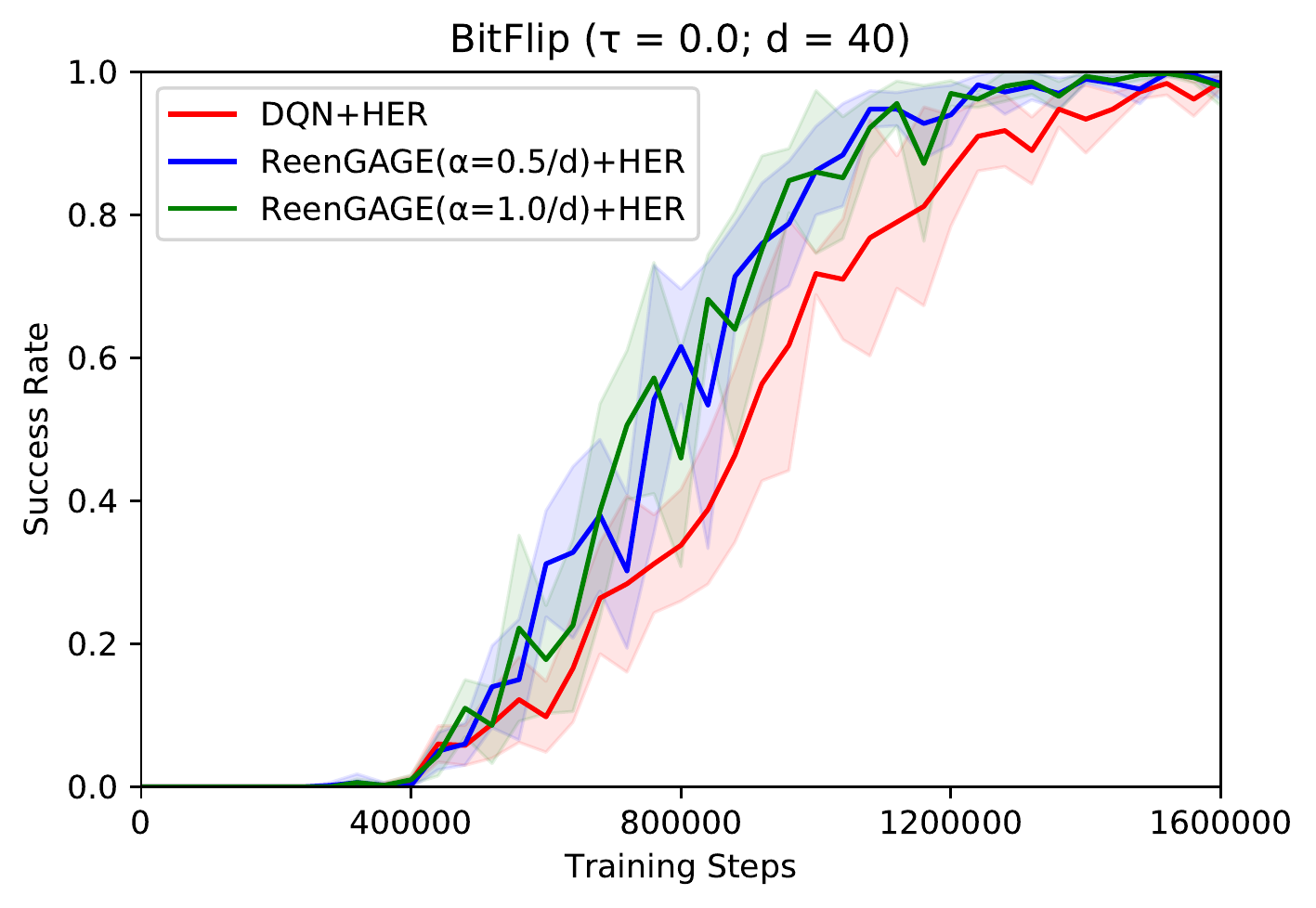}
     \end{subfigure}
     \hfill
     \begin{subfigure}[b]{0.44\textwidth}
         \centering
         \includegraphics[width=\textwidth]{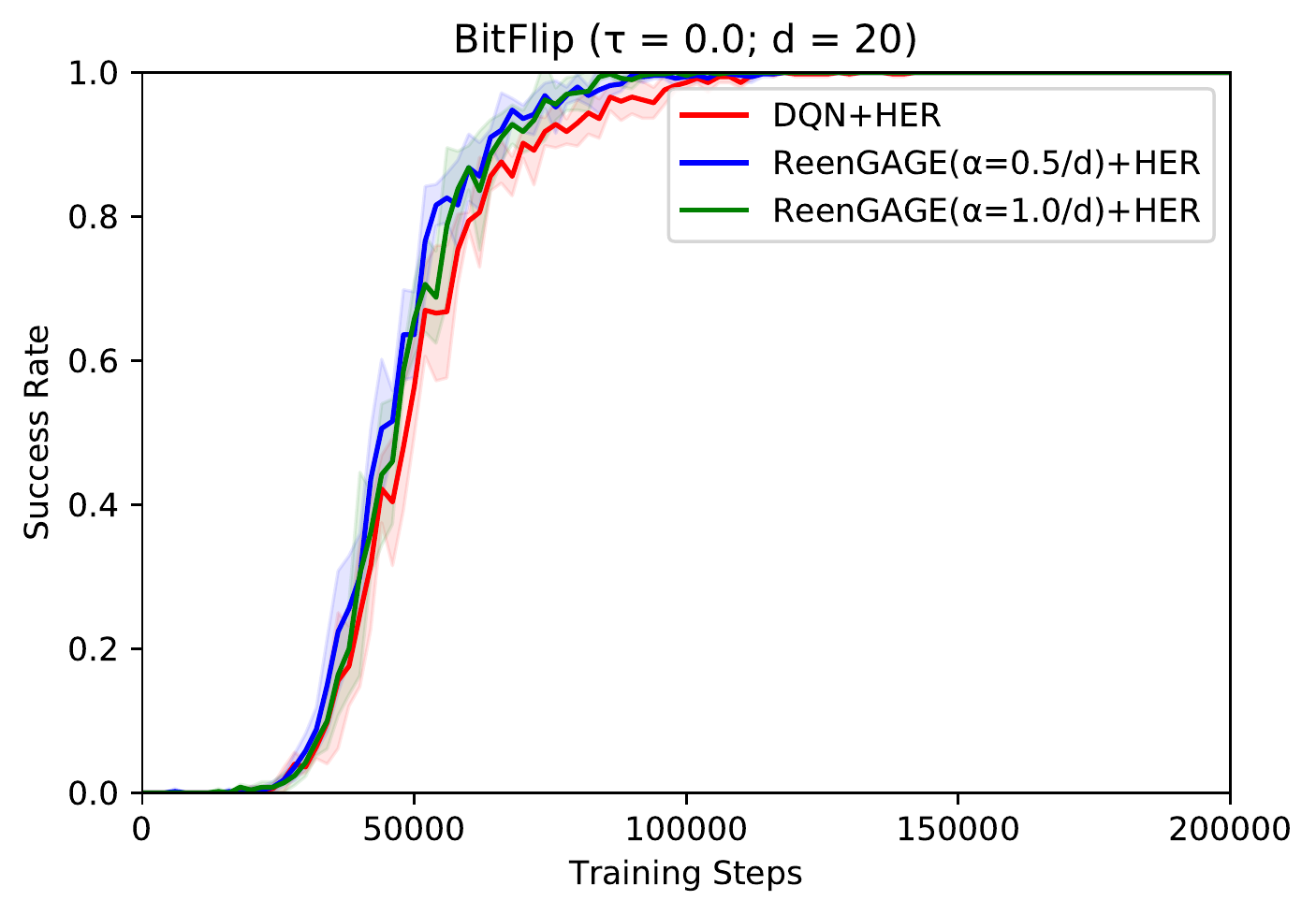}
     \end{subfigure}
     \hfill
     \begin{subfigure}[b]{0.44\textwidth}
         \centering
         \includegraphics[width=\textwidth]{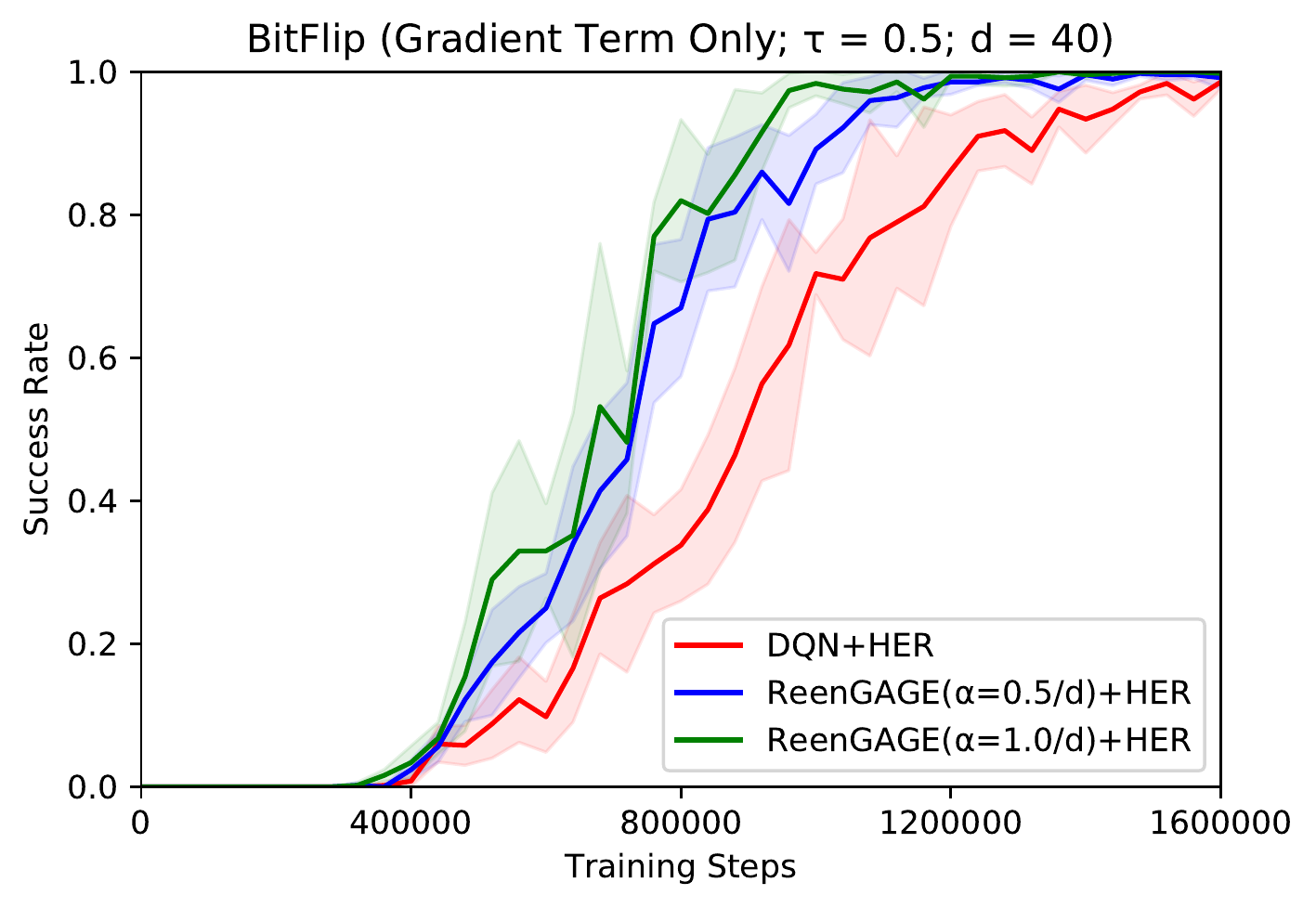}
     \end{subfigure}
     \hfill
     \begin{subfigure}[b]{0.44\textwidth}
         \centering
         \includegraphics[width=\textwidth]{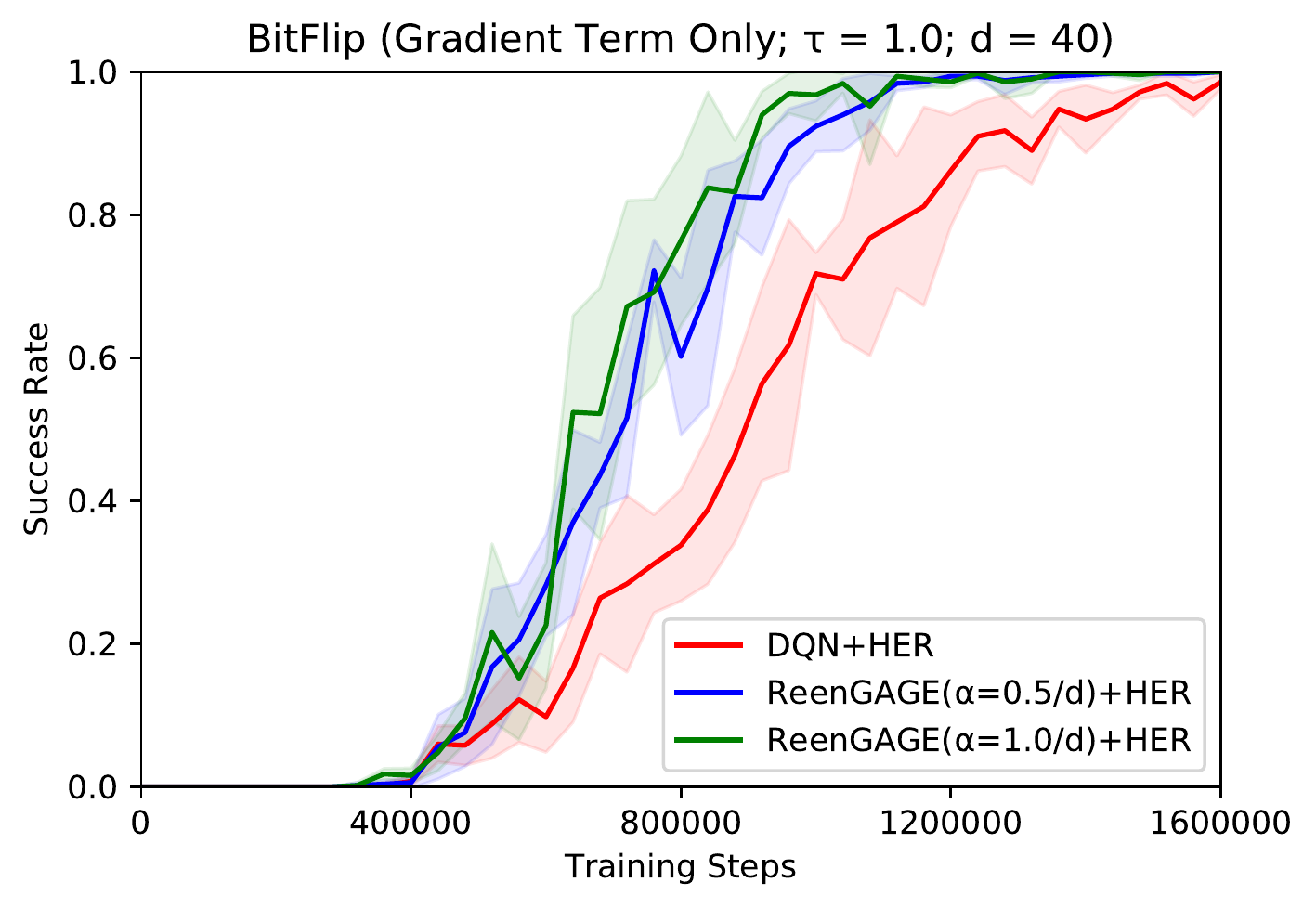}
     \end{subfigure}
     \hfill
     \begin{subfigure}[b]{0.44\textwidth}
         \centering
         \includegraphics[width=\textwidth]{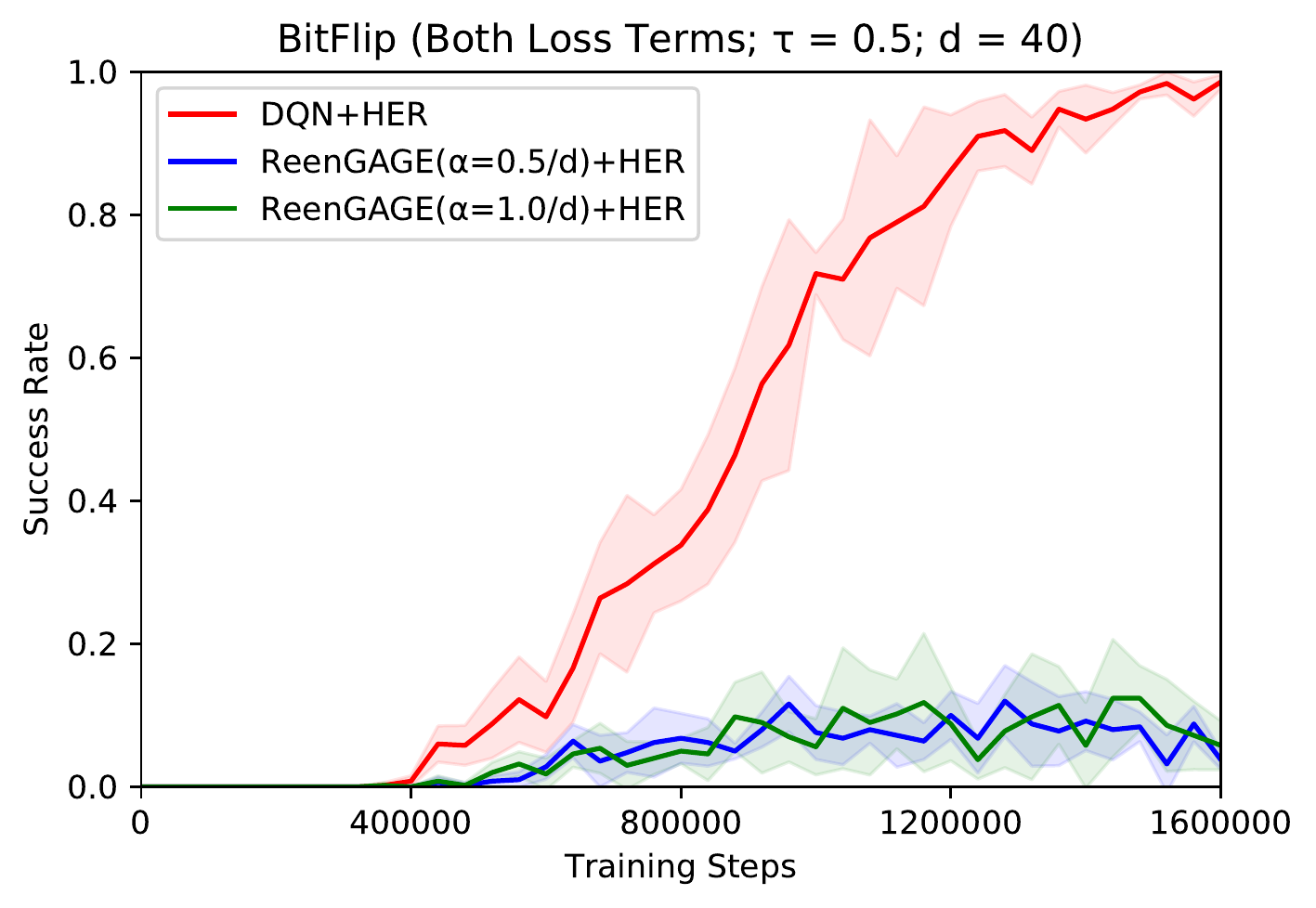}
     \end{subfigure}
     \hfill
     \begin{subfigure}[b]{0.44\textwidth}
         \centering
         \includegraphics[width=\textwidth]{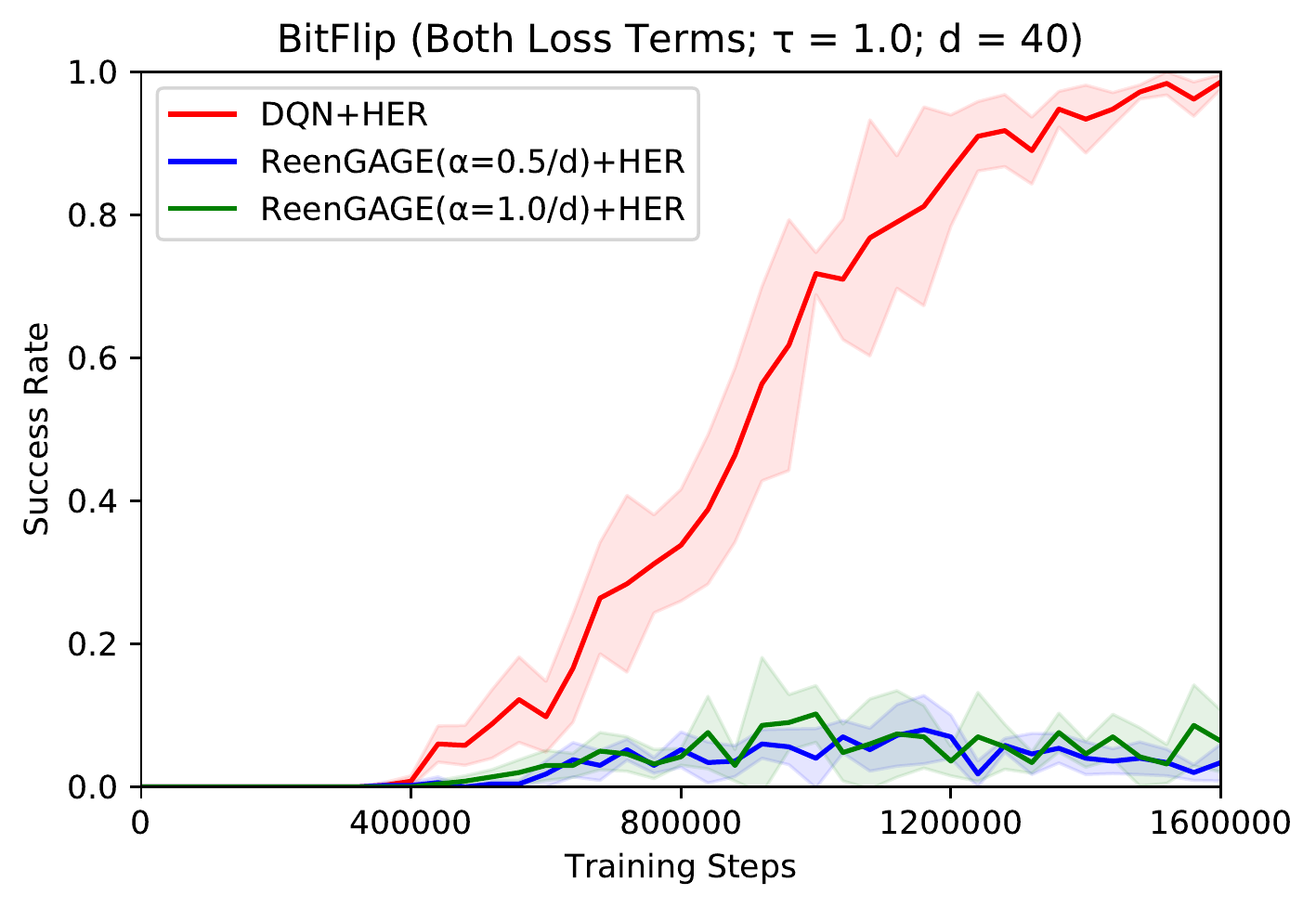}
     \end{subfigure}
     \hfill
     \begin{subfigure}[b]{0.44\textwidth}
         \centering
         \includegraphics[width=\textwidth]{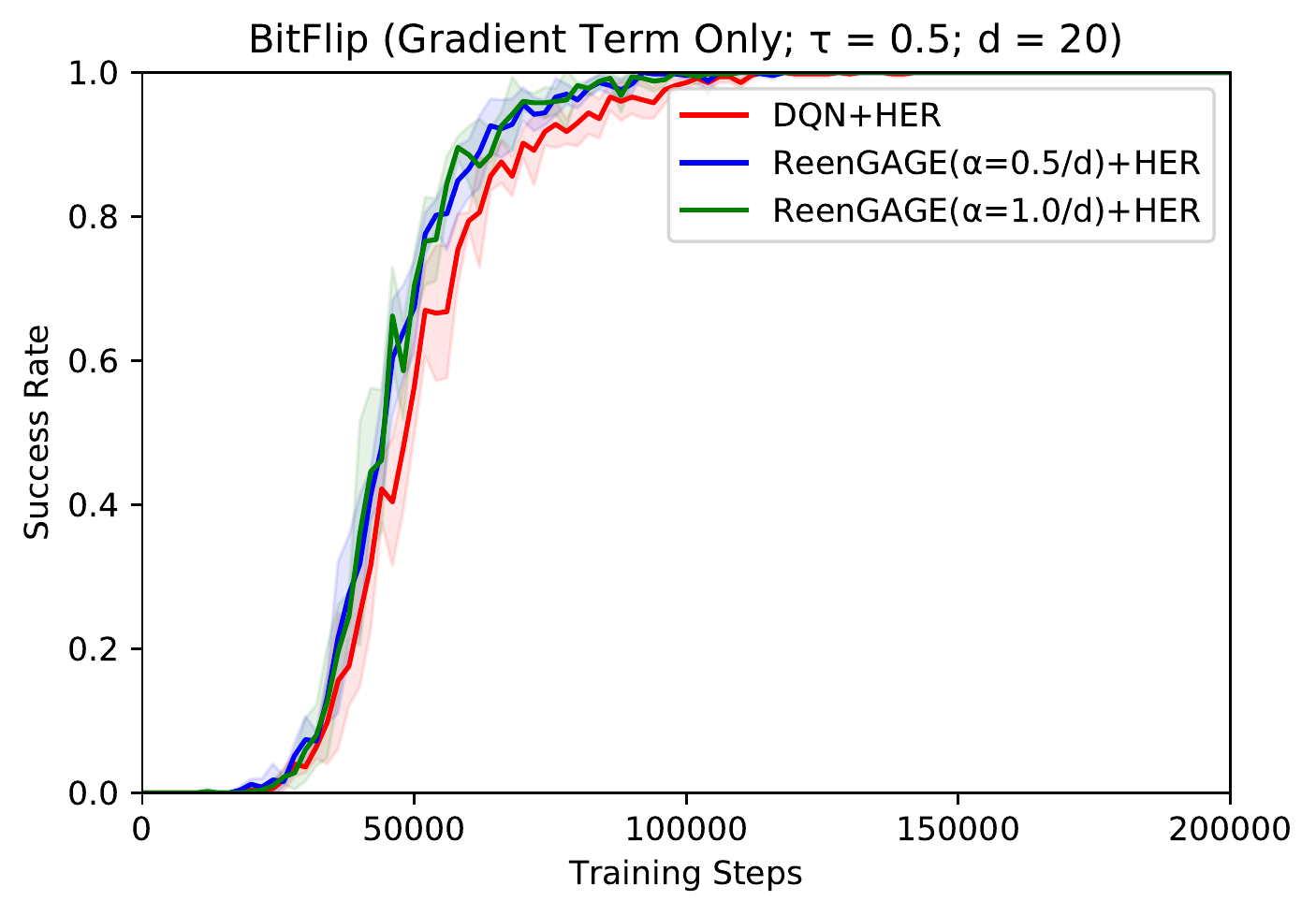}
     \end{subfigure}
     \hfill
     \begin{subfigure}[b]{0.44\textwidth}
         \centering
         \includegraphics[width=\textwidth]{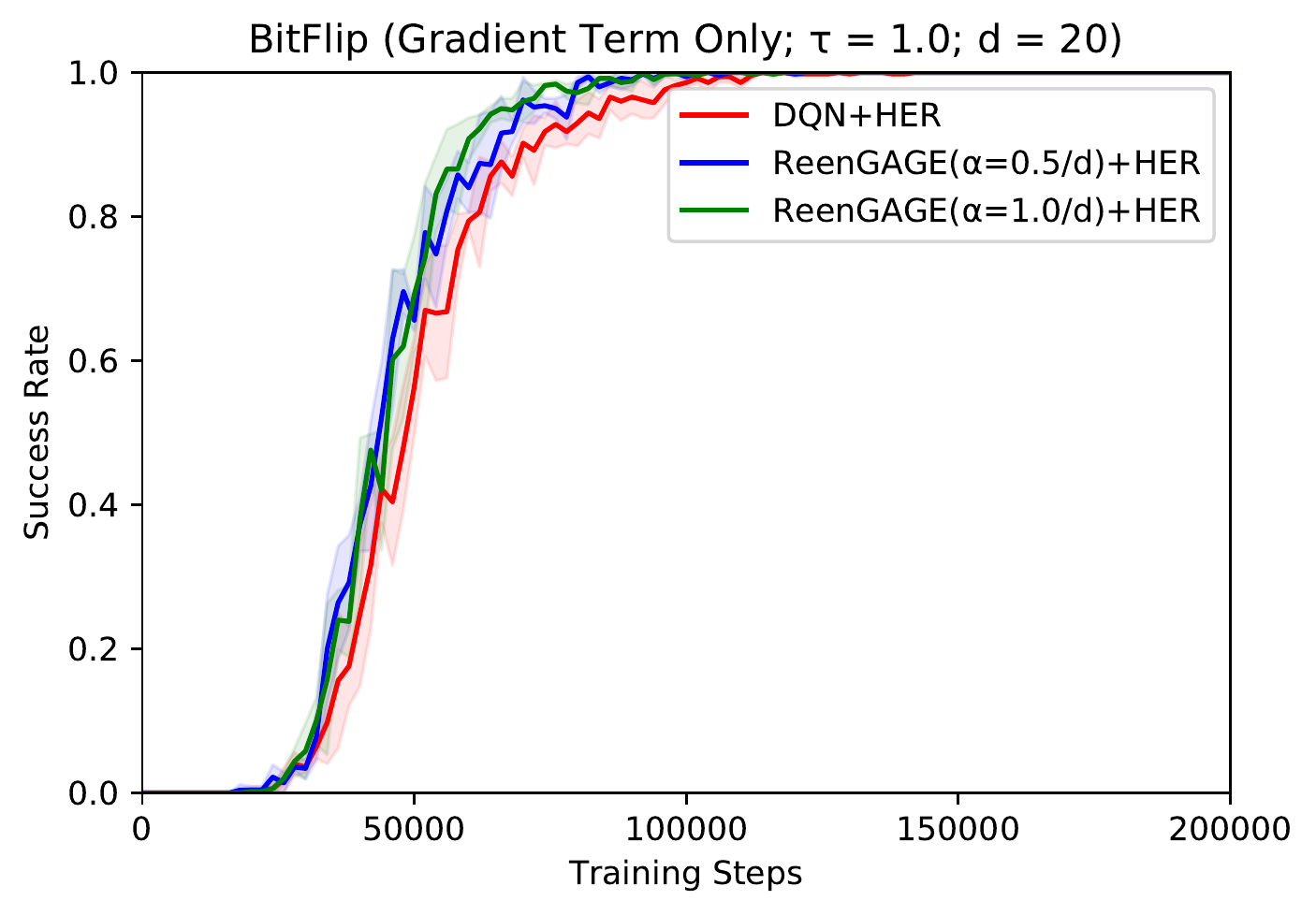}
     \end{subfigure}
        \caption{Results for the Bit-Flipping environment.}
        \label{fig:bit_flip}
\end{figure*}

Note that Equation \ref{eq:method_dense} requires that the gradient:
\begin{equation}
  \nabla_g Q_{\theta'}  (s',\pi_{\phi'}(s', g), g)
\end{equation}
be computable. This means that $\pi_{\phi'}(s', g)$ must be continuous and differentiable, which implies a continuous action space. To extend our method to discrete action spaces, we instead consider the target Q-value of DQN \cite{mnih2015human}, the standard  baseline method for discrete Q-learning, in a goal-conditioned setting:
\begin{equation}
  R(s',g)  + \gamma \, \max_a Q_{\theta',a}  (s', g), \label{eq:discrete_target}
\end{equation}
where $Q_{\theta} \in S \times G  \to \mathbb{R}^{|A|}$. This is not differentiable everywhere with respect to $g$: in particular, at points where, for some pair of actions $a',a''$, we have:
\begin{equation}
   \max_a Q_{\theta',a}  (s', g) =   Q_{\theta',a'}  (s', g) =   Q_{\theta',a''}  (s', g), 
\end{equation}
the gradient with respect to $g$ is not necessarily defined. In practice, this means that naively using auto-differentiation to take the gradient of Equation \ref{eq:discrete_target} produces the gradient of the target with respect to the goal \textit{assuming the optimal action remains constant}. To overcome this, we consider instead using a \textit{soft target}:
\begin{equation}
\begin{split}
         \text{SoftQTarget}(s',g):=& R(s',g) + \gamma\, \text{SoftMax}  [ Q_{\theta'}  (s', g)/\tau] \cdot   Q_{\theta'}  (s', g) \\
         =&R(s',g)  + \gamma \frac{ \sum_{a\in A} Q_{\theta',a}  (s', g) e^{Q_{\theta',a}  (s', g)/\tau}}{\sum_{a'\in A}  e^{Q_{\theta',a'}  (s', g)/\tau}} ,
\end{split}
\end{equation}
Where $\tau$ is the temperature hyperparameter. (Note that this approaches the standard DQN target as $\tau \rightarrow  0$.) We tested three uses of this soft target:

\begin{itemize}
    \item Soft target for \textit{gradient loss only}:
    \begin{equation}
\begin{split}
        \mathcal{L} =  \mathop\mathbb{E}_{(s,a,s',g)\sim \text{Buffer}} \Bigg[ &\mathcal{L}_{\text{Huber}} \Big[Q_{\theta,a}(s,g),
        R(s',g)+  \gamma \, \max_a Q_{\theta',a}  (s', g)\Big] \\
        + \alpha  &\mathcal{L_{\text{mse}}} \Big[\nabla_g Q_{\theta,a}(s,g), 
        \gamma \nabla_g \text{SoftQTarget}(s',g)\Big]\Bigg]  
\end{split}
\end{equation}
Following \cite{mnih2015human}, we use the Huber loss for the main loss term, although we use MSE for the gradient loss.
    \item Soft target for \textit{both loss terms}:
    \begin{equation}
\begin{split}
        \mathcal{L} =  \mathop\mathbb{E}_{(s,a,s',g)\sim \text{Buffer}} \Bigg[ &\mathcal{L}_{\text{Huber}} \Big[Q_{\theta,a}(s,g),
        \text{SoftQTarget}(s',g)\Big] \\
        + \alpha  &\mathcal{L_{\text{mse}}} \Big[\nabla_g Q_{\theta,a}(s,g), 
        \gamma \nabla_g \text{SoftQTarget}(s',g)\Big]\Bigg]  
\end{split}
\end{equation}
    \item Soft target for both loss terms, \textit{and} take actions nondeterministically, with a probability distribution given by $\text{SoftMax}  [ Q_{\theta'}  (s', g)/\tau]$.
\end{itemize}
We test on an implementation of the ``Bit-Flipping'' sparse-reward environment from \cite{andrychowicz2017hindsight}, with dimensionality $d=40$ bits using DQN with HER as the baseline. Because code for this experiment is not provided by \cite{andrychowicz2017hindsight}, we re-implemented the experiment using the Stable-Baselines3 package \cite{JMLR:v22:20-1364}. We used the hyperparameters specified by \cite{andrychowicz2017hindsight}, with the following minor modifications (note that for some of these specifications, \cite{andrychowicz2017hindsight} is unclear about whether the specification was applied for the Bit-Flipping environment, or only in the continuous control, DDPG, environments tested in that work):
\begin{itemize}
    \item We train on a single GPU, rather than averaging gradient updates from 8 workers.
    \item While we use Adam optimizer with learning rate of 0.001 as specified, we use PyTorch defaults for other Adam hyperparameters, rather than TensorFlow defaults.
    \item We do not clip the target Q-values (this feature is not part of the DQN implementation of \cite{JMLR:v22:20-1364}).
    \item We evaluate using the current, rather than target, Q-value function.
    \item We do not normalize observations (which are already $\{0,1\}$).
\end{itemize}
The baselines we present are from our re-implementation, to provide a fair comparison. For our method, we performed a grid search on $\alpha \in \{0.5,1.0\}$ and $\tau \in \{0.0,0.5,1.0\}$. Results are presented in Figure \ref{fig:bit_flip}. We observed that the ``gradient loss only'' method was effective at improving performance, both over standard DQN and over ReenGAGE with standard DQN targets $(\tau=0)$. By contrast, using soft targets for both loss terms made the performance worse, and using nondeterministic actions with soft targets for both loss terms brought the success rate to zero (and hence is not shown). We tested all three methods with $d=40$ bits, and then applied the successful method (``gradient loss only') to $d=20$ as well; improvements over baseline for $d=20$ were minor.

\begin{table}
    \centering
    \begin{tabular}{|c|c|}
    \hline
        Replay Buffer Size &  1000000\\
        Frequency of Training & Every 1 environment step \\
        Gradient Descent Steps per Training & 1 \\
        Initial Steps before Training & 1000\\
        Discount $\gamma$ & 0.95\\
        Polyak Update $\tau$ & 0.005 \\
        Normal action noise for training $\sigma$ & 0.03\\
        Architecture (both actor and critic) & Fully Connected; 2 hidden layers of width 256; ReLU activations \\
        HER number relabeled goals k & 4\\
        HER relabeling strategy & `Future'\\
        Evaluation episodes & 50\\
        Evaluation Frequency & Every 2000 environment steps\\
    \hline
    \end{tabular}
    \caption{Hyperparameters for ContinuousSeek}
    \label{tab:continuous_seek_hyperparams}
\end{table}
\section{Hyperparameters and Full Results for ContinuousSeek}
\begin{figure*}
     \centering
     \begin{subfigure}[b]{0.33\textwidth}
         \centering
         \includegraphics[width=\textwidth]{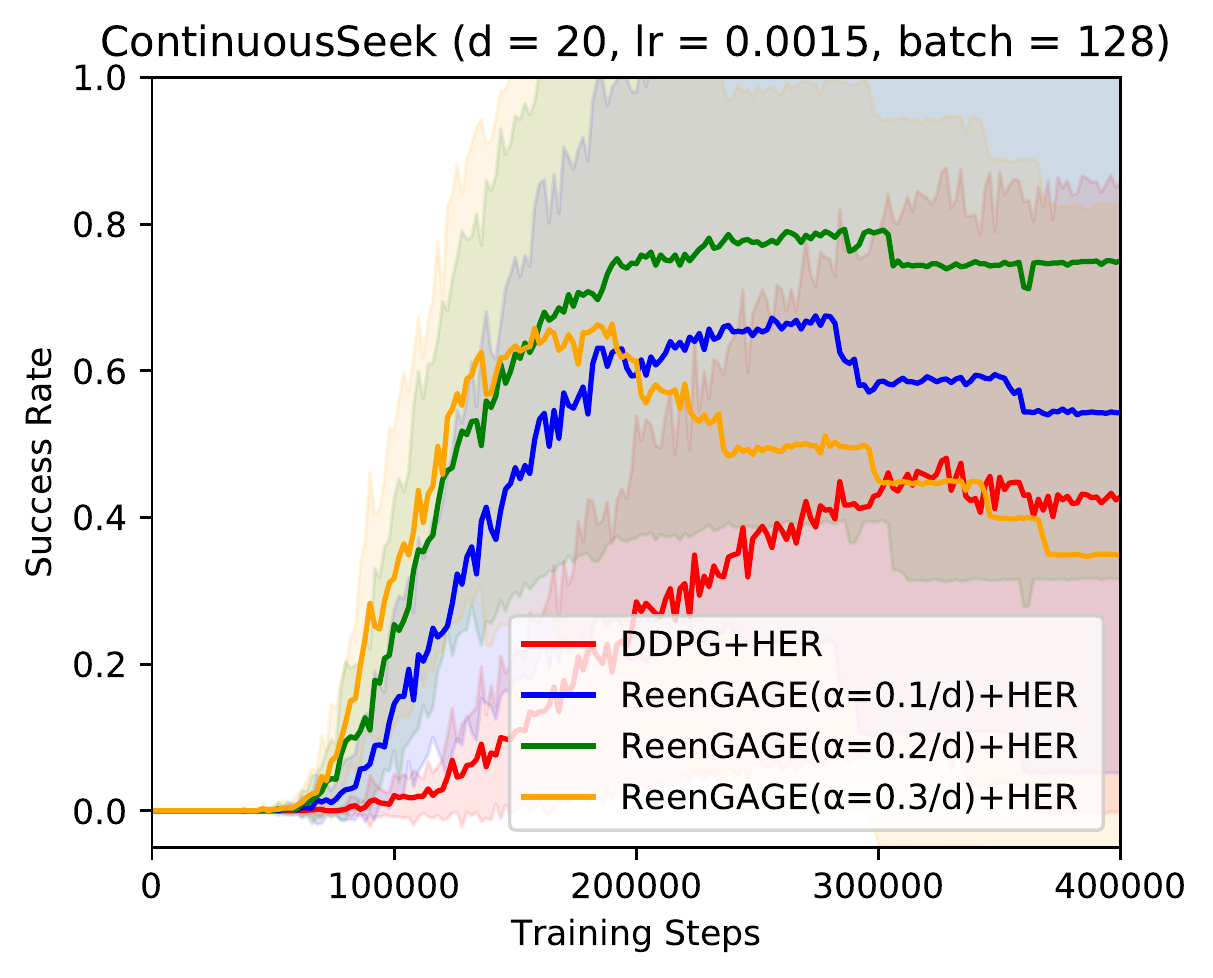}
     \end{subfigure}
     \hfill
     \begin{subfigure}[b]{0.33\textwidth}
         \centering
         \includegraphics[width=\textwidth]{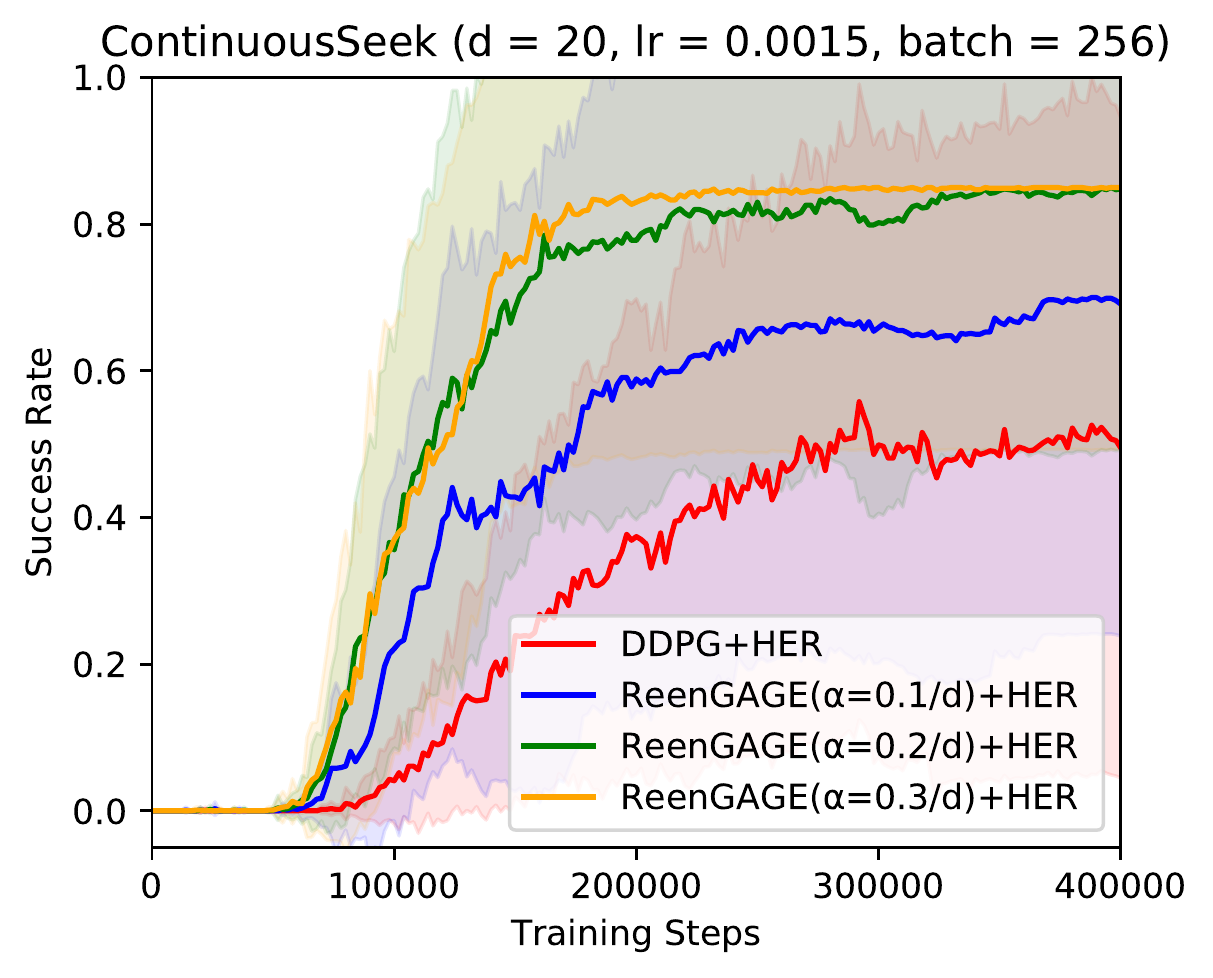}
     \end{subfigure}
     \hfill
     \begin{subfigure}[b]{0.33\textwidth}
         \centering
         \includegraphics[width=\textwidth]{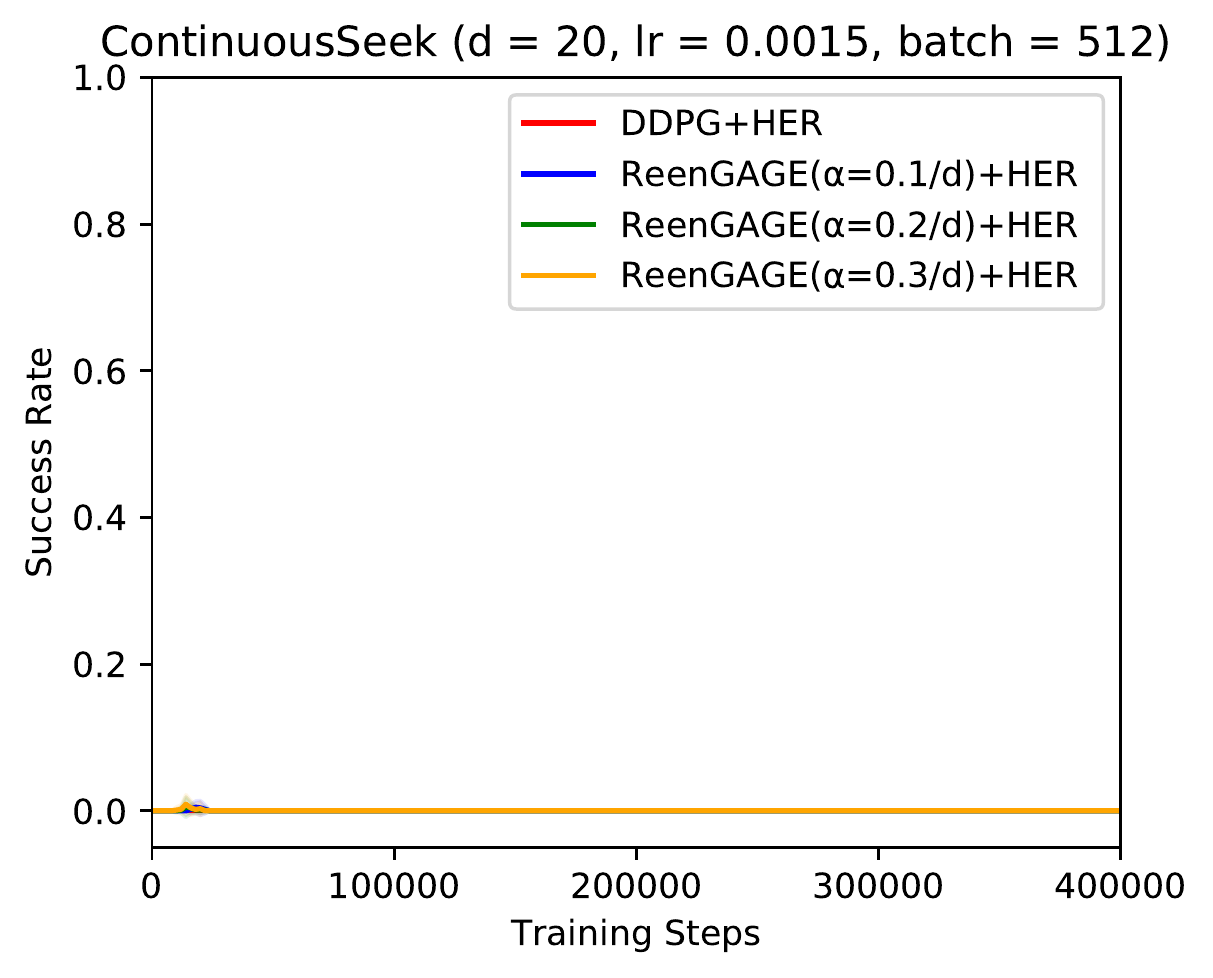}
     \end{subfigure}
     \begin{subfigure}[b]{0.33\textwidth}
         \centering
         \includegraphics[width=\textwidth]{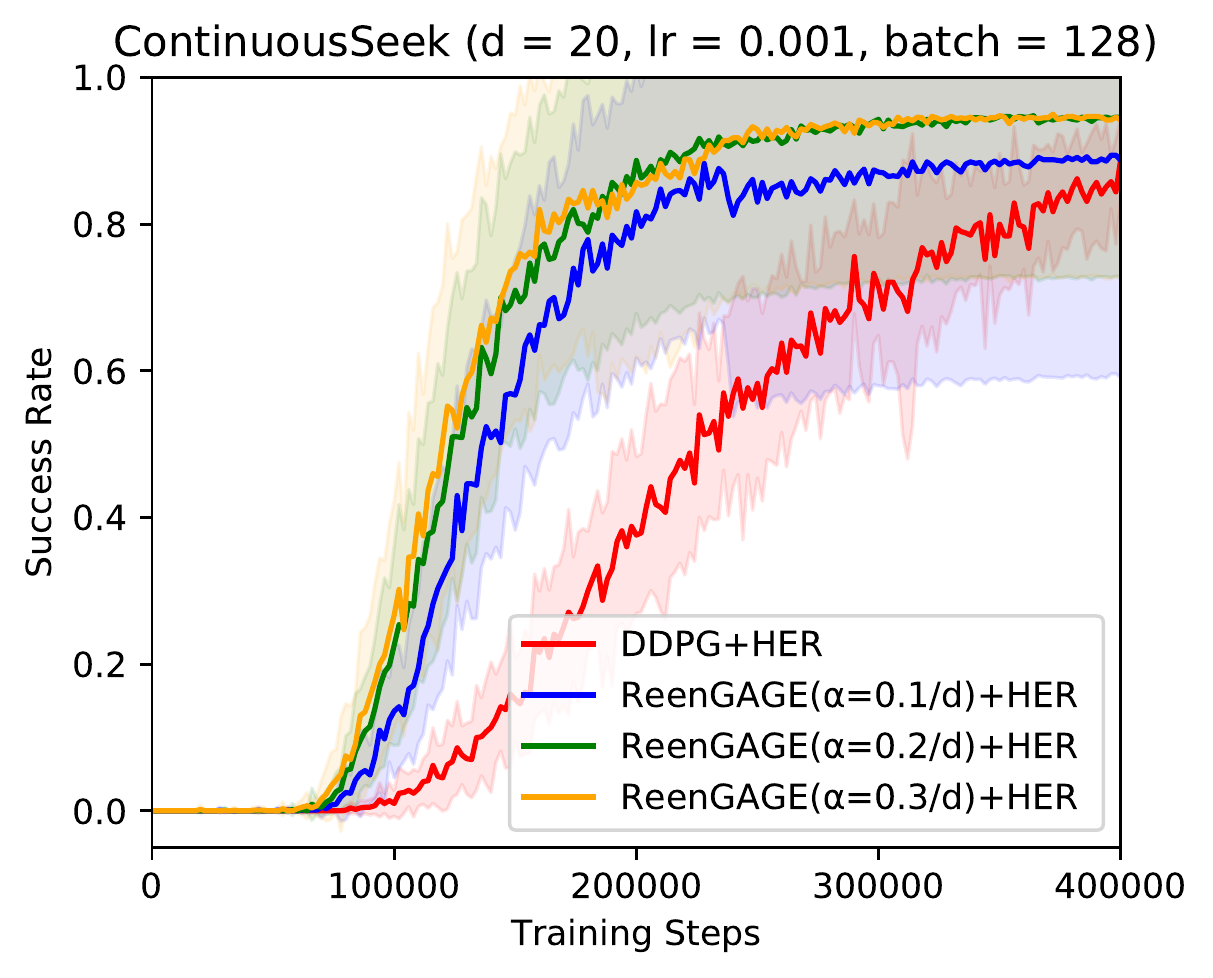}
     \end{subfigure}
     \hfill
     \begin{subfigure}[b]{0.33\textwidth}
         \centering
         \includegraphics[width=\textwidth]{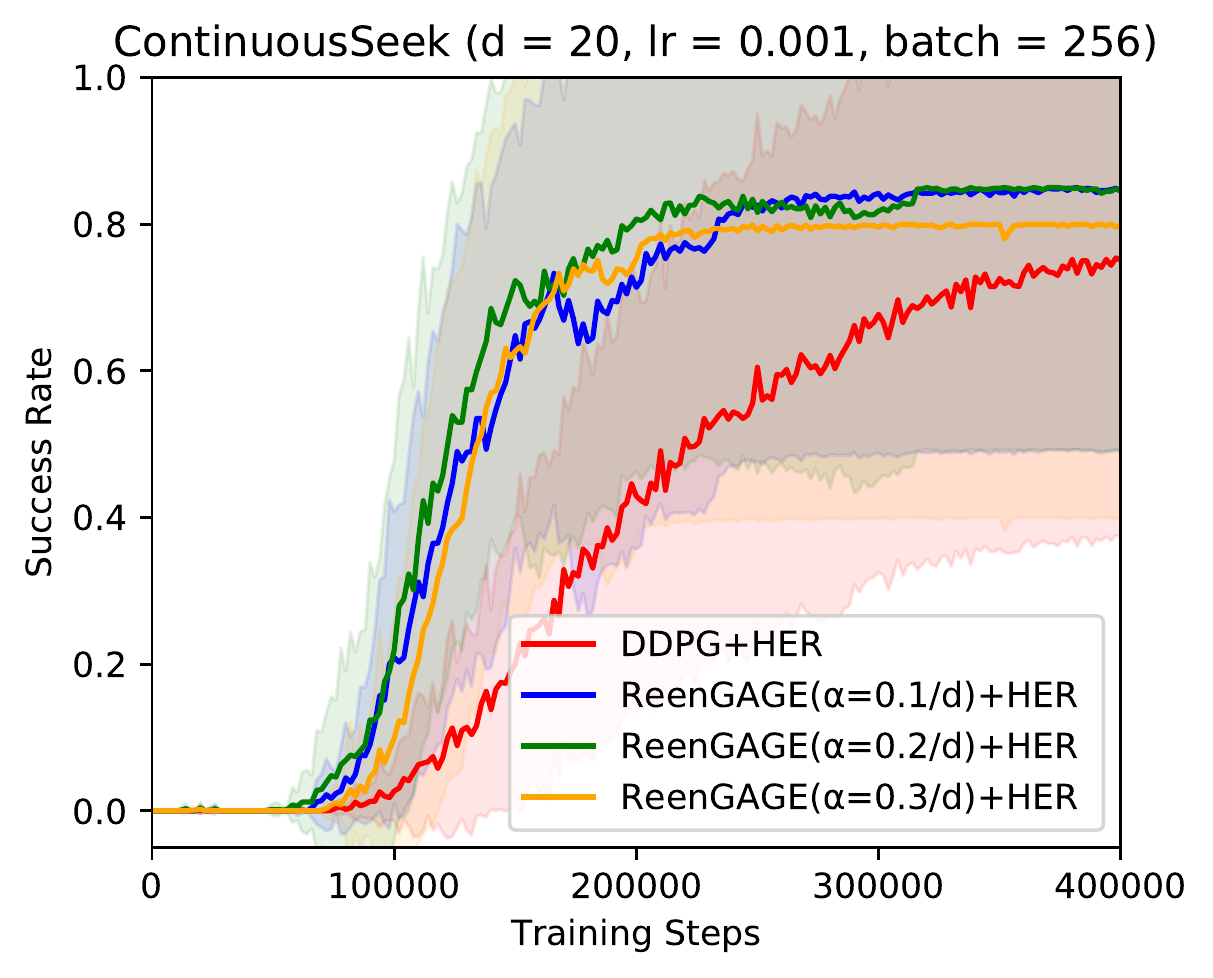}
     \end{subfigure}
     \hfill
     \begin{subfigure}[b]{0.33\textwidth}
         \centering
         \includegraphics[width=\textwidth]{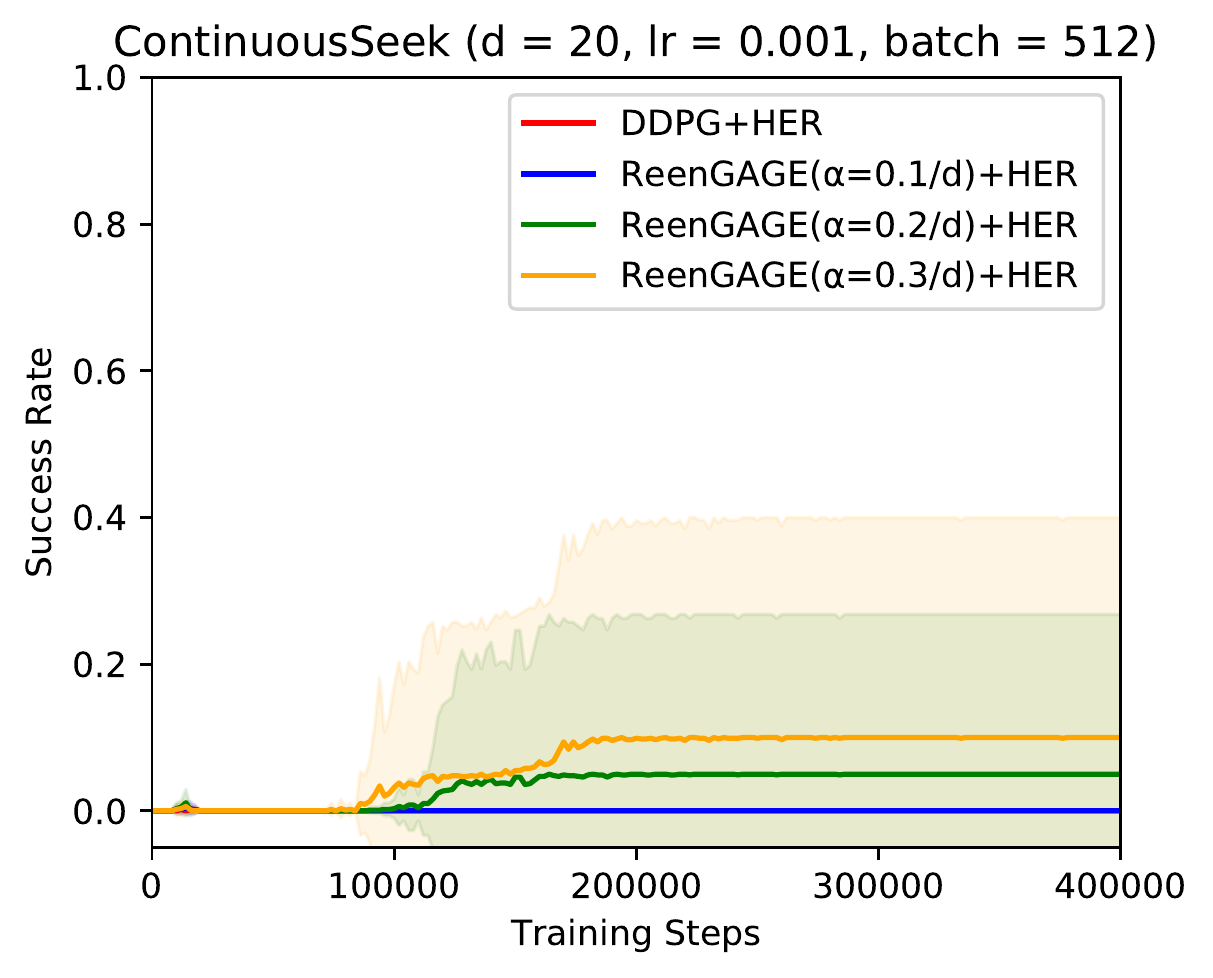}
     \end{subfigure}
     \begin{subfigure}[b]{0.33\textwidth}
         \centering
         \includegraphics[width=\textwidth]{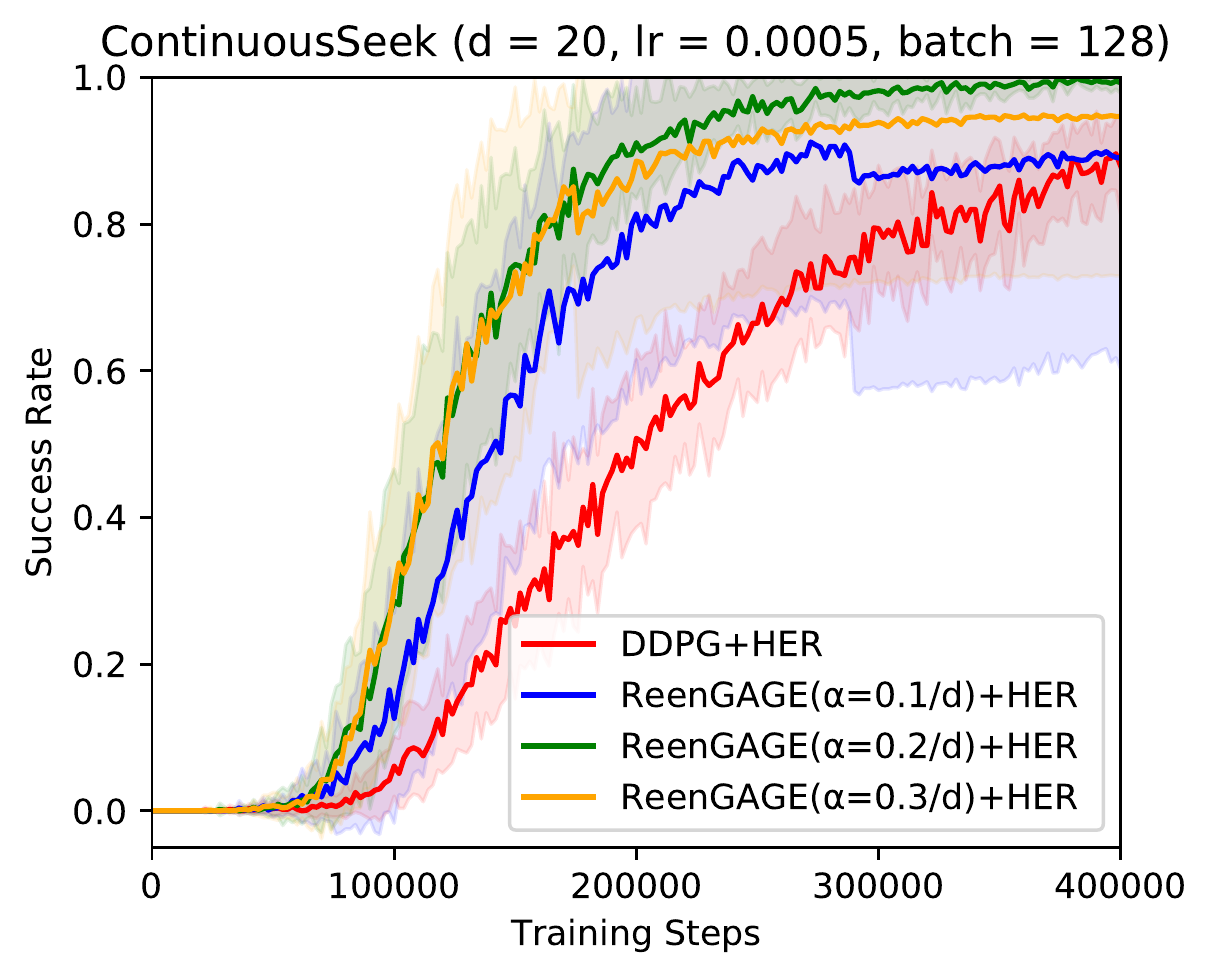}
     \end{subfigure}
     \hfill
     \begin{subfigure}[b]{0.33\textwidth}
         \centering
         \includegraphics[width=\textwidth]{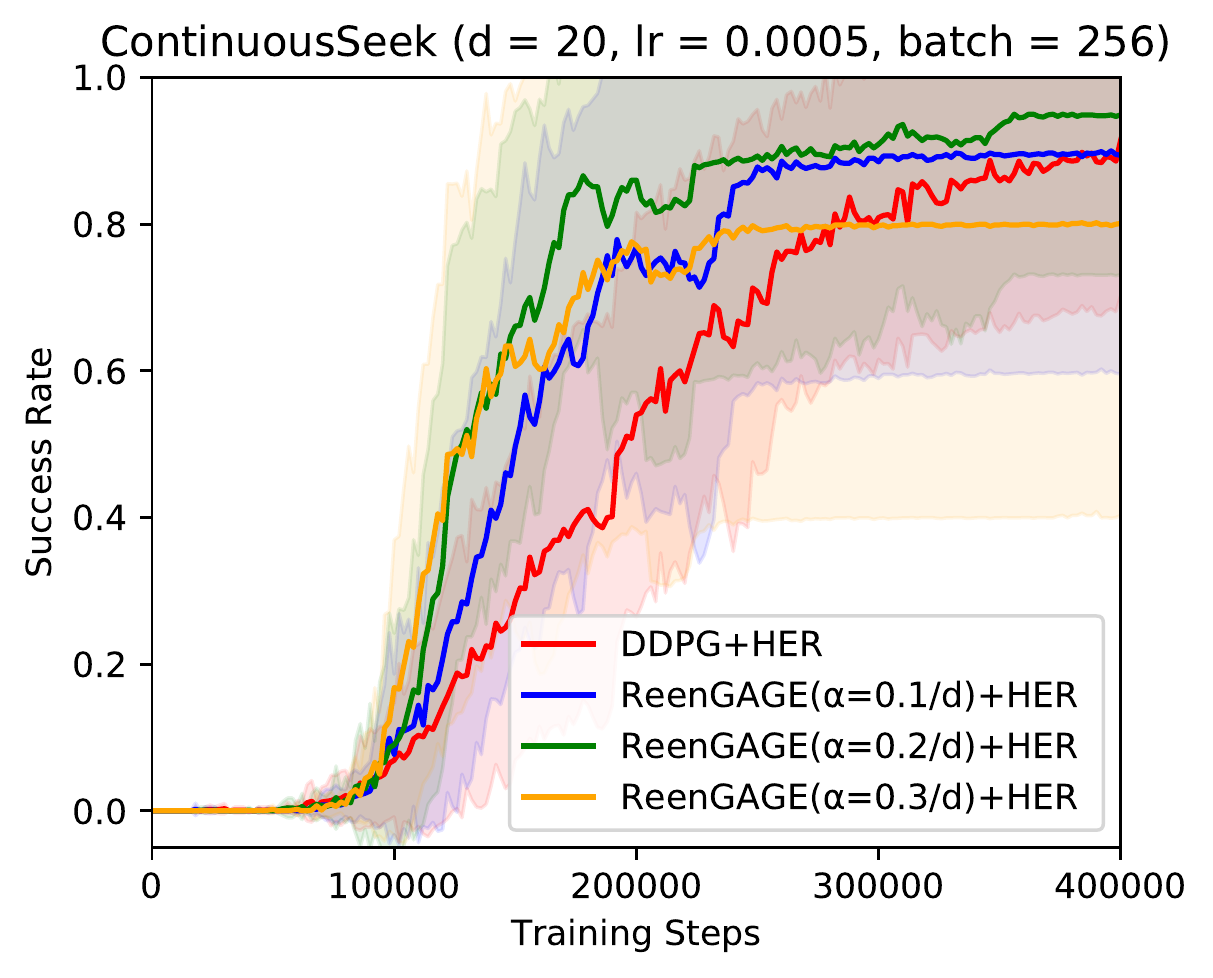}
     \end{subfigure}
     \hfill
     \begin{subfigure}[b]{0.33\textwidth}
         \centering
         \includegraphics[width=\textwidth]{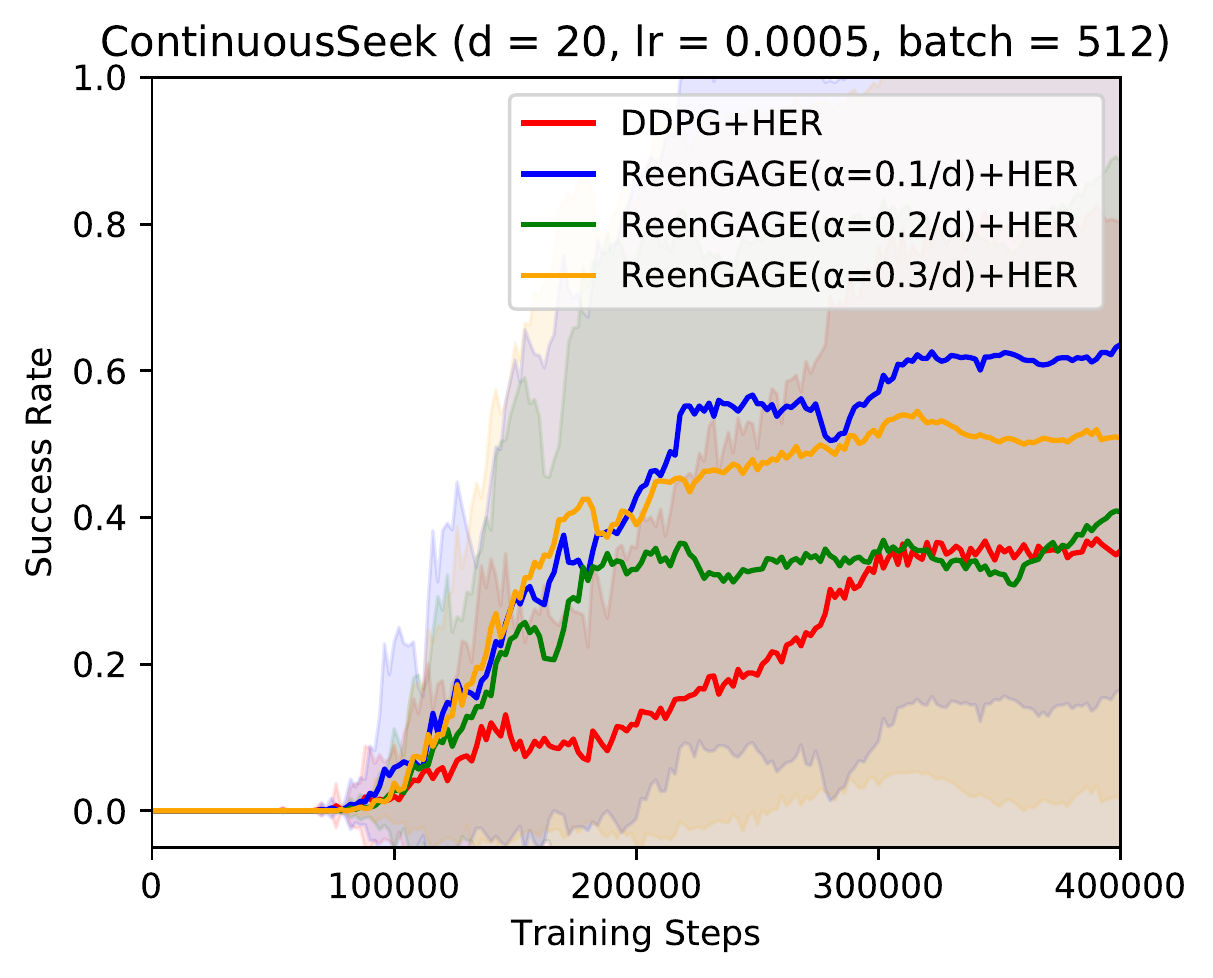}
     \end{subfigure}
     \begin{subfigure}[b]{0.33\textwidth}
         \centering
         \includegraphics[width=\textwidth]{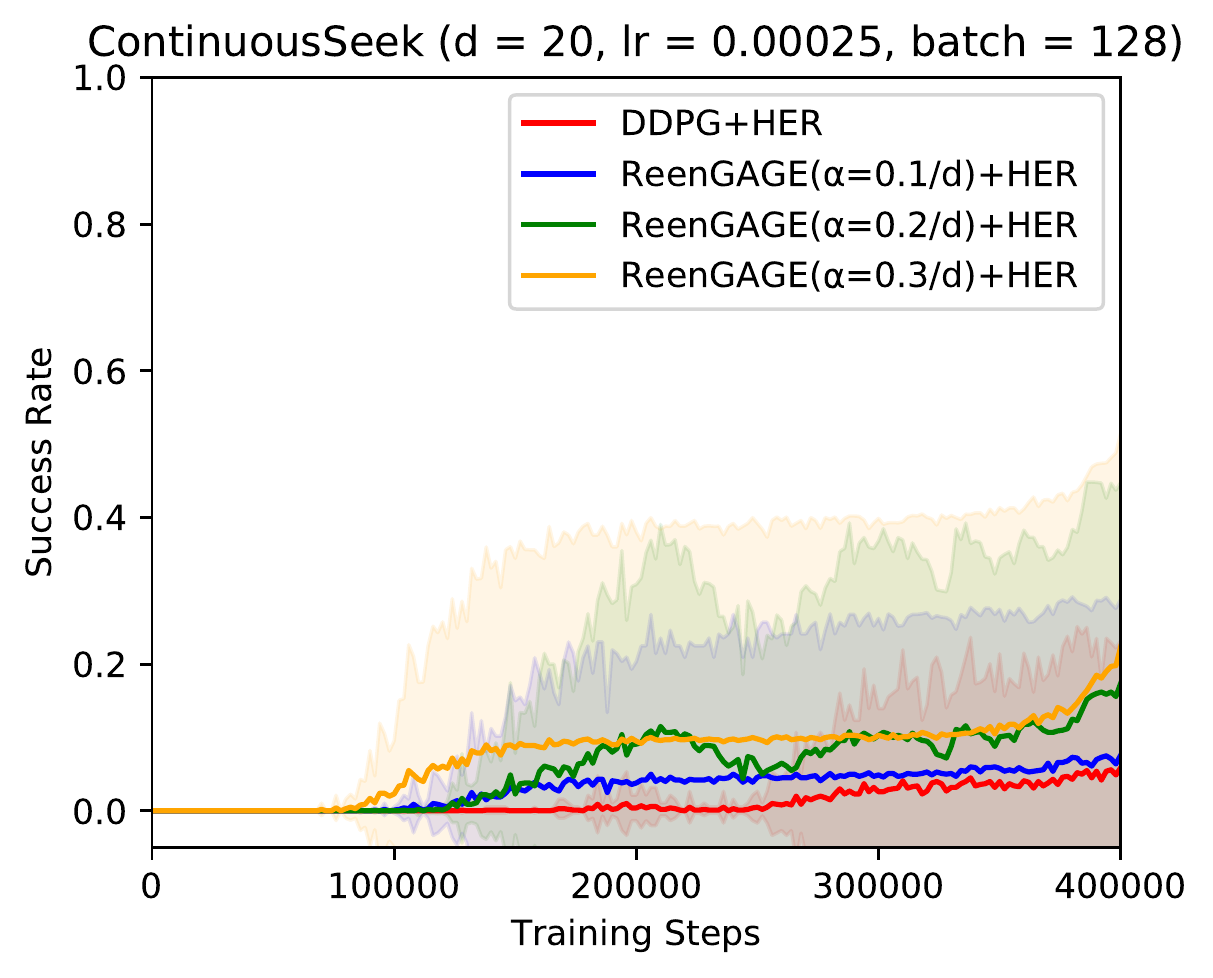}
     \end{subfigure}
     \hfill
     \begin{subfigure}[b]{0.33\textwidth}
         \centering
         \includegraphics[width=\textwidth]{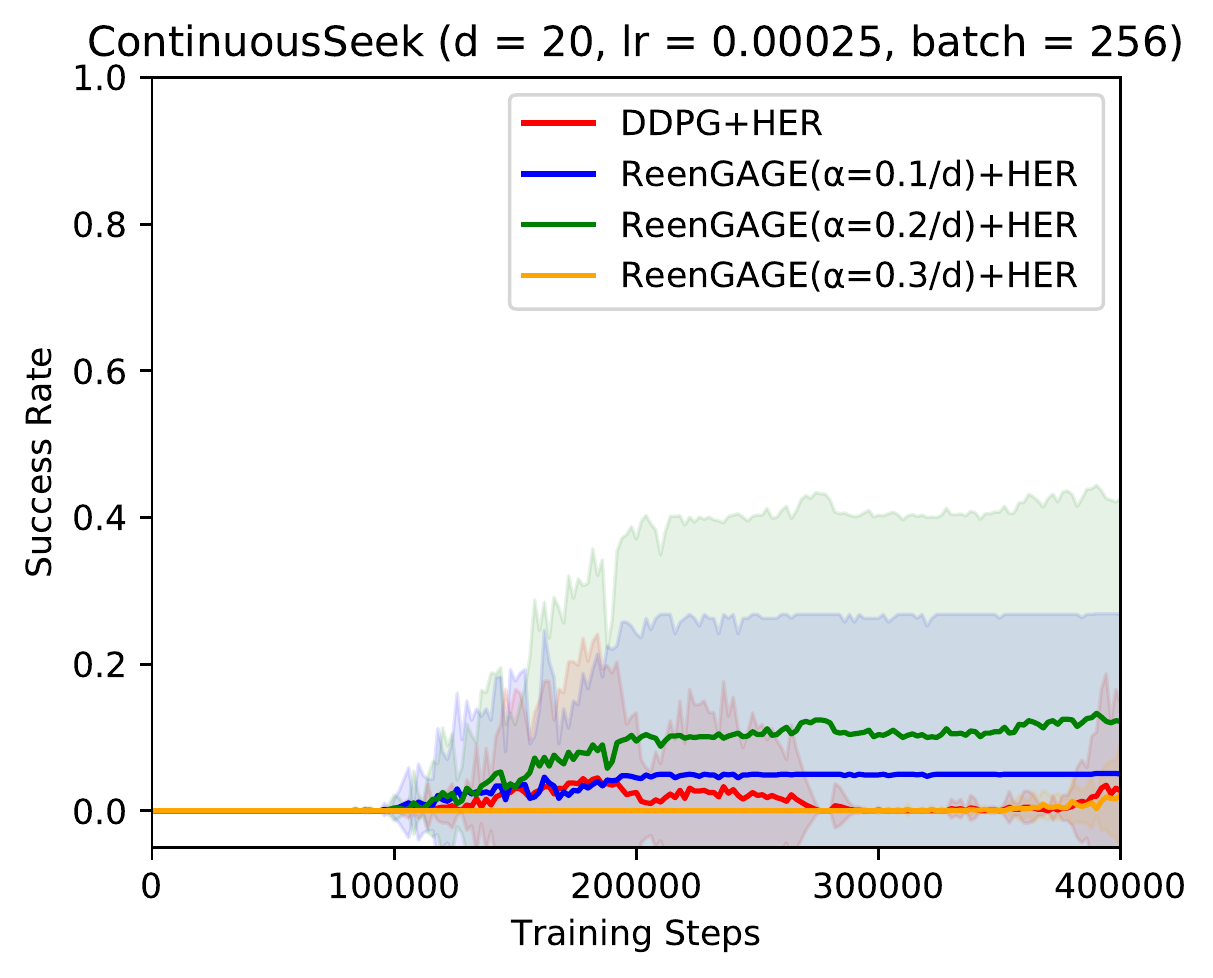}
     \end{subfigure}
     \hfill
     \begin{subfigure}[b]{0.33\textwidth}
         \centering
         \includegraphics[width=\textwidth]{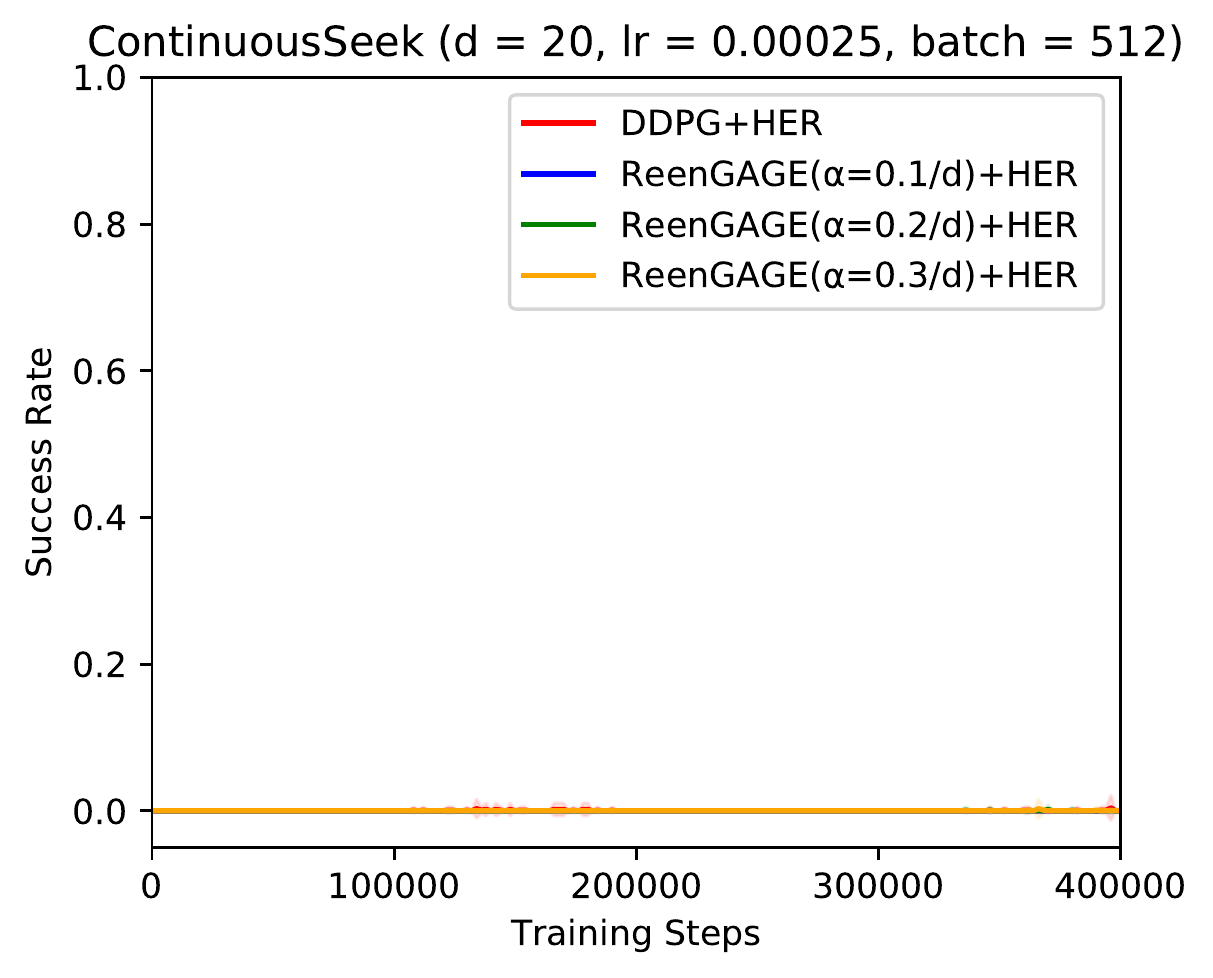}
     \end{subfigure}
        \caption{Complete ContinuousSeek Results for $d=20$.}
        \label{fig:continuous_seek_appdx_20}
\end{figure*}

\begin{figure*}
     \centering
     \begin{subfigure}[b]{0.33\textwidth}
         \centering
         \includegraphics[width=\textwidth]{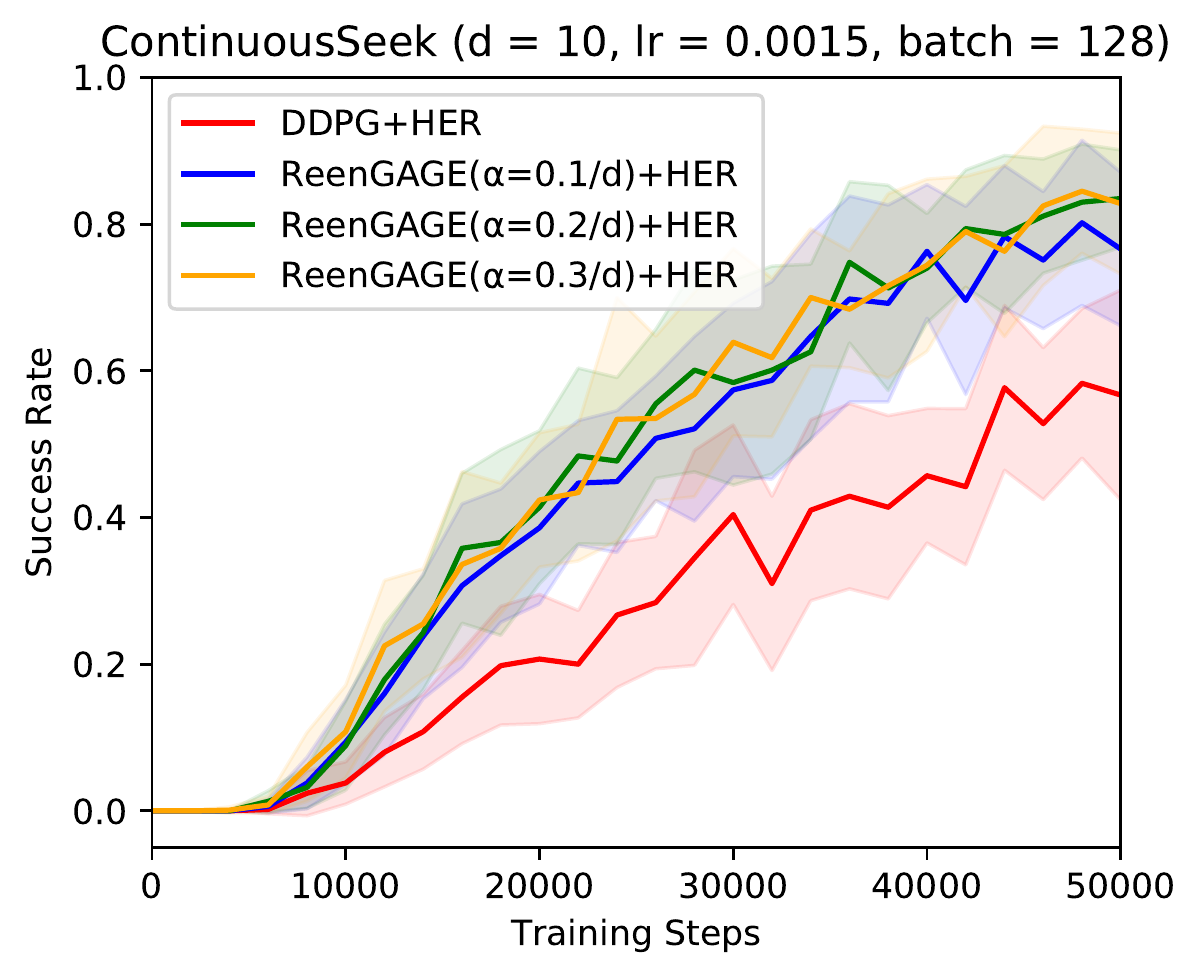}
     \end{subfigure}
     \hfill
     \begin{subfigure}[b]{0.33\textwidth}
         \centering
         \includegraphics[width=\textwidth]{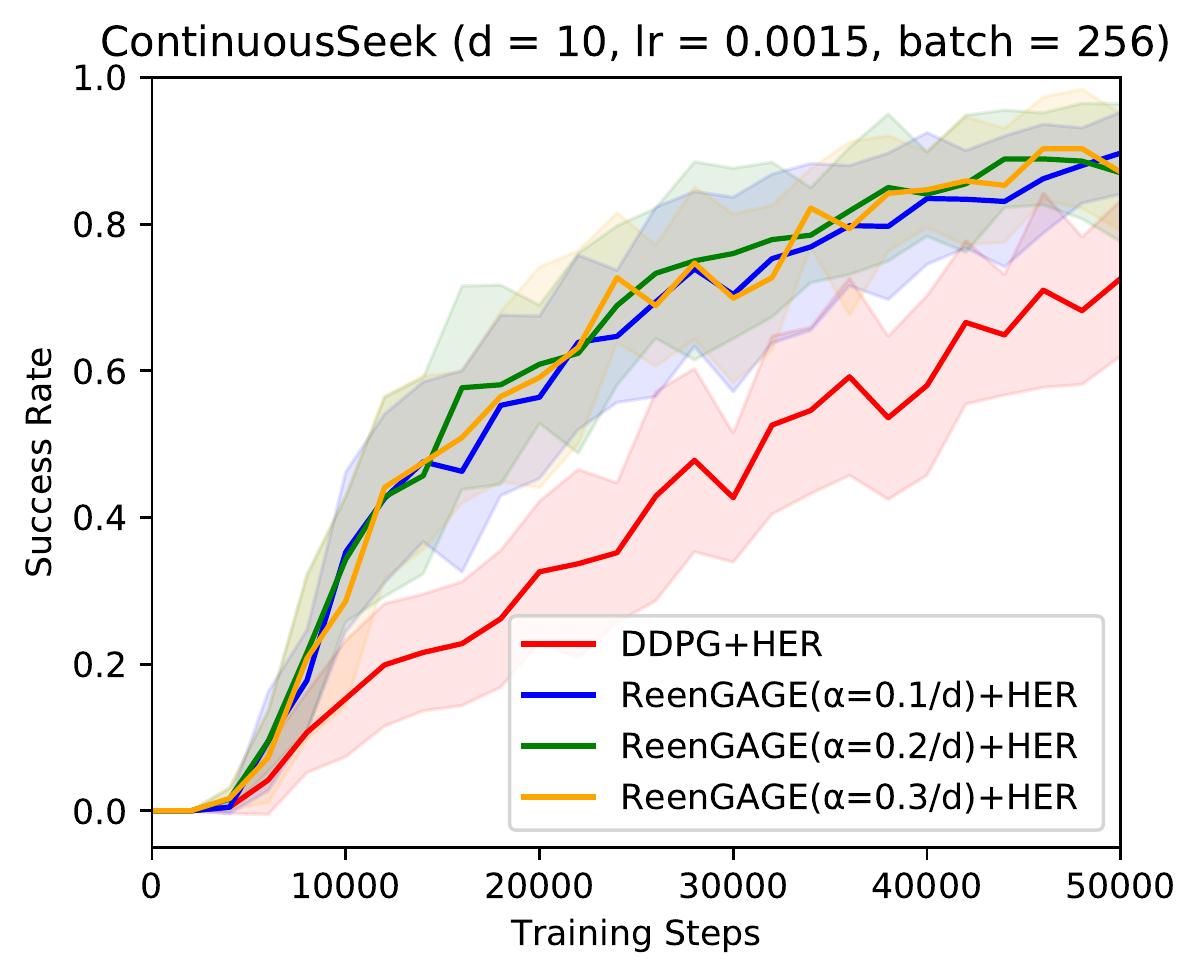}
     \end{subfigure}
     \hfill
     \begin{subfigure}[b]{0.33\textwidth}
         \centering
         \includegraphics[width=\textwidth]{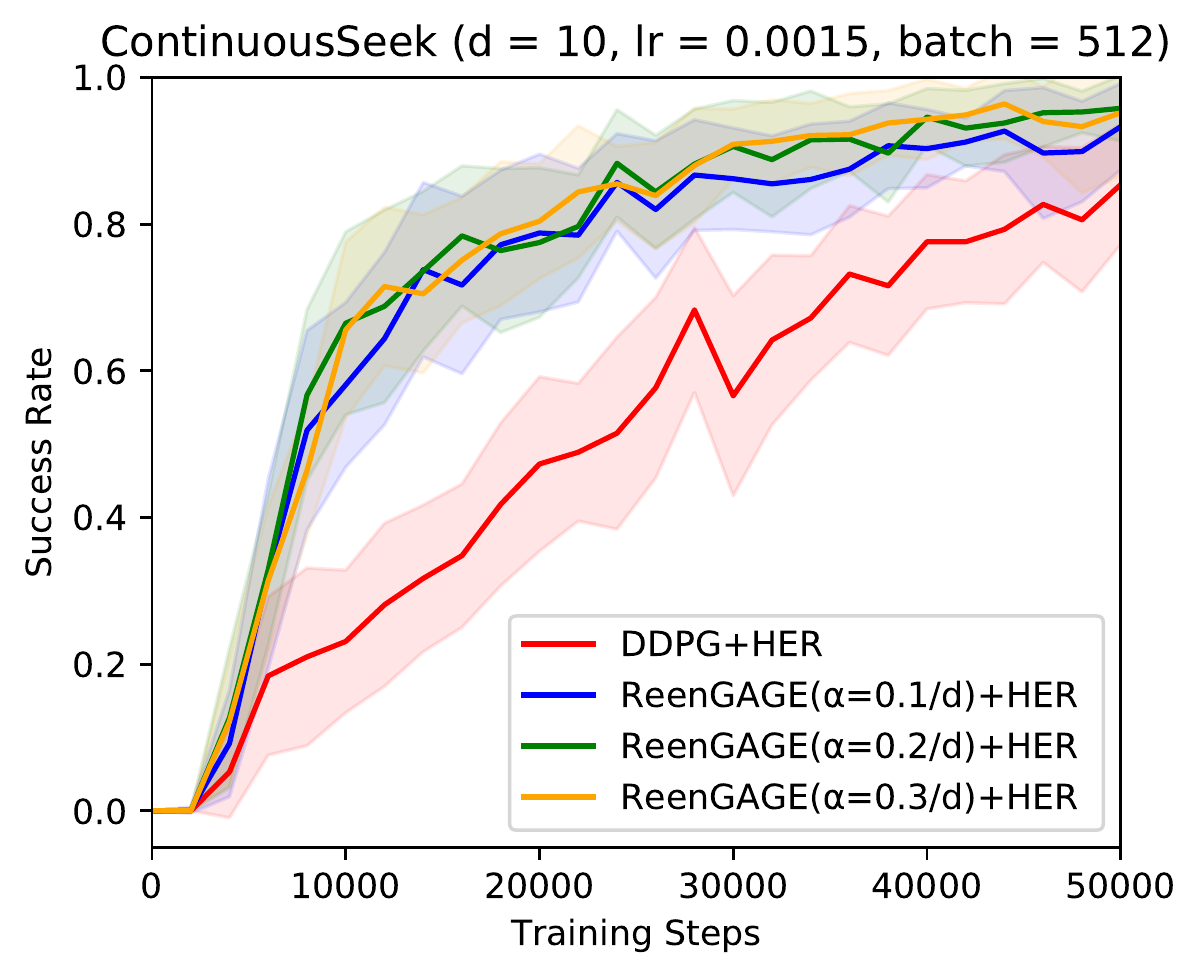}
     \end{subfigure}
     \begin{subfigure}[b]{0.33\textwidth}
         \centering
         \includegraphics[width=\textwidth]{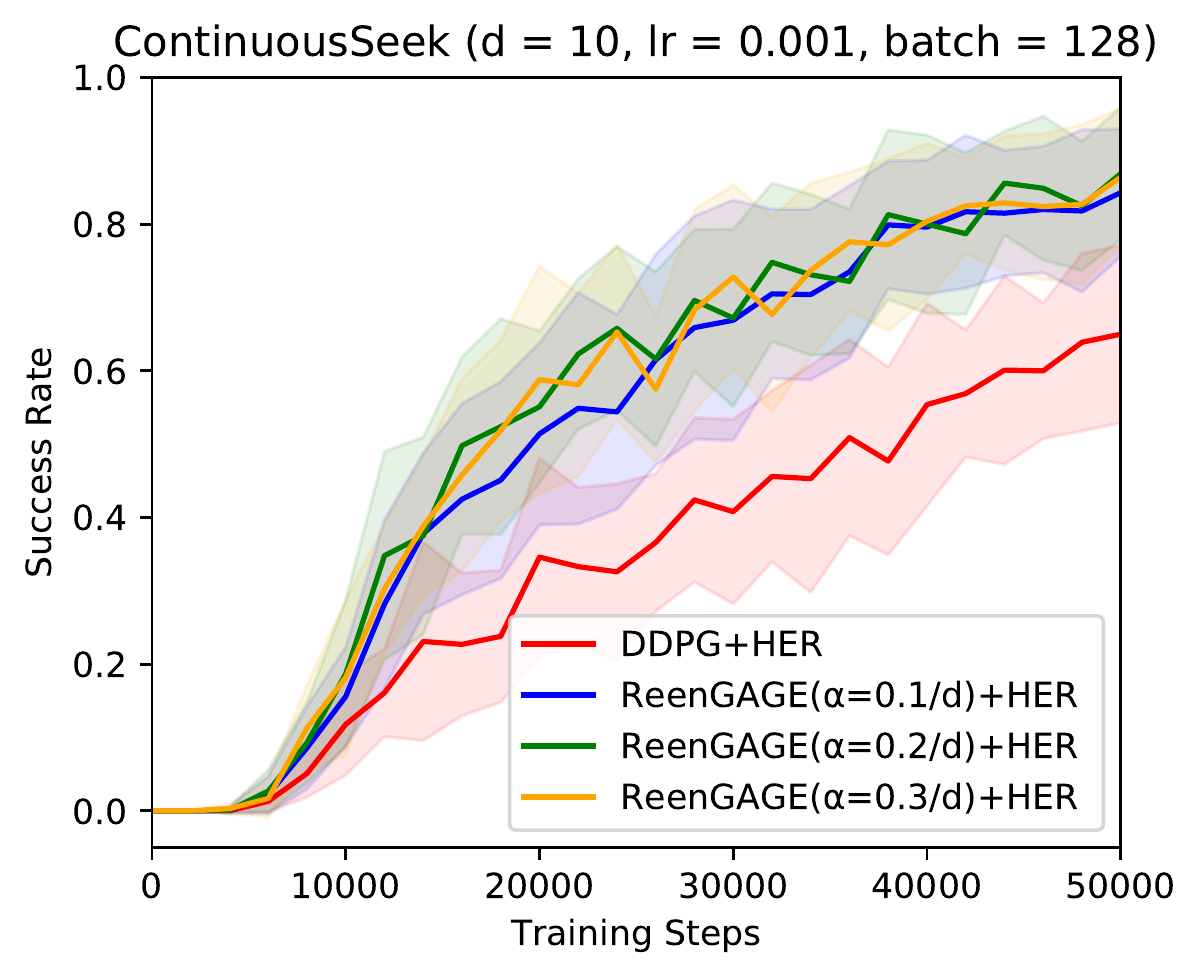}
     \end{subfigure}
     \hfill
     \begin{subfigure}[b]{0.33\textwidth}
         \centering
         \includegraphics[width=\textwidth]{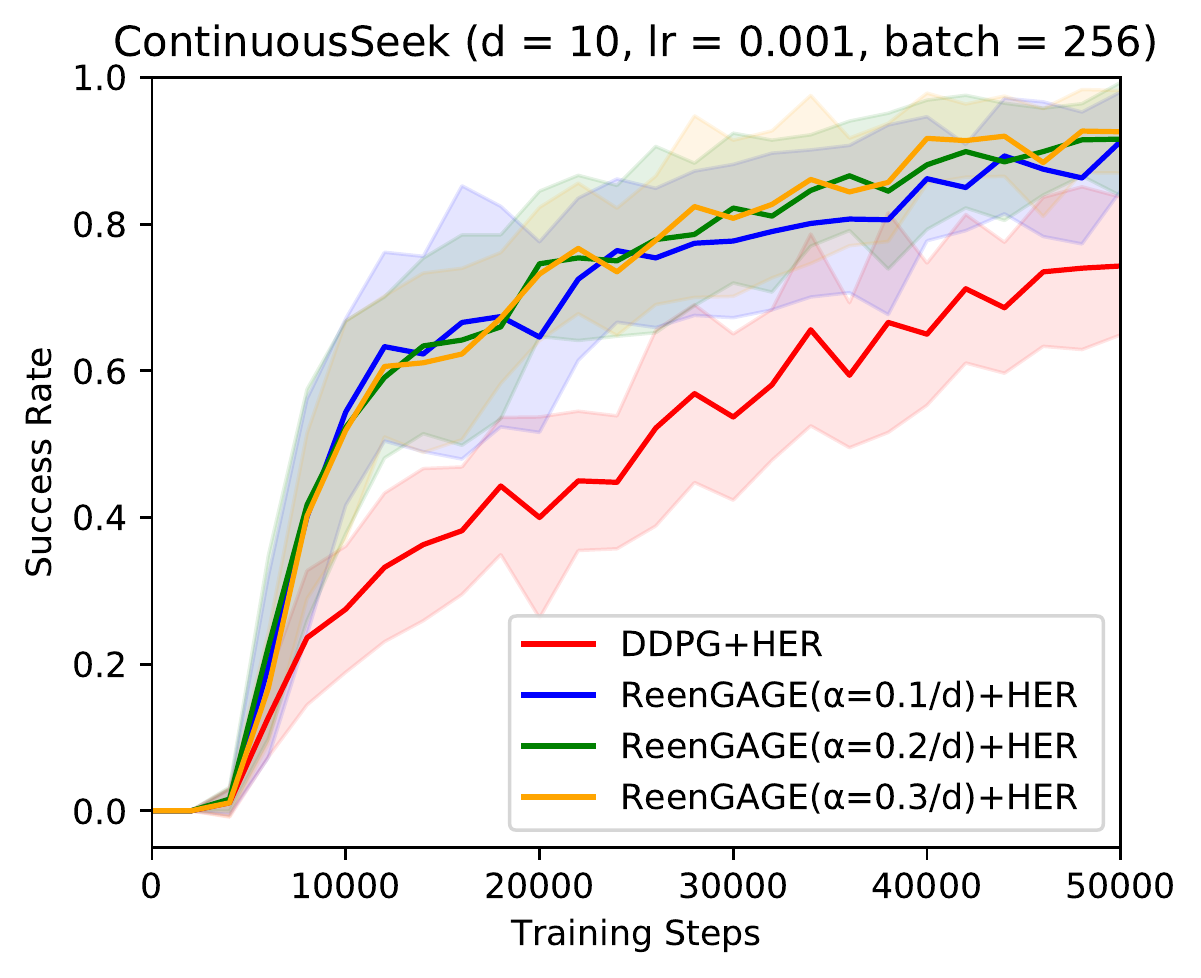}
     \end{subfigure}
     \hfill
     \begin{subfigure}[b]{0.33\textwidth}
         \centering
         \includegraphics[width=\textwidth]{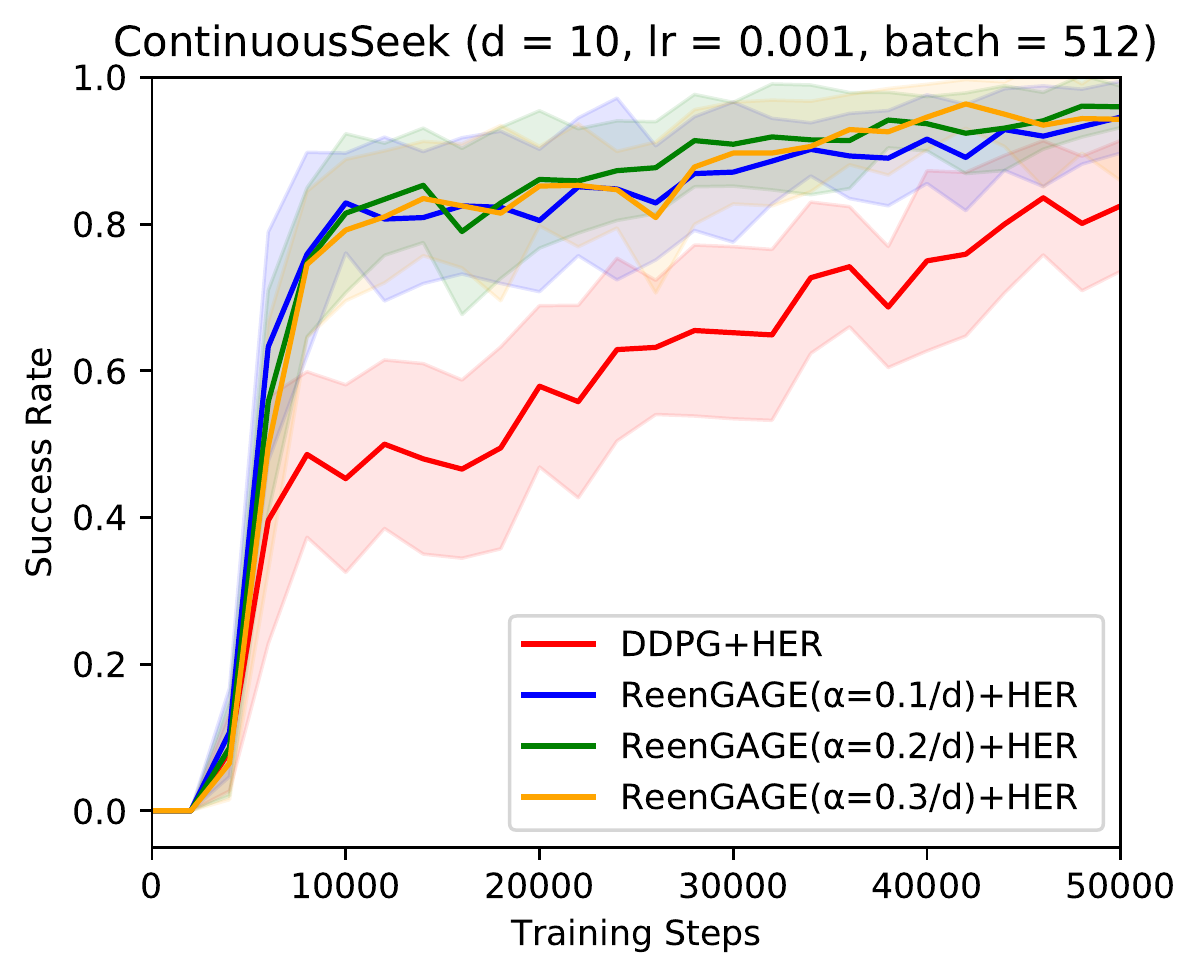}
     \end{subfigure}
     \begin{subfigure}[b]{0.33\textwidth}
         \centering
         \includegraphics[width=\textwidth]{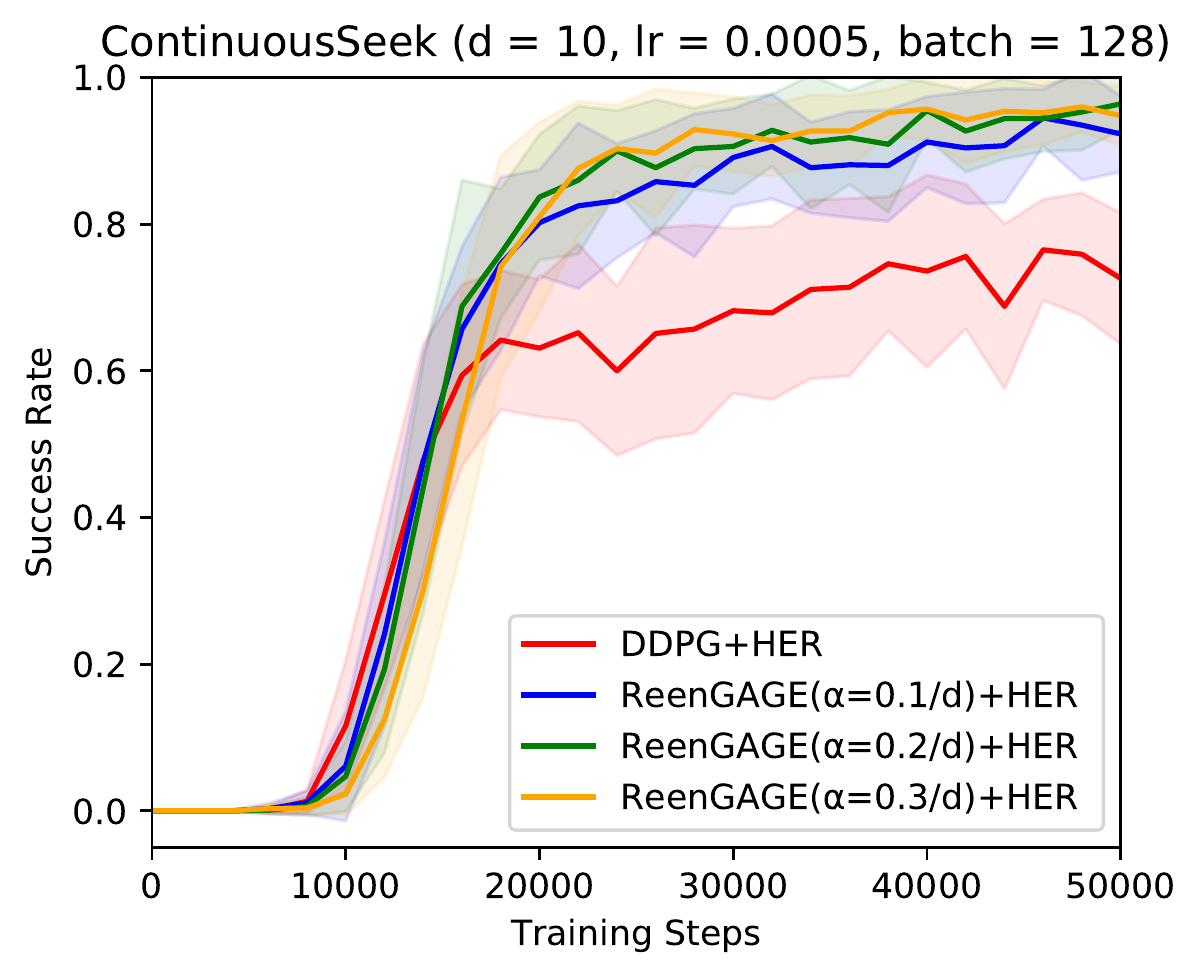}
     \end{subfigure}
     \hfill
     \begin{subfigure}[b]{0.33\textwidth}
         \centering
         \includegraphics[width=\textwidth]{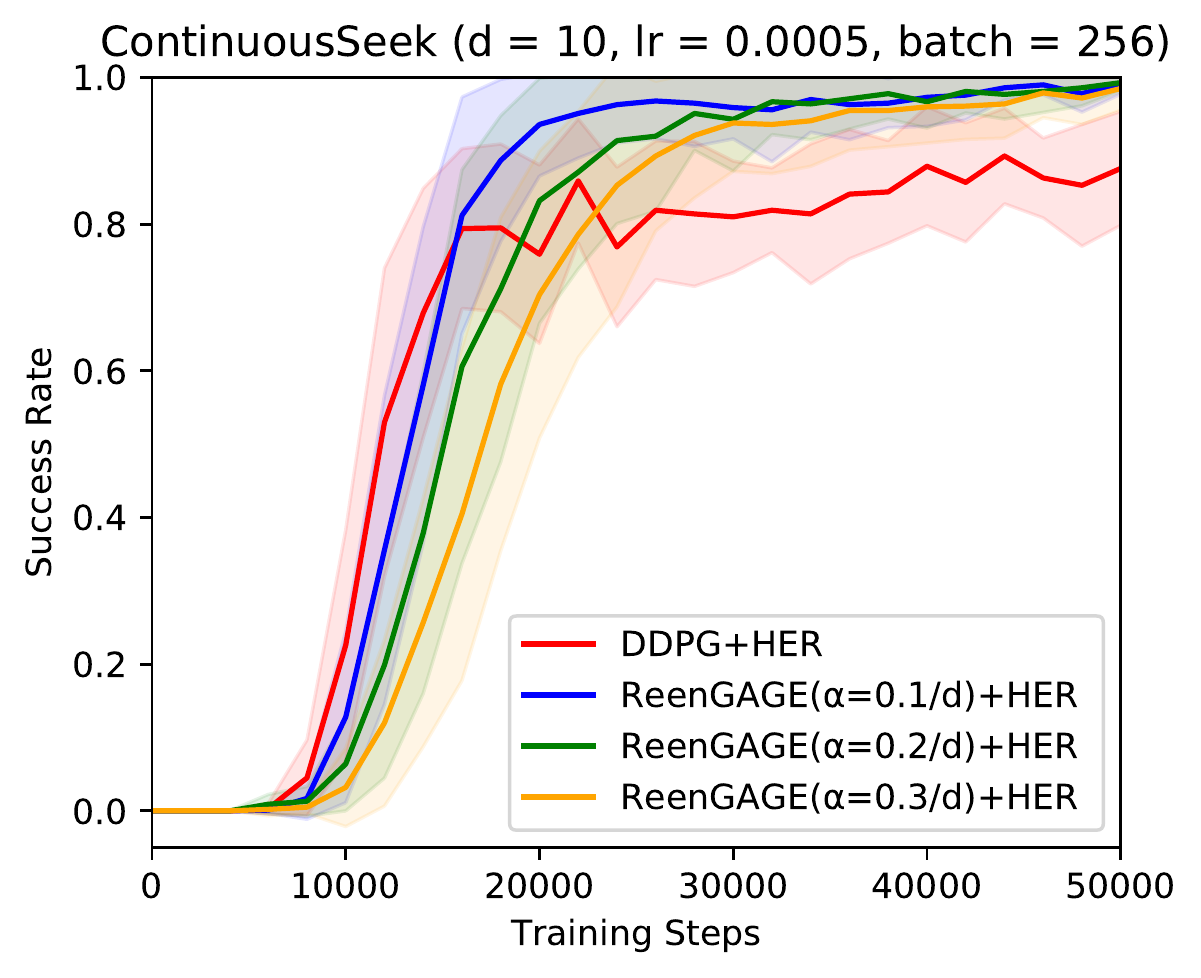}
     \end{subfigure}
     \hfill
     \begin{subfigure}[b]{0.33\textwidth}
         \centering
         \includegraphics[width=\textwidth]{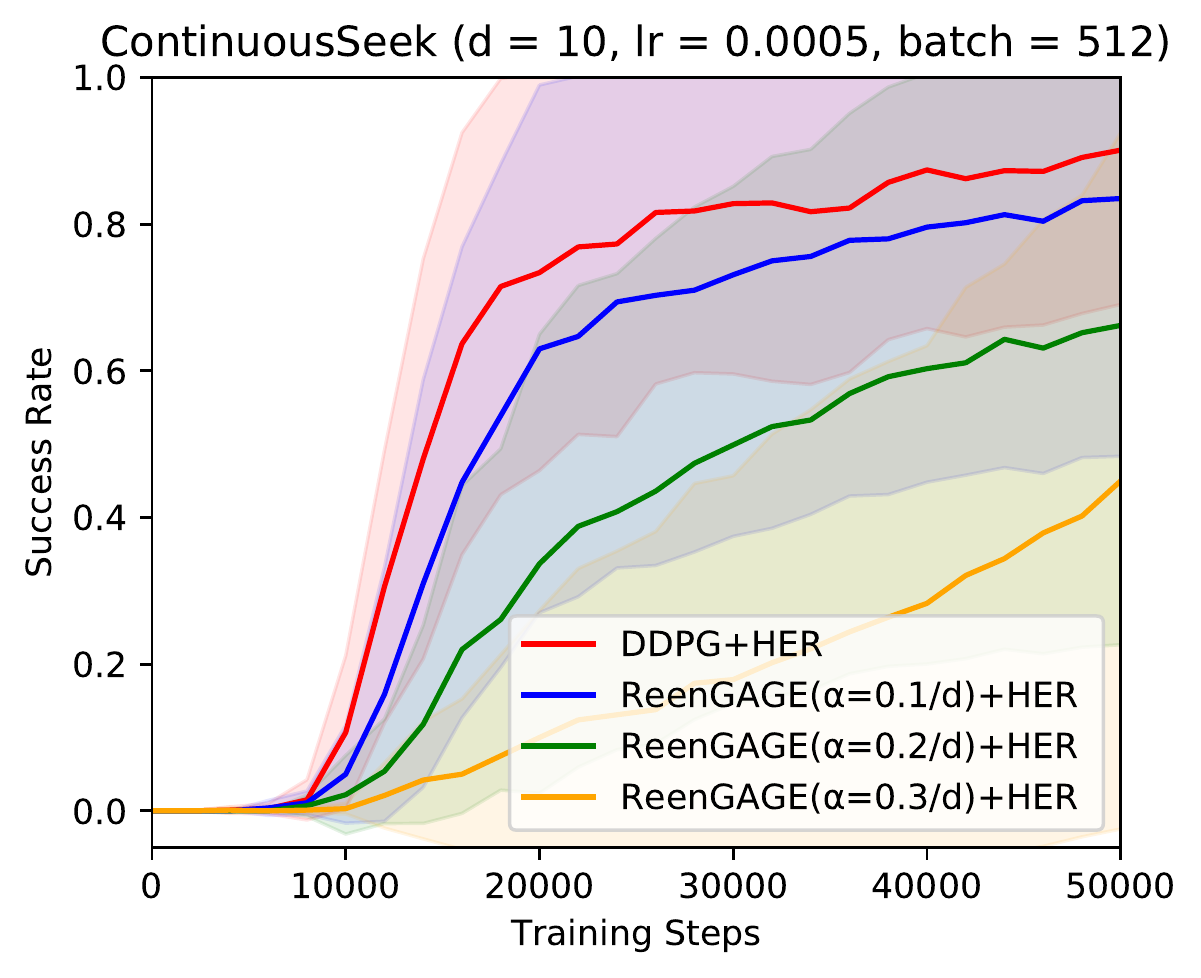}
     \end{subfigure}
     \begin{subfigure}[b]{0.33\textwidth}
         \centering
         \includegraphics[width=\textwidth]{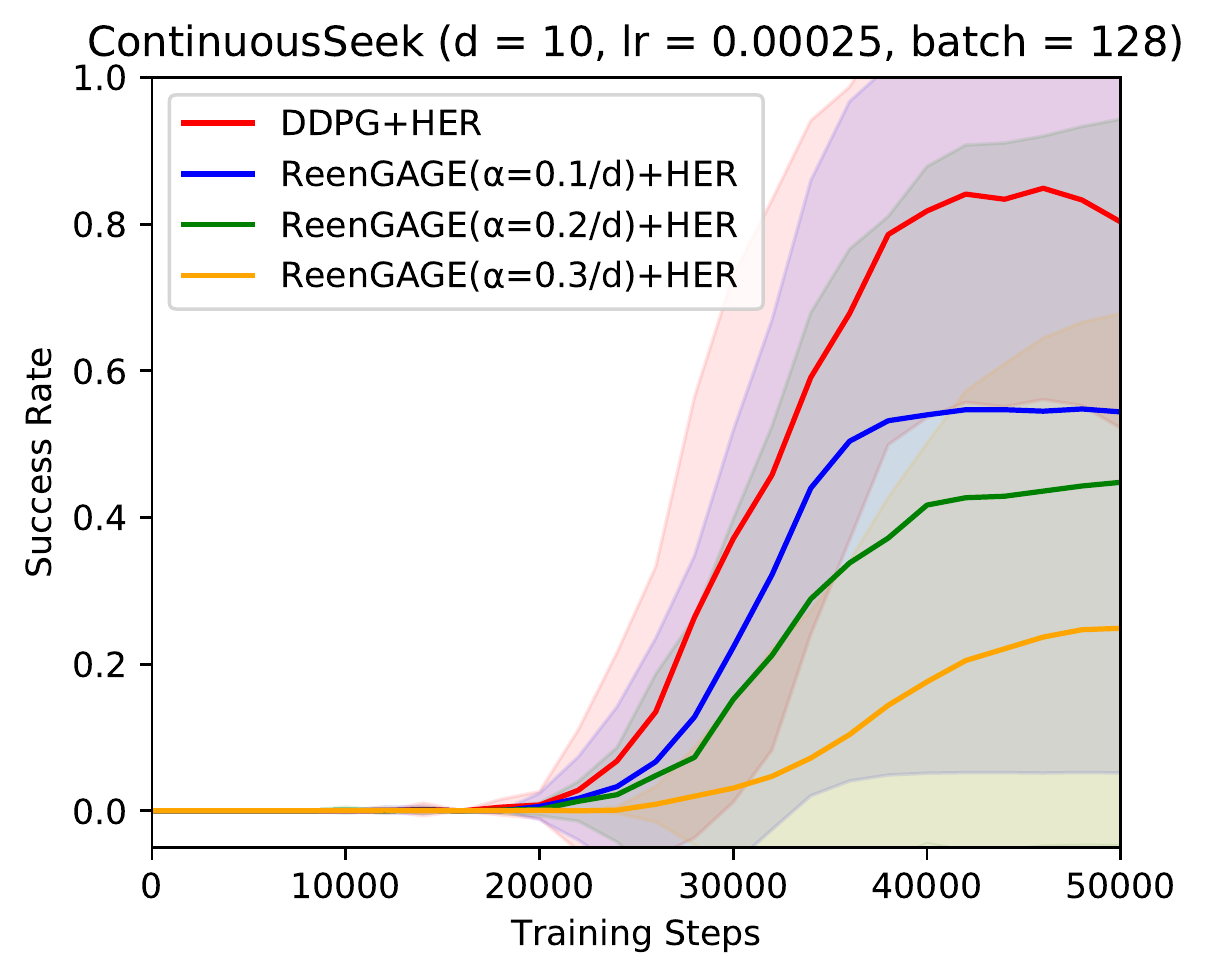}
     \end{subfigure}
     \hfill
     \begin{subfigure}[b]{0.33\textwidth}
         \centering
         \includegraphics[width=\textwidth]{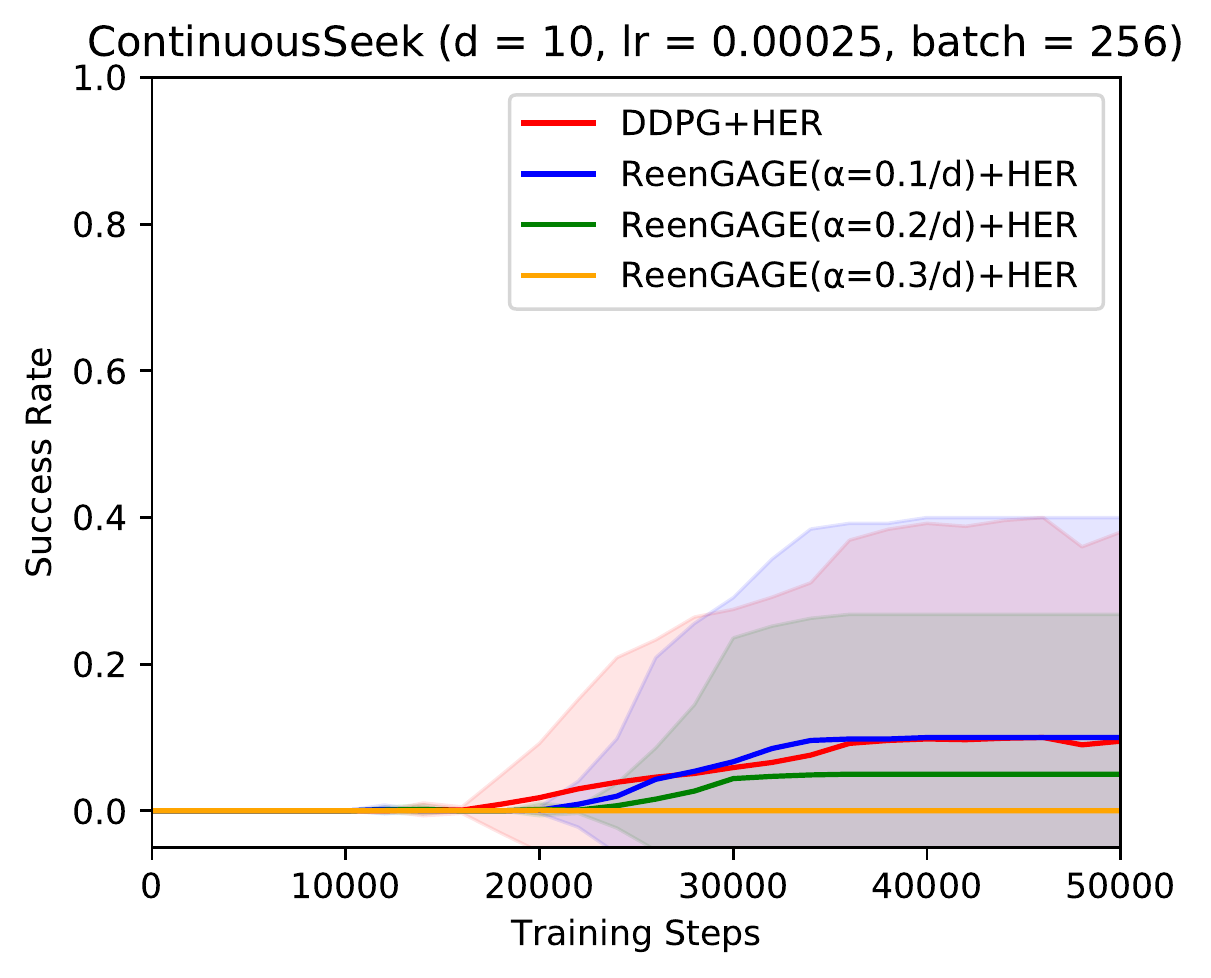}
     \end{subfigure}
     \hfill
     \begin{subfigure}[b]{0.33\textwidth}
         \centering
         \includegraphics[width=\textwidth]{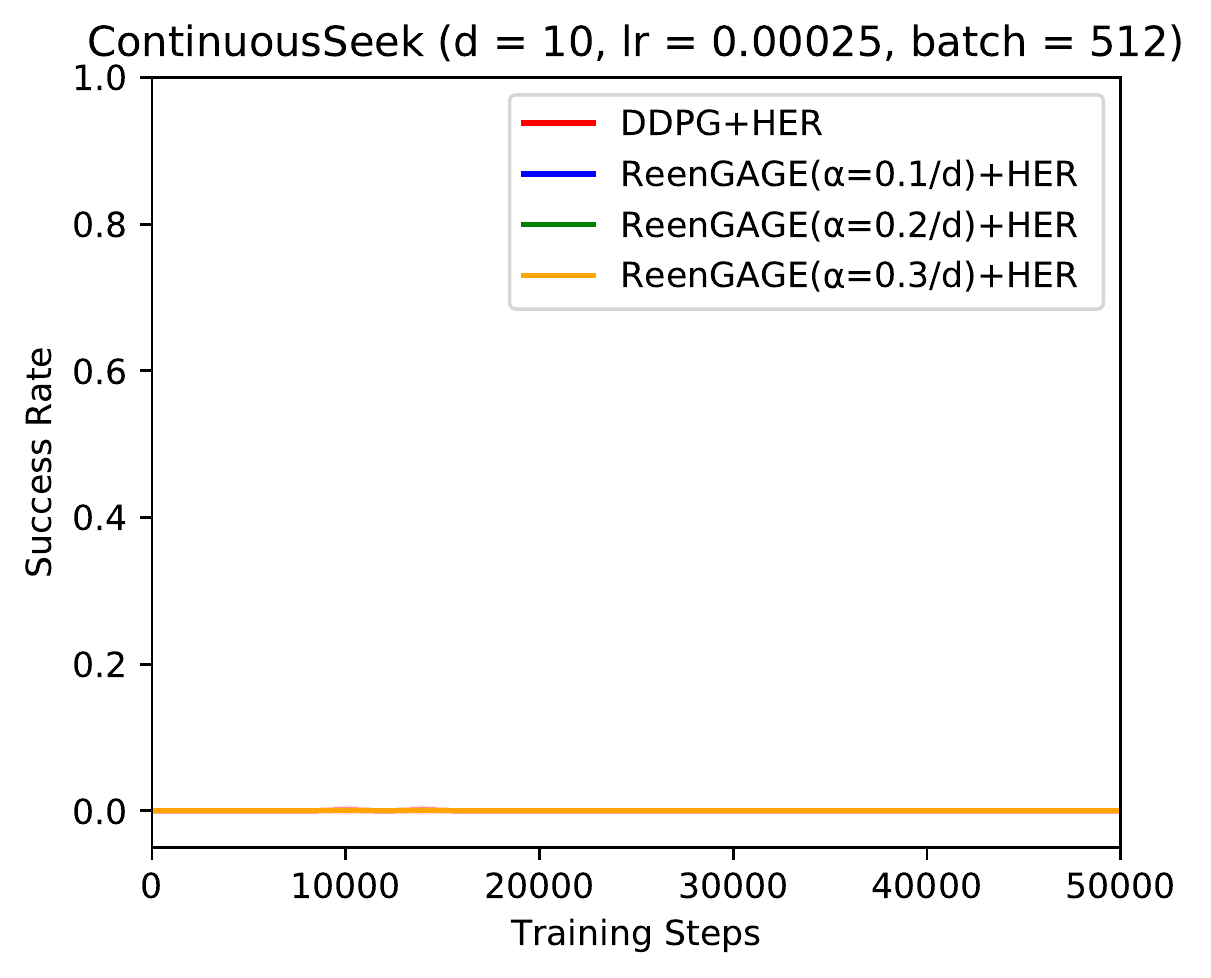}
     \end{subfigure}
        \caption{Complete ContinuousSeek Results for $d=10$.}
        \label{fig:continuous_seek_appdx_10}
\end{figure*}

\begin{figure*}
     \centering
     \begin{subfigure}[b]{0.33\textwidth}
         \centering
         \includegraphics[width=\textwidth]{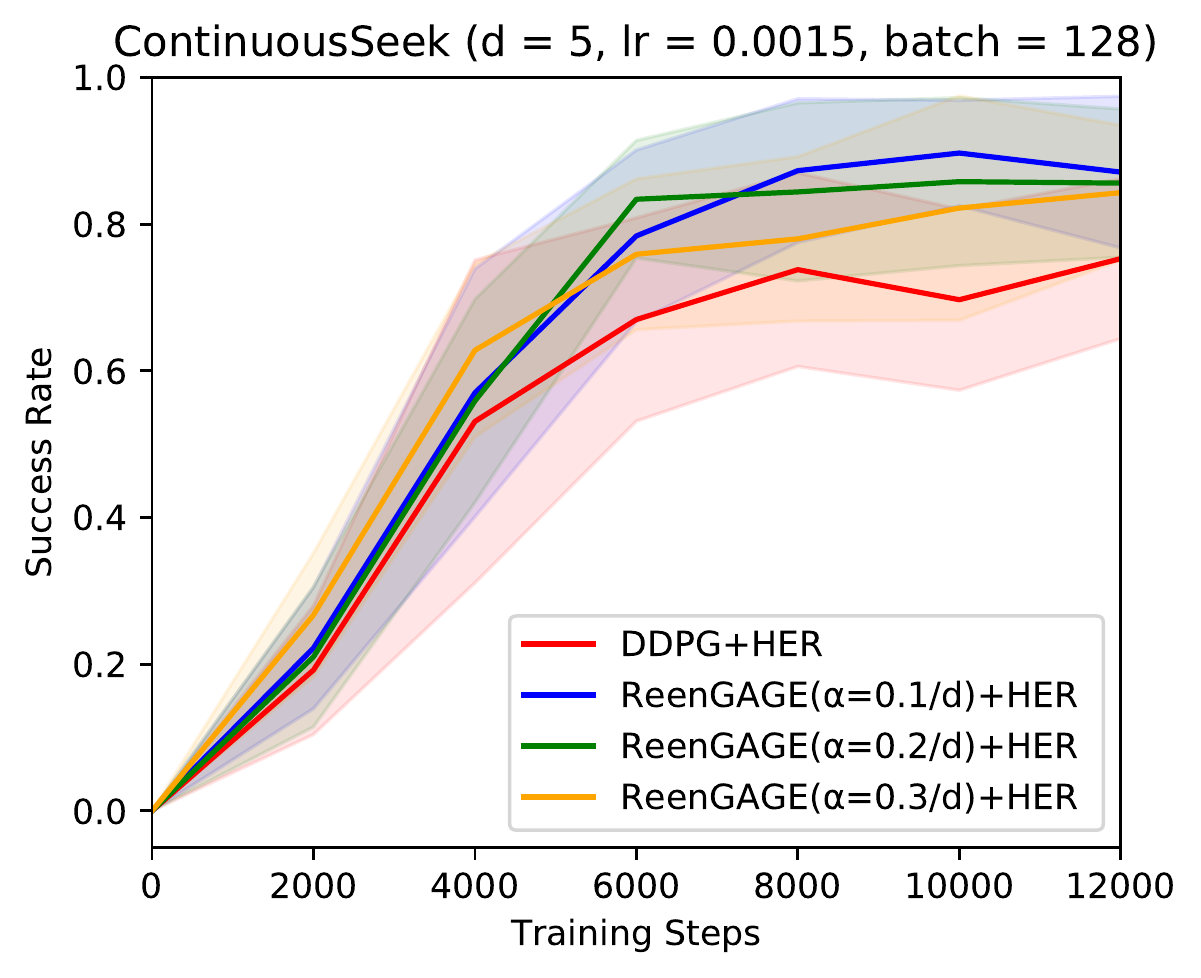}
     \end{subfigure}
     \hfill
     \begin{subfigure}[b]{0.33\textwidth}
         \centering
         \includegraphics[width=\textwidth]{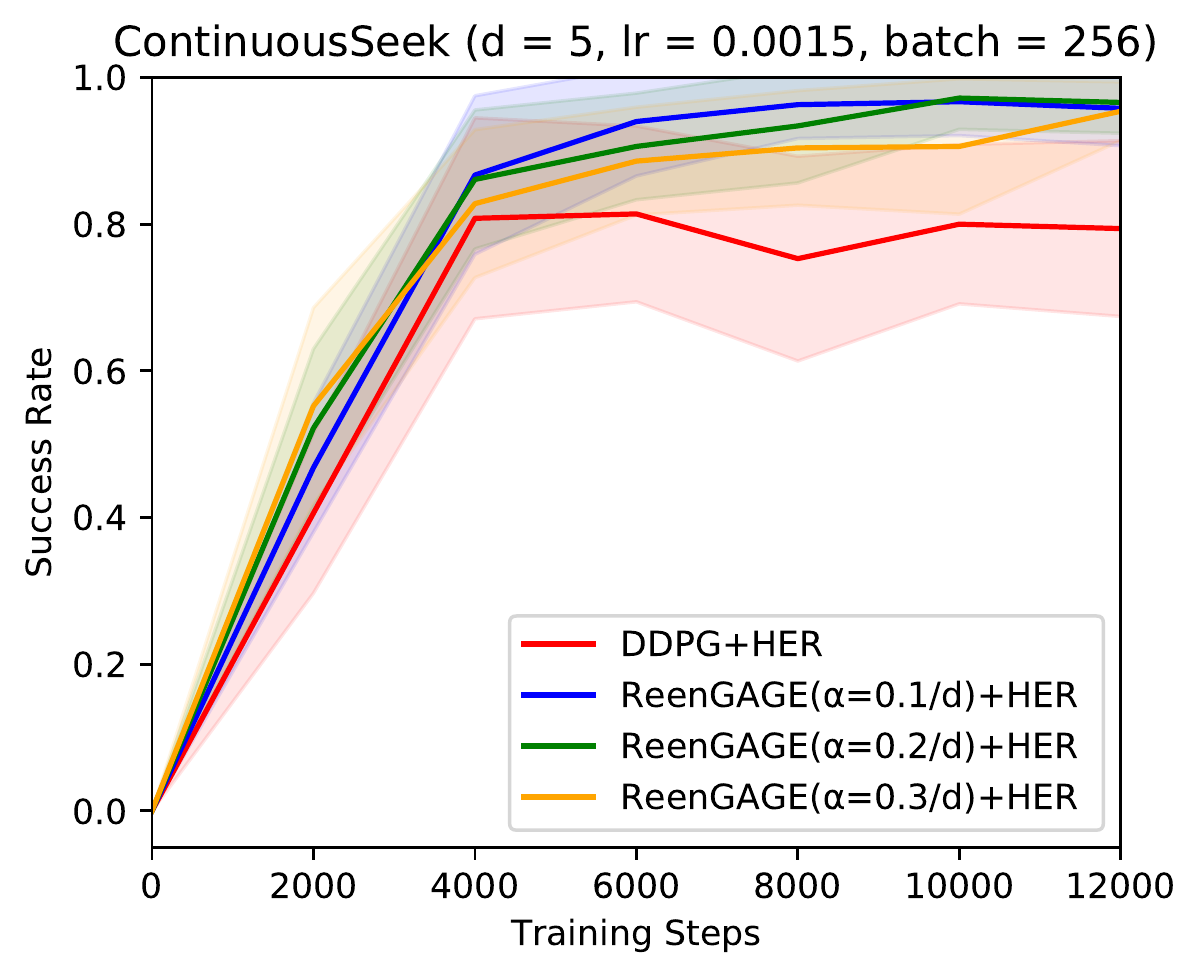}
     \end{subfigure}
     \hfill
     \begin{subfigure}[b]{0.33\textwidth}
         \centering
         \includegraphics[width=\textwidth]{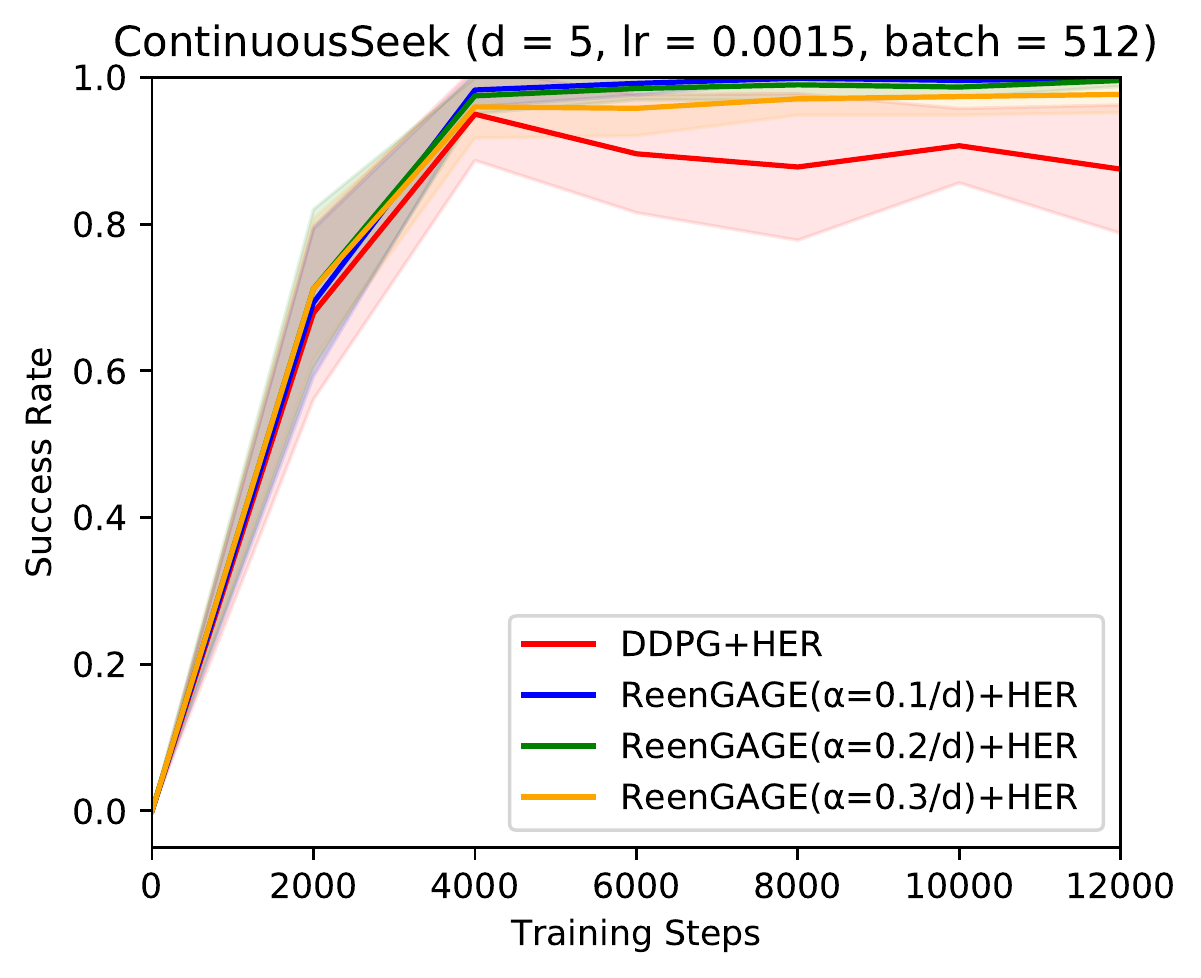}
     \end{subfigure}
     \begin{subfigure}[b]{0.33\textwidth}
         \centering
         \includegraphics[width=\textwidth]{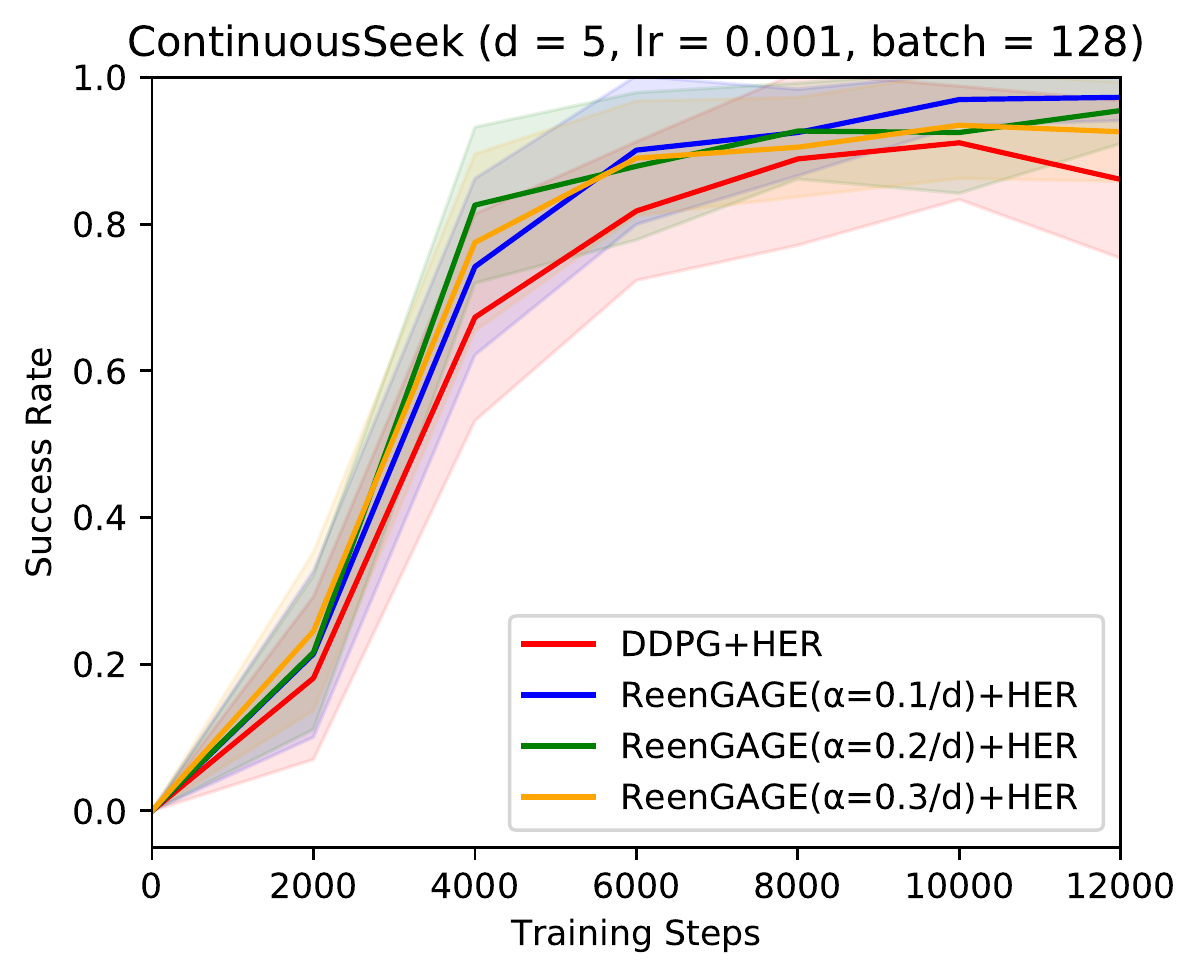}
     \end{subfigure}
     \hfill
     \begin{subfigure}[b]{0.33\textwidth}
         \centering
         \includegraphics[width=\textwidth]{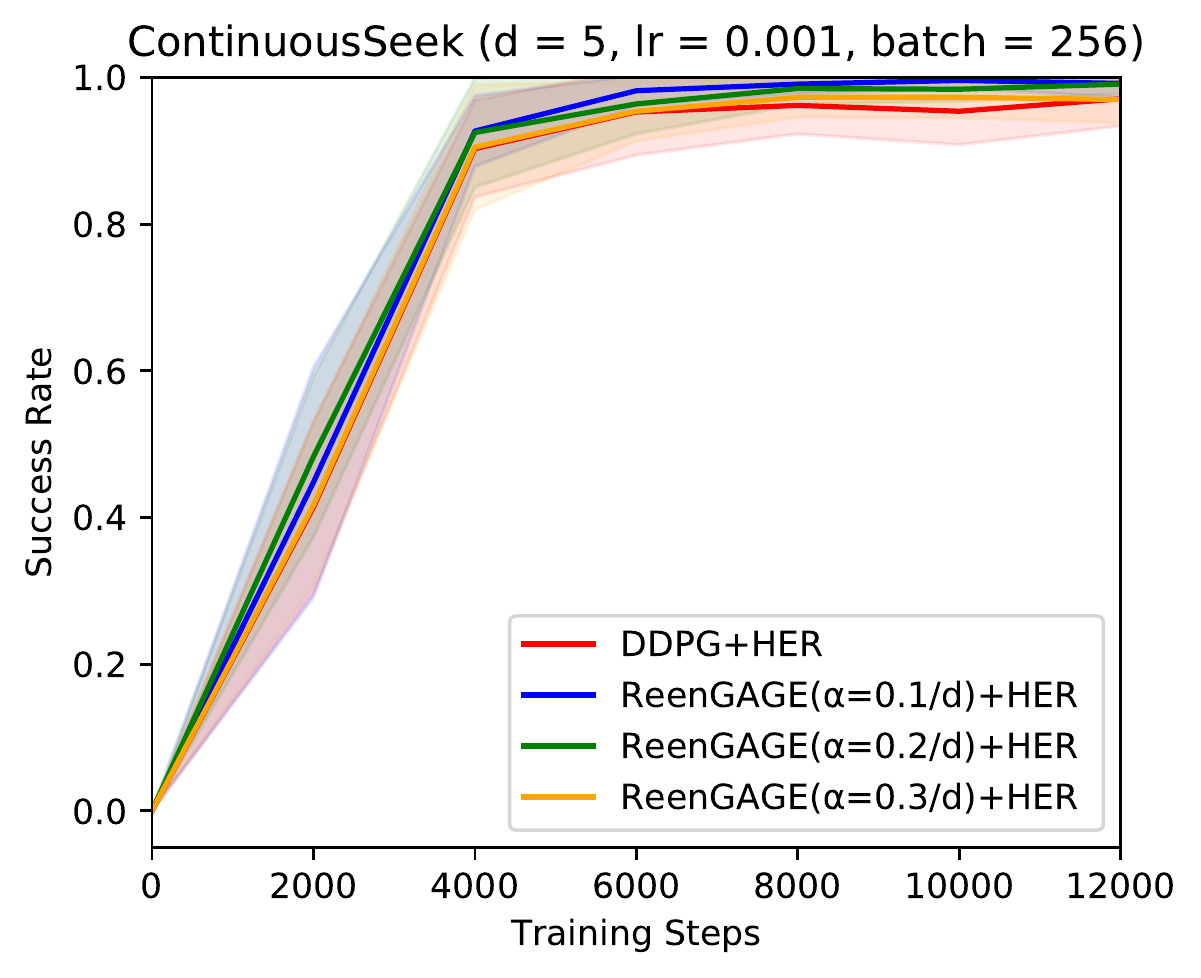}
     \end{subfigure}
     \hfill
     \begin{subfigure}[b]{0.33\textwidth}
         \centering
         \includegraphics[width=\textwidth]{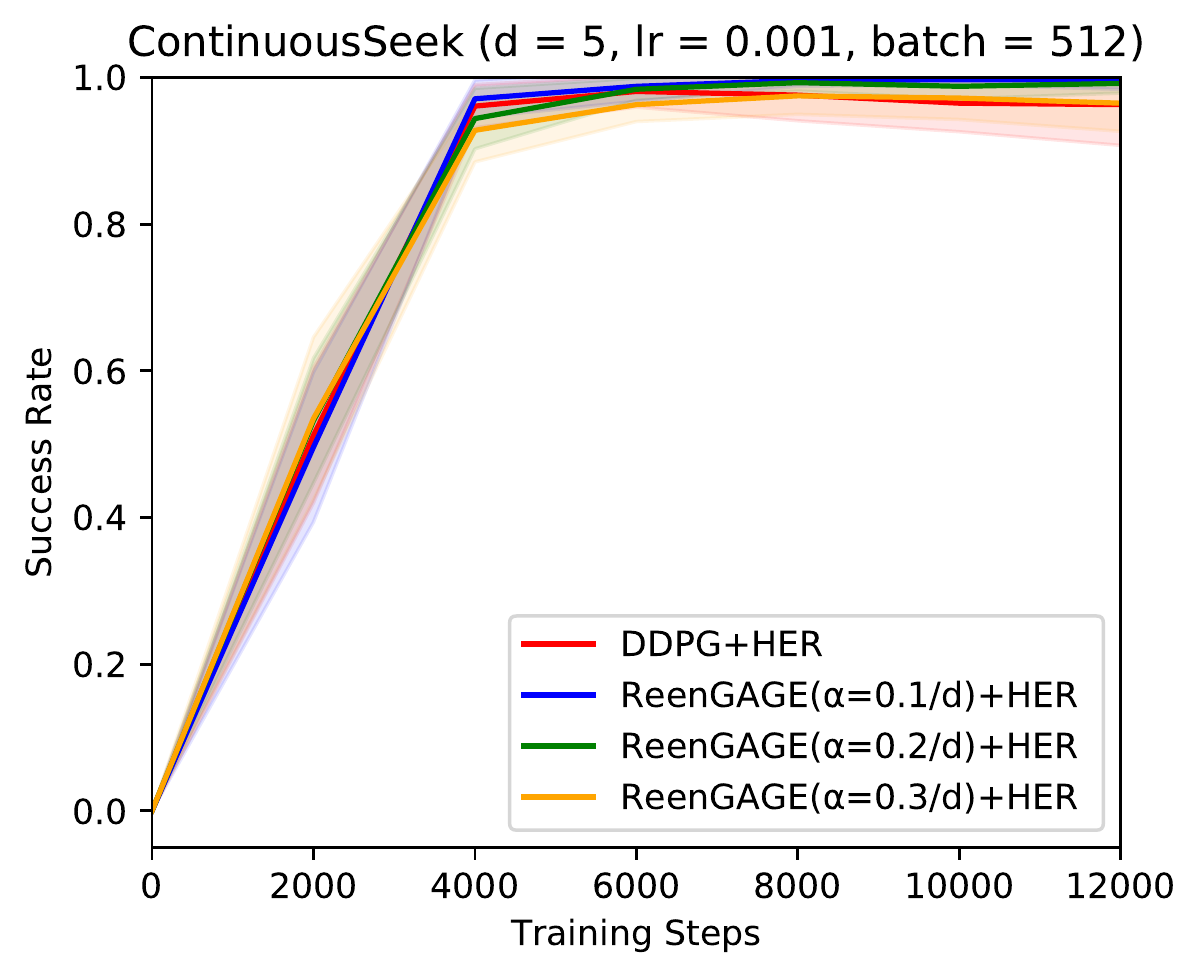}
     \end{subfigure}
     \begin{subfigure}[b]{0.33\textwidth}
         \centering
         \includegraphics[width=\textwidth]{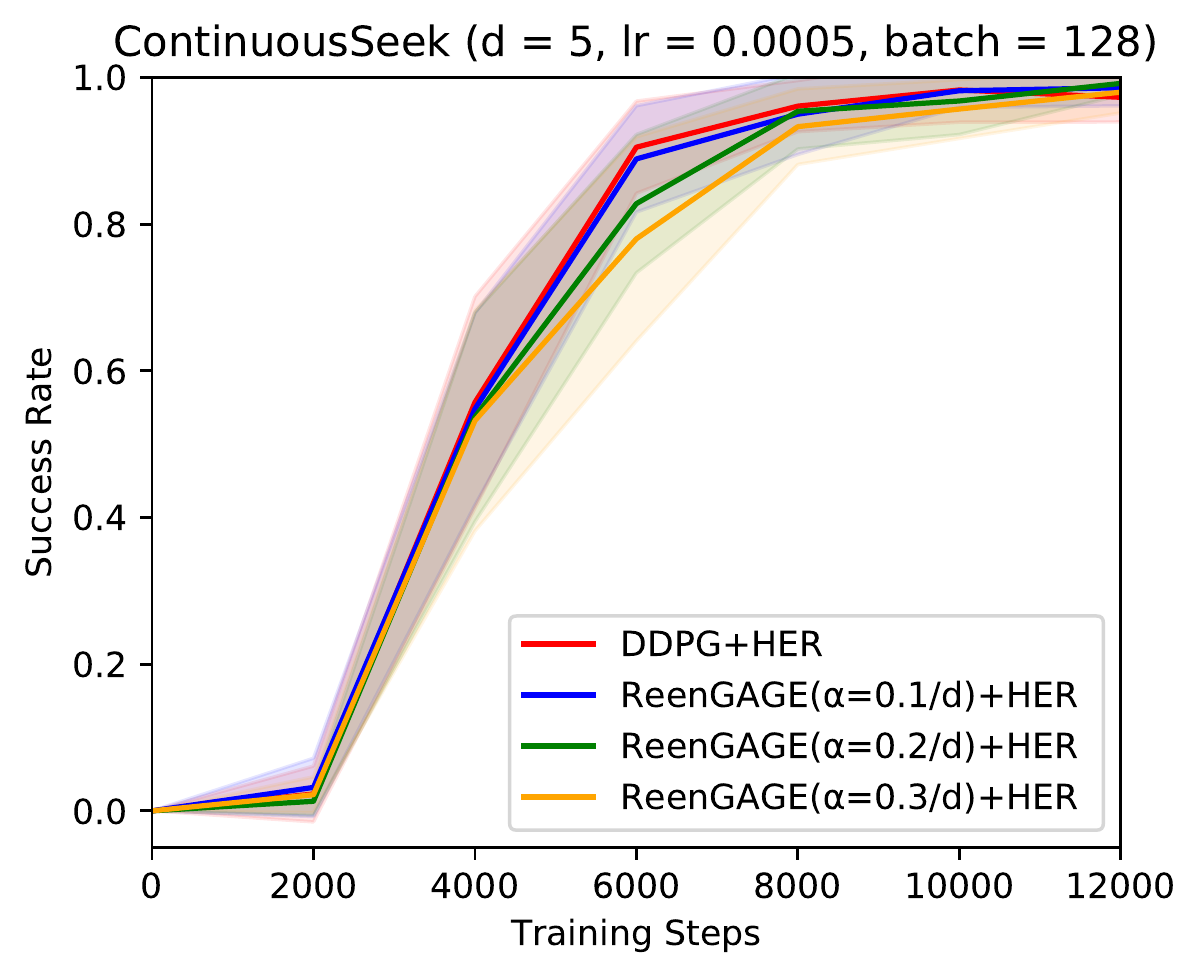}
     \end{subfigure}
     \hfill
     \begin{subfigure}[b]{0.33\textwidth}
         \centering
         \includegraphics[width=\textwidth]{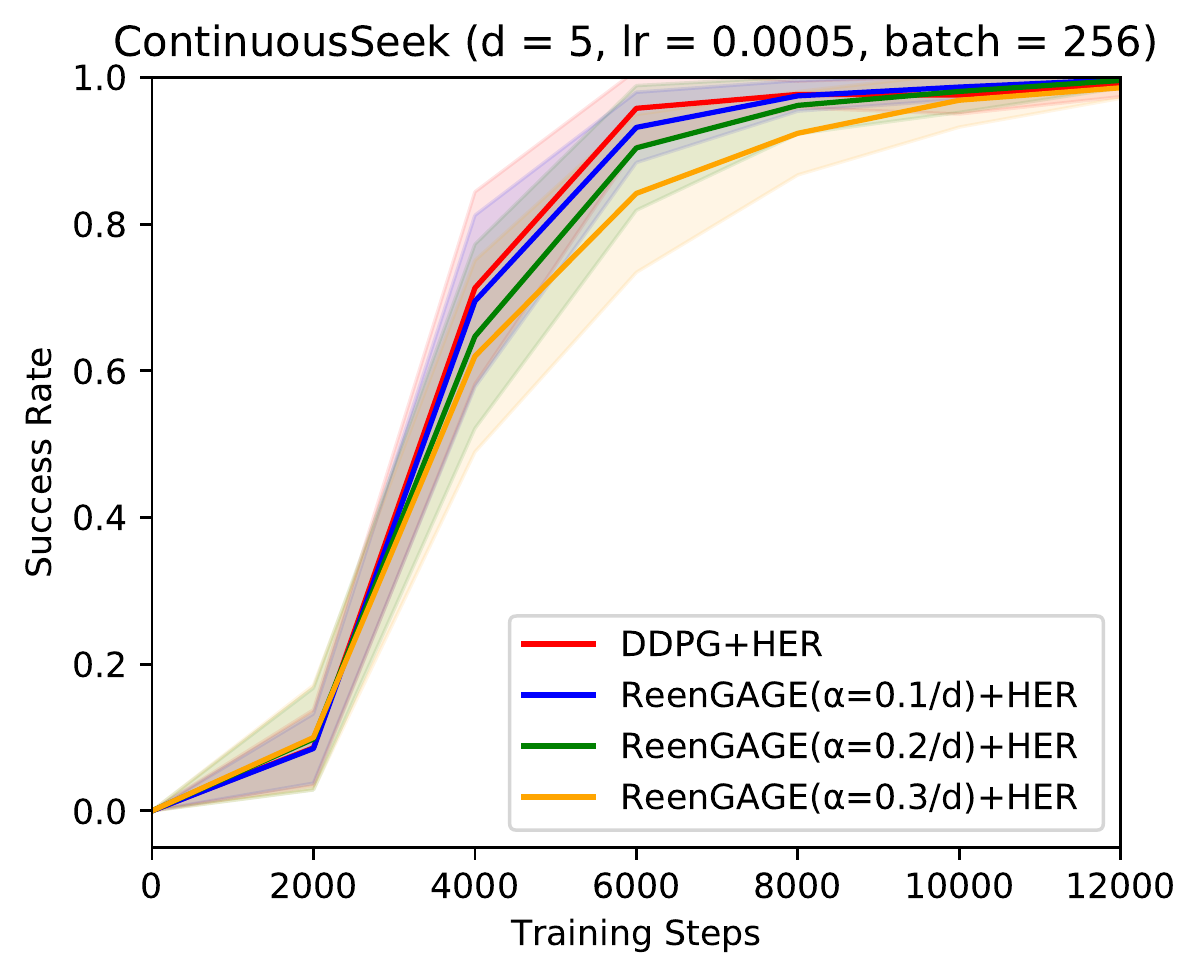}
     \end{subfigure}
     \hfill
     \begin{subfigure}[b]{0.33\textwidth}
         \centering
         \includegraphics[width=\textwidth]{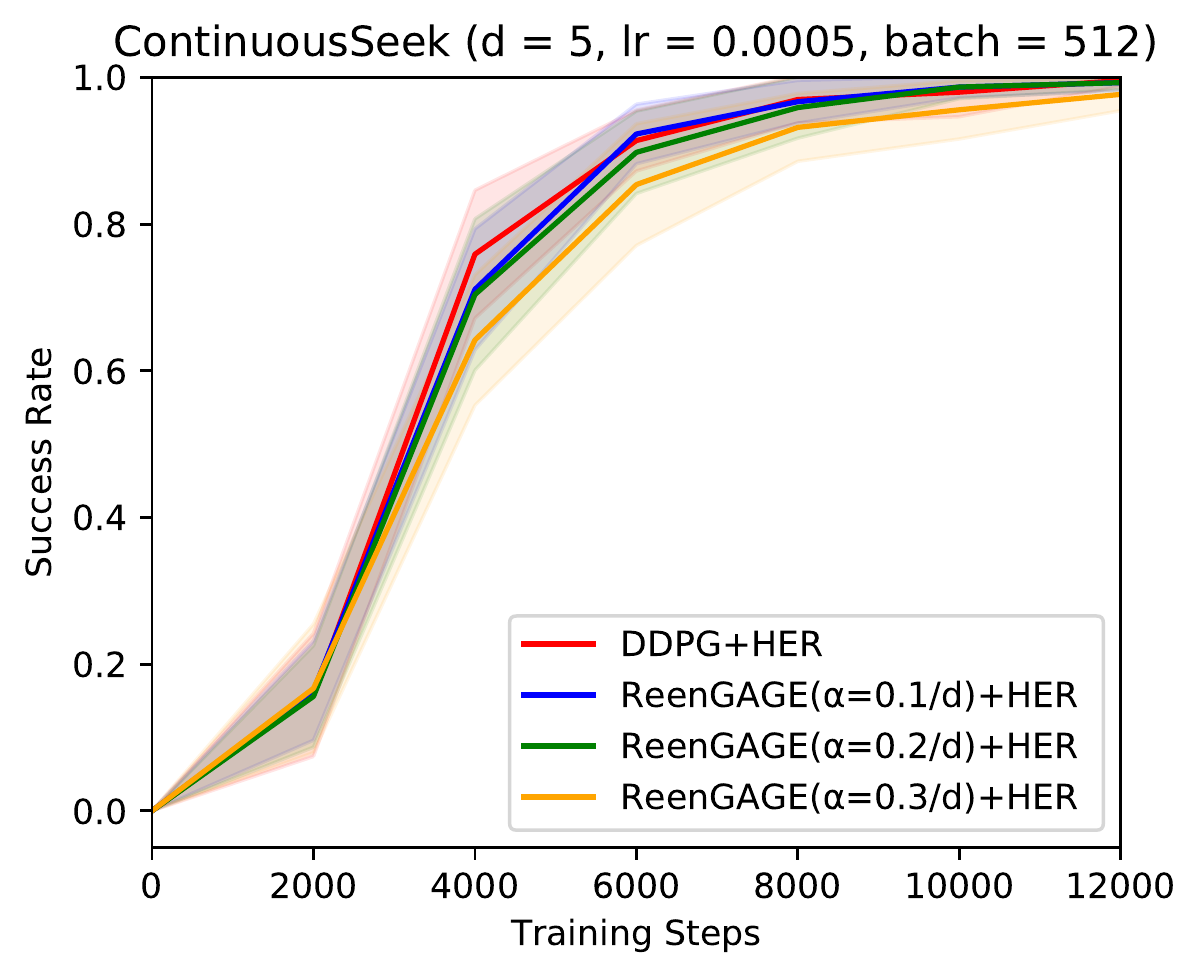}
     \end{subfigure}
     \begin{subfigure}[b]{0.33\textwidth}
         \centering
         \includegraphics[width=\textwidth]{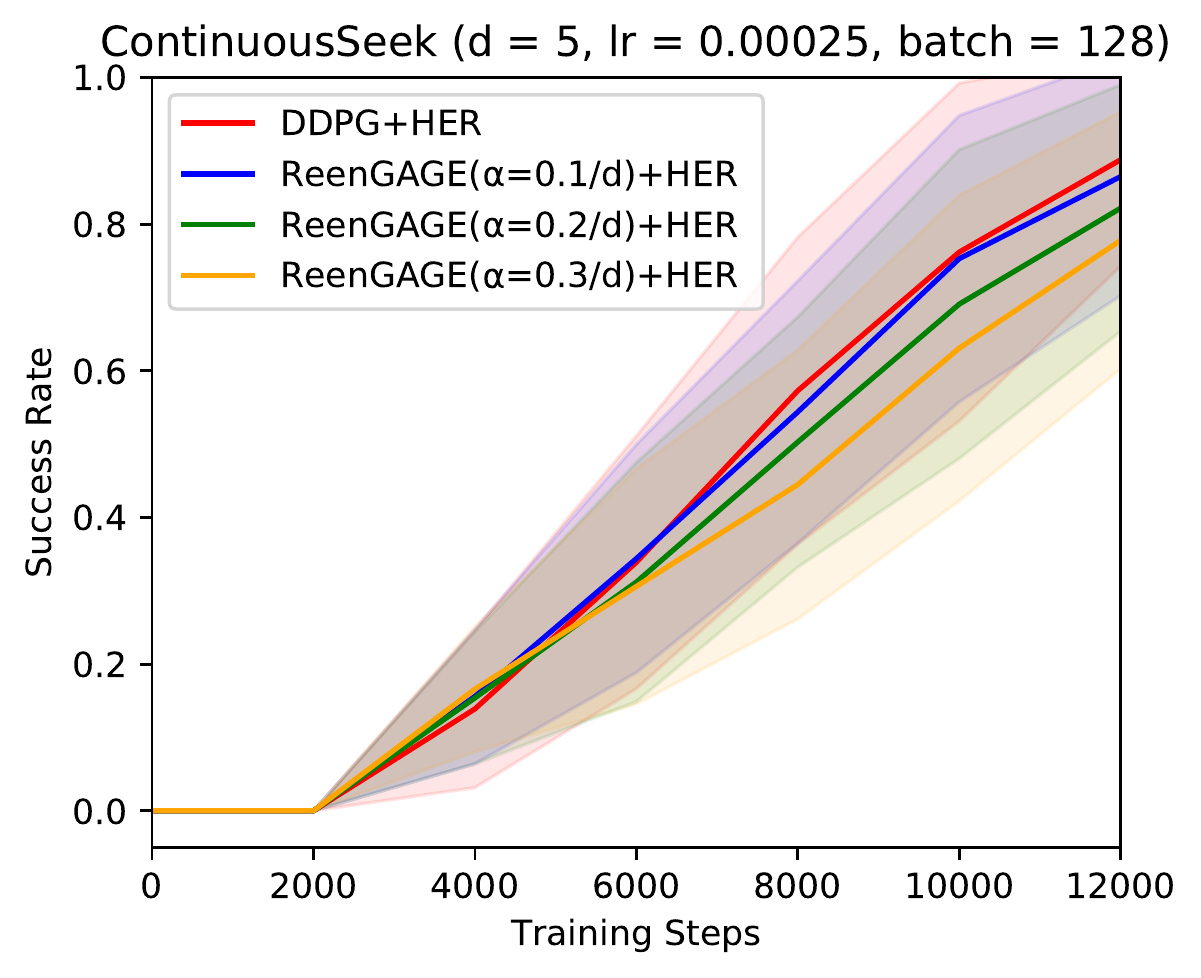}
     \end{subfigure}
     \hfill
     \begin{subfigure}[b]{0.33\textwidth}
         \centering
         \includegraphics[width=\textwidth]{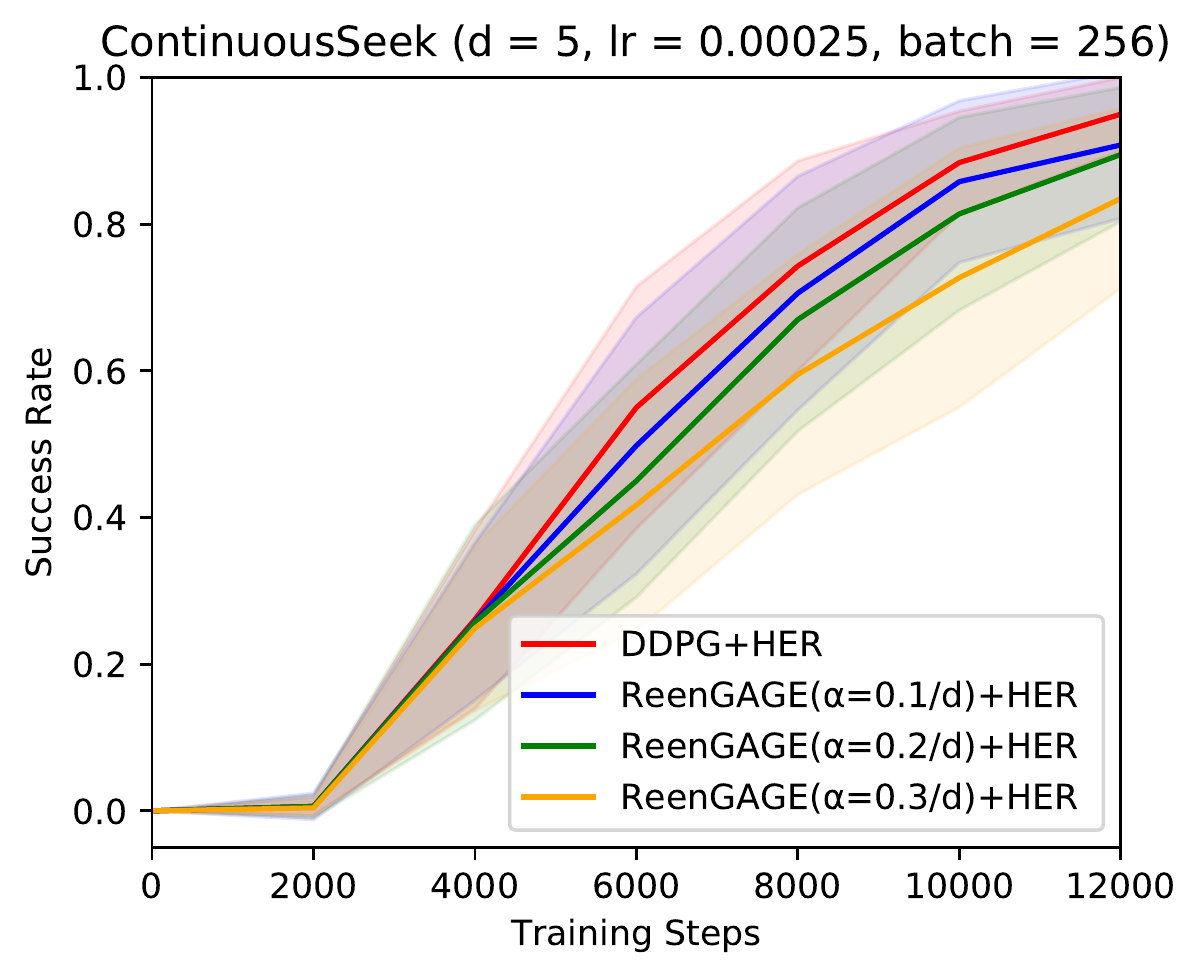}
     \end{subfigure}
     \hfill
     \begin{subfigure}[b]{0.33\textwidth}
         \centering
         \includegraphics[width=\textwidth]{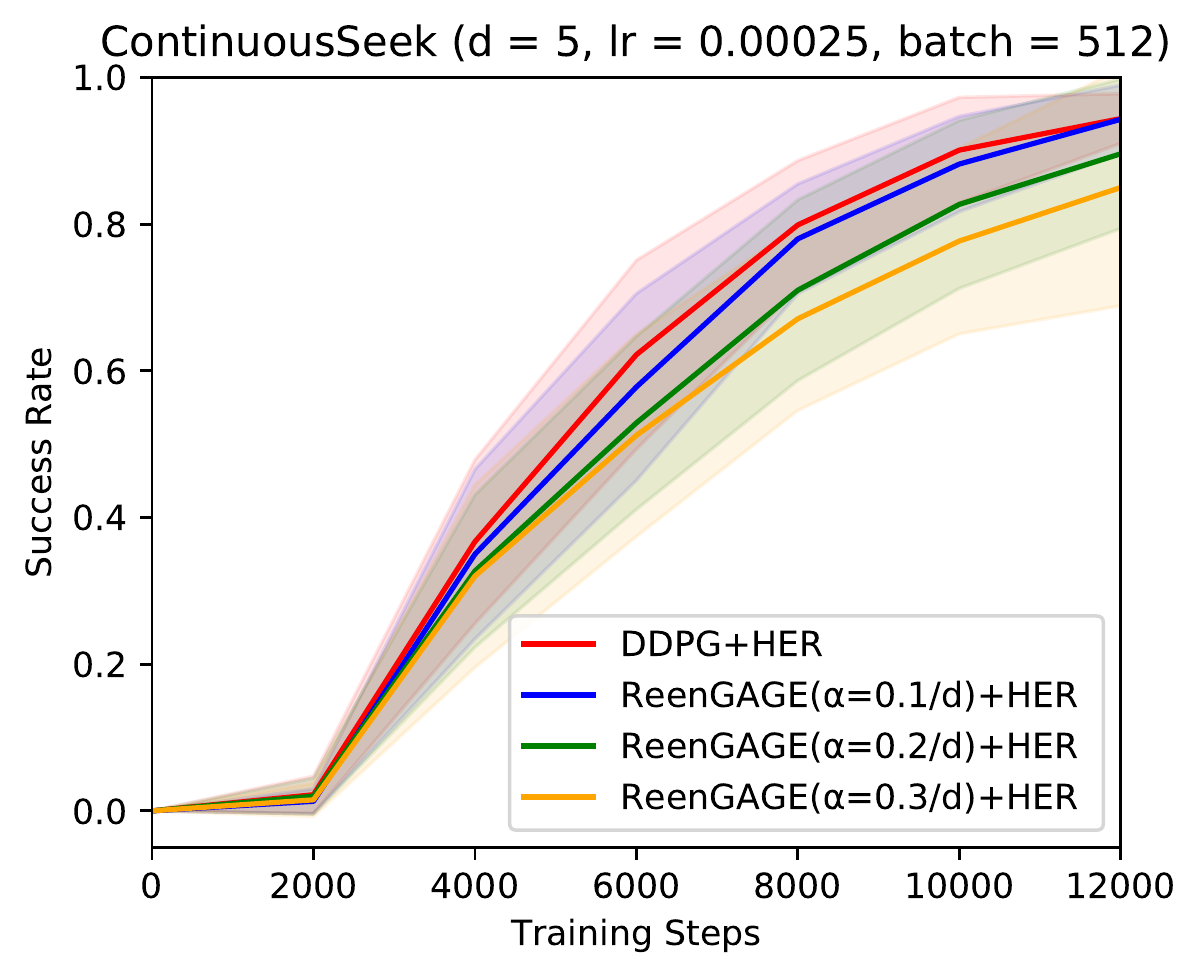}
     \end{subfigure}
        \caption{Complete ContinuousSeek Results for $d=5$.}
        \label{fig:continuous_seek_appdx_5}
\end{figure*}

As mentioned in the main text, we performed a full hyperparameter search over the batch size (in $\{128,256,512\}$) and learning rate (in $\{0.00025,0.0005,0.001,0.0015\}$). The results in the main paper represent the best single curve (as defined by area under the curve, or in other words best average score over time) for the baseline and each value of the ReenGAGE $\alpha$ term. Complete results are presented in Figures \ref{fig:continuous_seek_appdx_20}, \ref{fig:continuous_seek_appdx_10}, and \ref{fig:continuous_seek_appdx_5}. The values of these parameters which yielded the best average performance, and were therefore reported in the main text, were as follows (Table \ref{tab:main_text_batch_lr}): 
\begin{table}[H]
    \centering
    \begin{tabular}{|c|c|c|c|c|c|c|c|c|}
    \hline
         & \multicolumn{2}{|c|}{DDPG+HER}   & \multicolumn{2}{|c|}{ReenGAGE($\alpha=0.1$)+HER} & \multicolumn{2}{|c|}{ReenGAGE($\alpha=0.2$)+HER} & \multicolumn{2}{|c|}{ReenGAGE($\alpha=0.3$)+HER}  \\
    \hline
         & Batch& LR & Batch& LR & Batch& LR & Batch& LR\\
    \hline     
        $d=5$  &512&0.001&512&0.0015&512&0.0015&512&0.0015\\
    \hline
        $d=10$  &256&0.0005&512&0.001&512&0.001&512&0.001\\
    \hline
        $d=20$  &256&0.0005&128&0.0005&128&0.0005&128&0.0005\\
    \hline
    \end{tabular}
    \caption{``Best'' batch sizes and learning rates for ContinuousSeek for DDPG and ReenGAGE. }
    \label{tab:main_text_batch_lr}
\end{table}
Note that for the larger-scale experiments ($d=10$ and $d=20$) the ``best'' hyperparameters for the baseline DDPG+HER model lie in the interior of the search space of both the batch size and learning rate: this would seem to imply (assuming concavity) that increasing the range of the search space of these parameters would likely not improve the baseline. However, this is not the case for ReenGAGE: we could perhaps achieve even better performance for ReenGAGE by conducting a larger search, specifically in the batch size dimension.

Other hyperparameters were fixed, and are listed in Table \ref{tab:continuous_seek_hyperparams}. We use the implementation of HER with DDPG from Stable-Baselines3 \cite{JMLR:v22:20-1364} as our baseline; any unlisted hyperparameters are the default from this package. Note that we use the ``online'' variant of HER provided in the Stable-Baselines3 package.


\newpage
\section{Additional Results for Robotics Experiments}
\begin{figure*}
     \centering
     \begin{subfigure}[b]{0.44\textwidth}
         \centering
         \includegraphics[width=\textwidth]{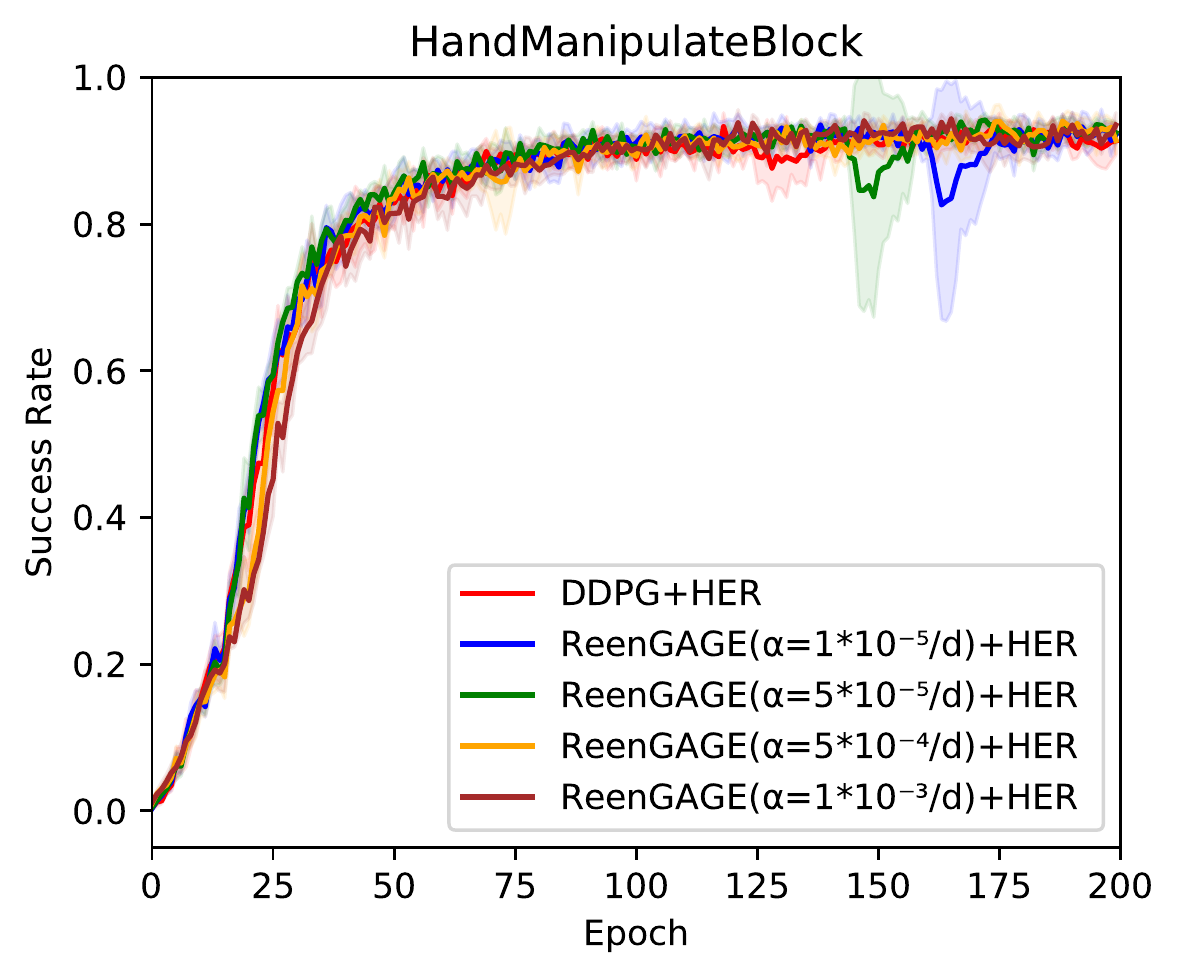}
         \caption{}
     \end{subfigure}
     \hfill
     \begin{subfigure}[b]{0.44\textwidth}
         \centering
         \includegraphics[width=\textwidth]{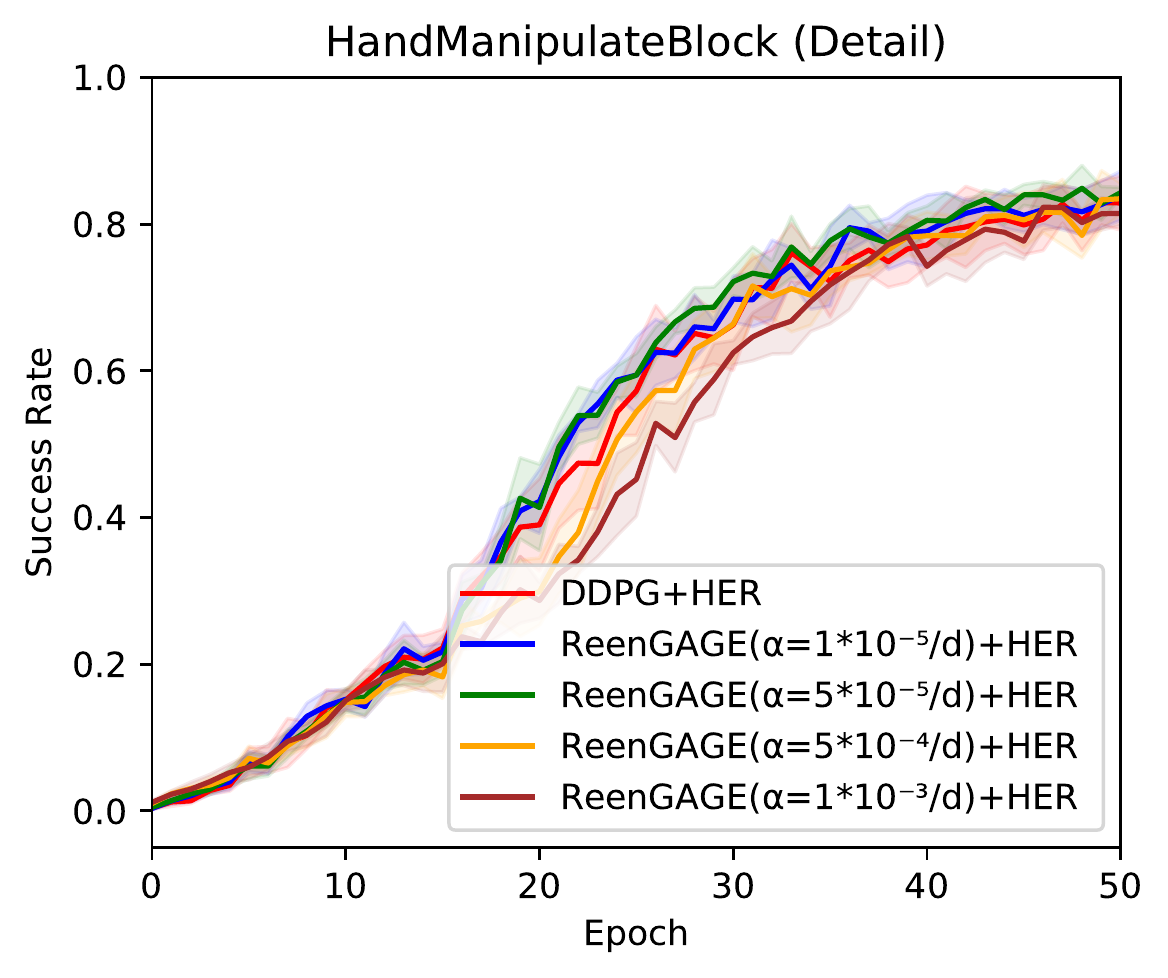}
         \caption{}
     \end{subfigure}
     \hfill
     \begin{subfigure}[b]{0.44\textwidth}
         \centering
         \includegraphics[width=\textwidth]{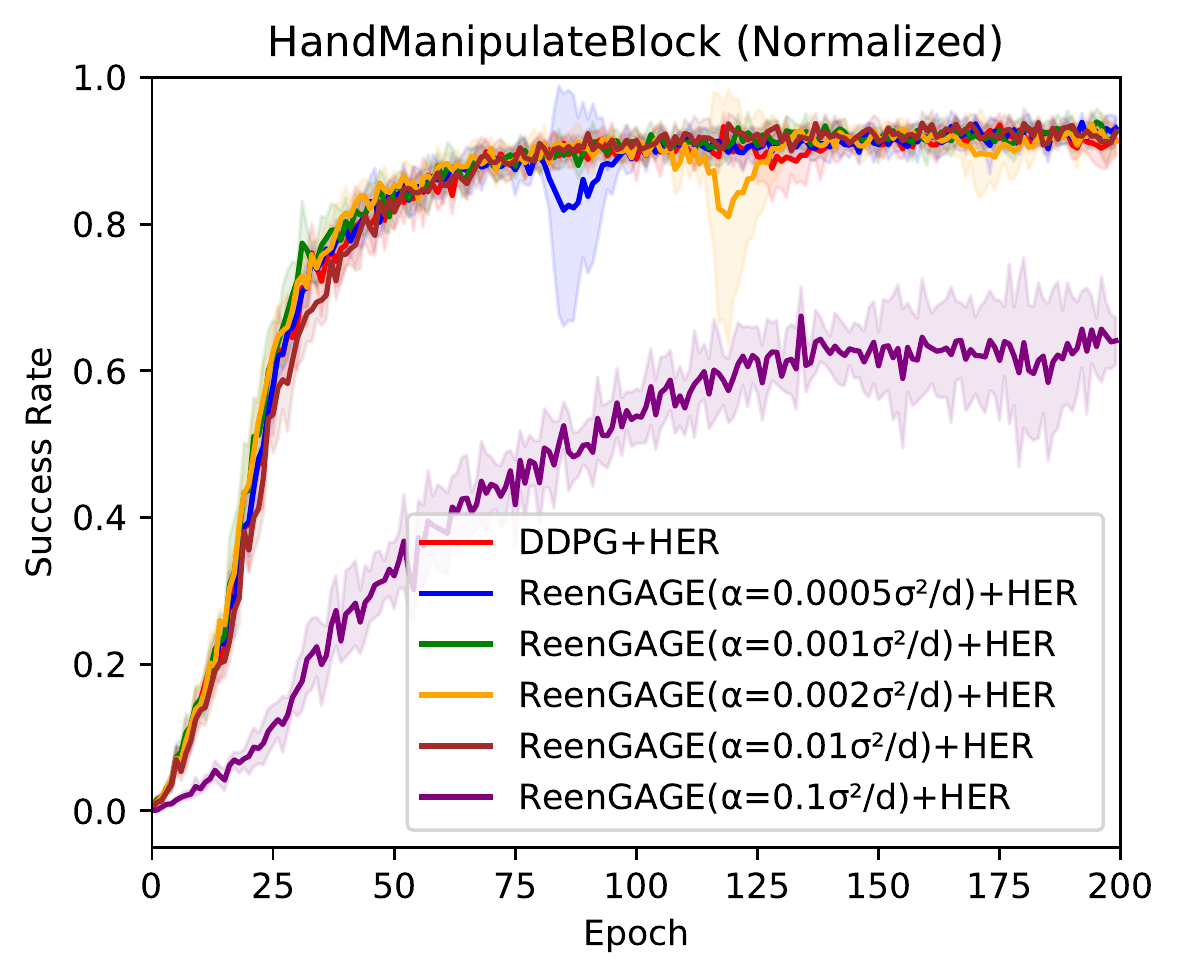}
         \caption{}
     \end{subfigure}
     \hfill
     \begin{subfigure}[b]{0.44\textwidth}
         \centering
         \includegraphics[width=\textwidth]{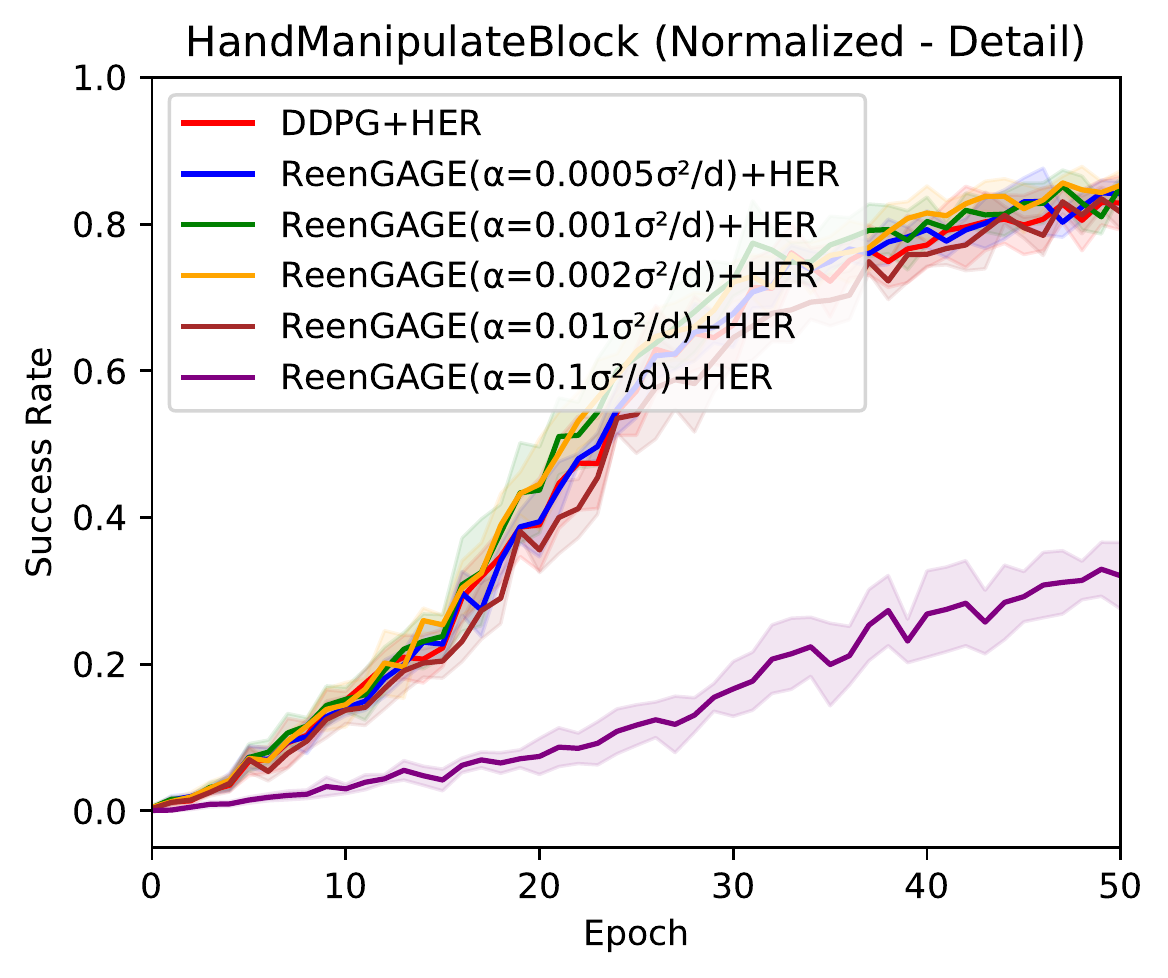}
         \caption{}
     \end{subfigure}
     \hfill
     \begin{subfigure}[b]{0.44\textwidth}
         \centering
         \includegraphics[width=\textwidth]{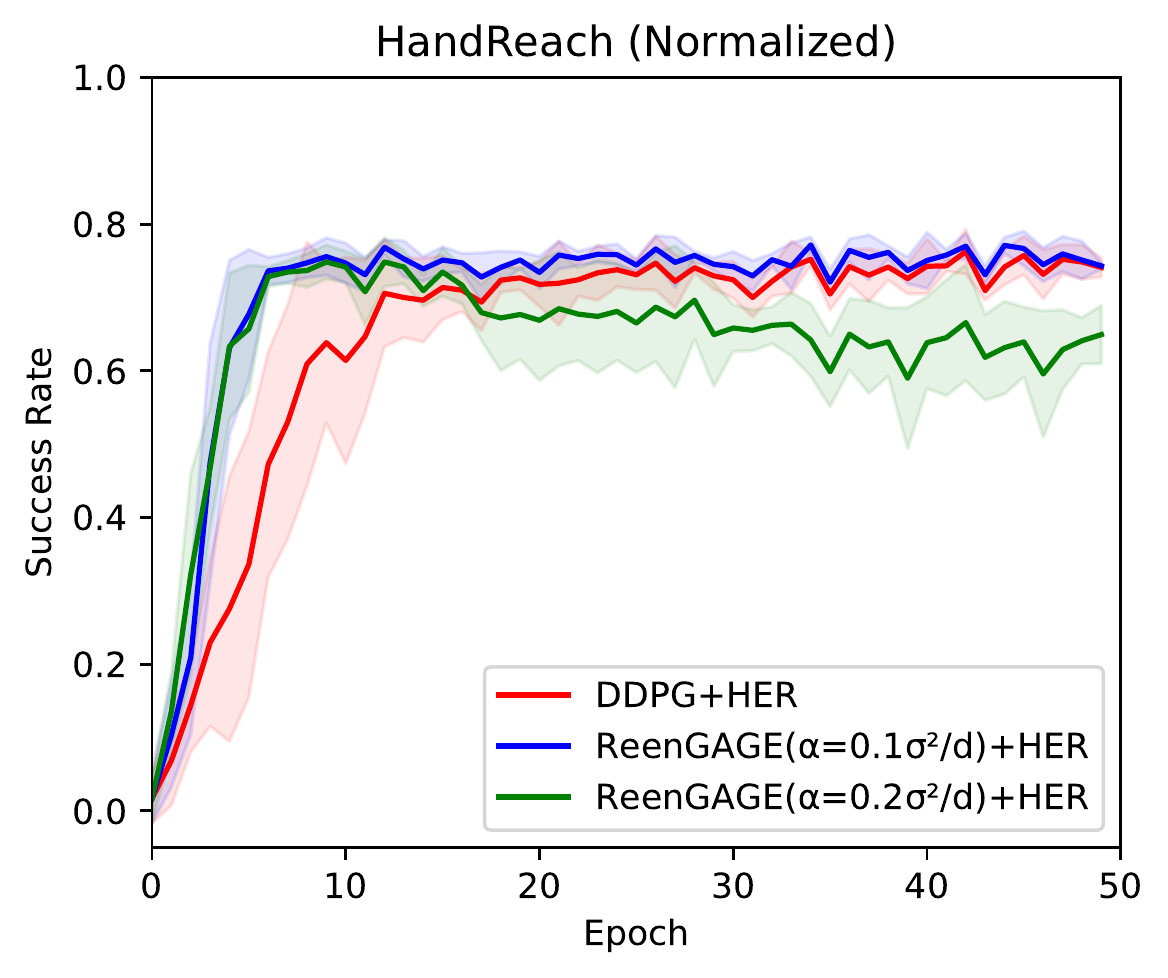}
         \caption{}
     \end{subfigure}
     \hfill
     \begin{subfigure}[b]{0.44\textwidth}
         \centering
         \includegraphics[width=\textwidth]{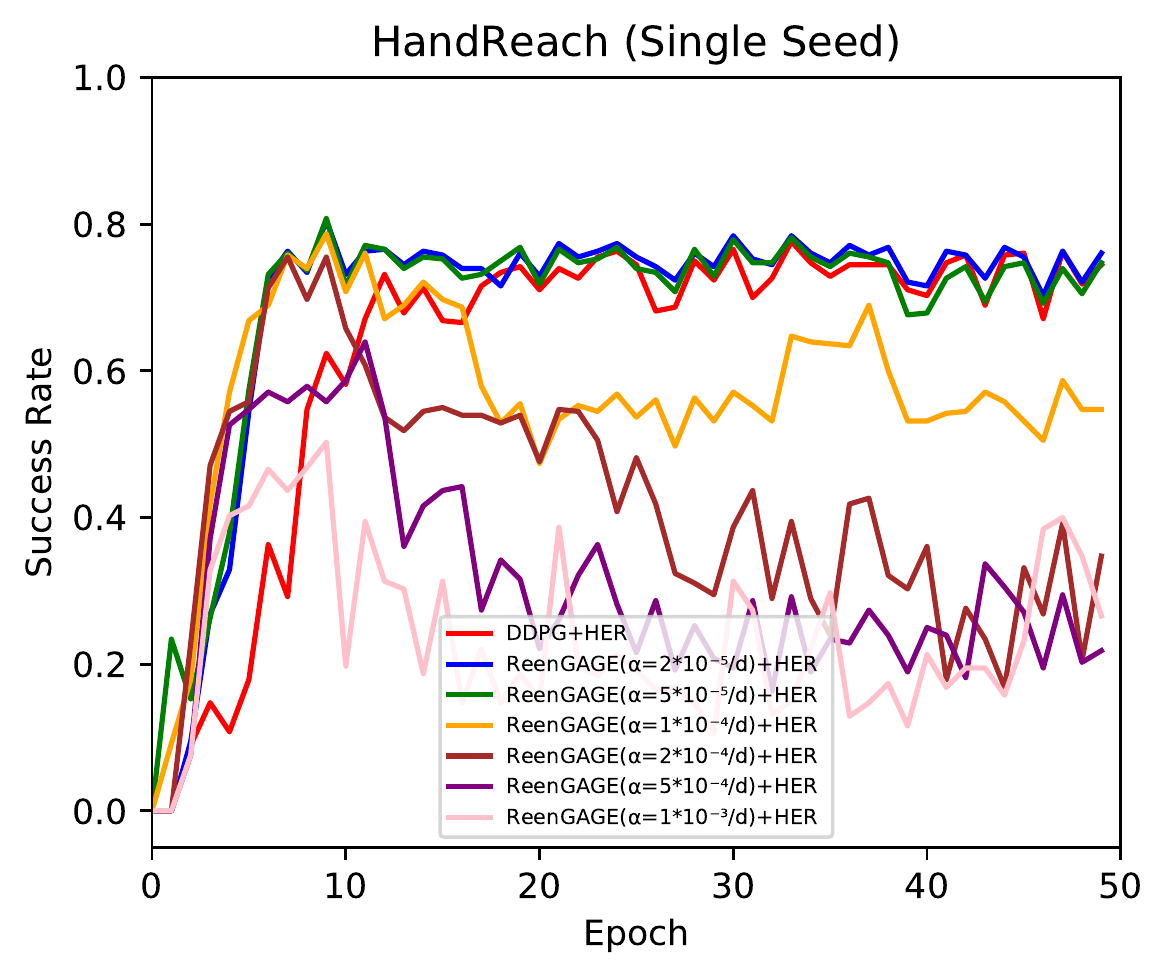}
         \caption{}
     \end{subfigure}
        \caption{Additional Results from Robotics experiments. See text for details.}
        \label{fig:additional_robotics}
\end{figure*}

In Figure \ref{fig:additional_robotics}, we provide additional results for the robotics experiments. In Figure \ref{fig:additional_robotics}-(a) and (b), we give results for the \textbf{HandManipulateBlock} environment: as mentioned in the text, we did not see a significant advantage to using ReenGAGE for this environment. 

In addition to the lower-dimensional goal space of this problem $(d=7$, versus $d=15$ for HandReach), one additional possible reason why ReenGAGE may underperform in this setting is that, unlike in the HandReach case, the dimensions of the goal vector represent diverse quantities of significantly varying scales: some represent angular measurements, while others represent position measurements. 

In order to handle this, we attempted normalizing each dimension of the ReenGAGE loss term. We accomplish this by multiplying the MSE loss for each coordinate $i$ by $\sigma_i^2$, where $\sigma_i$ is the running average standard deviation of dimension $i$ of the goals. (\cite{plappert2018}'s implementation already computes these averages in order to normalize inputs to neural networks.) Intuitively, we do this because the derivative in a given coordinate is in general inversely proportional to the scale of that coordinate (i.e., if $y = 2x$, then $\frac{df}{dy} = 0.5 \frac{df}{dx}$), and because the ReenGAGE MSE loss in each dimension is proportional to the square of the derivative.

However, unfortunately, this did not result in superior performance for HandManipulateBlock: see Figure \ref{fig:additional_robotics}-(c) and (d). As a sanity check, we confirmed that this method performed similarly to the non-normalized ReenGAGE method on HandReach (\ref{fig:additional_robotics}-(e)).

Finally, in Figure \ref{fig:additional_robotics}-(f), we show some additional experiments on HandReach. Specifically, we show results using a single random seed for a wider range of $\alpha$ than shown in the main text: this was performed as a first pass to find the appropriate range of the hyperparameter $\alpha$ to use for the complete experiments. As shown in the main text results, ReenGAGE becomes unstable if too-high values of $\alpha$ are used.

\newpage
\section{ReenGAGE with SAC}
\begin{figure*}[t]
     \centering
     \begin{subfigure}[b]{0.33\textwidth}
         \centering
         \includegraphics[width=\textwidth]{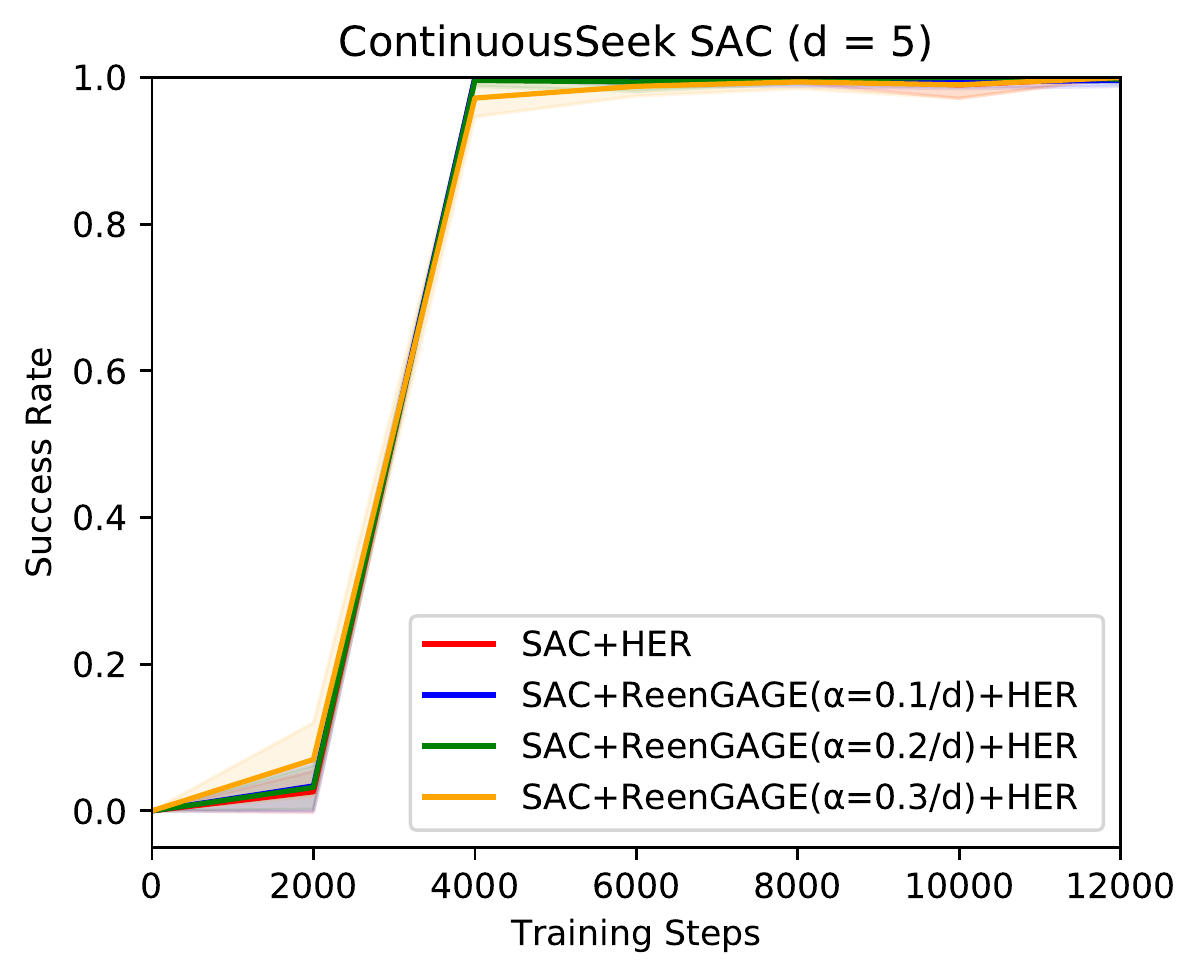}
     \end{subfigure}
     \hfill
     \begin{subfigure}[b]{0.33\textwidth}
         \centering
         \includegraphics[width=\textwidth]{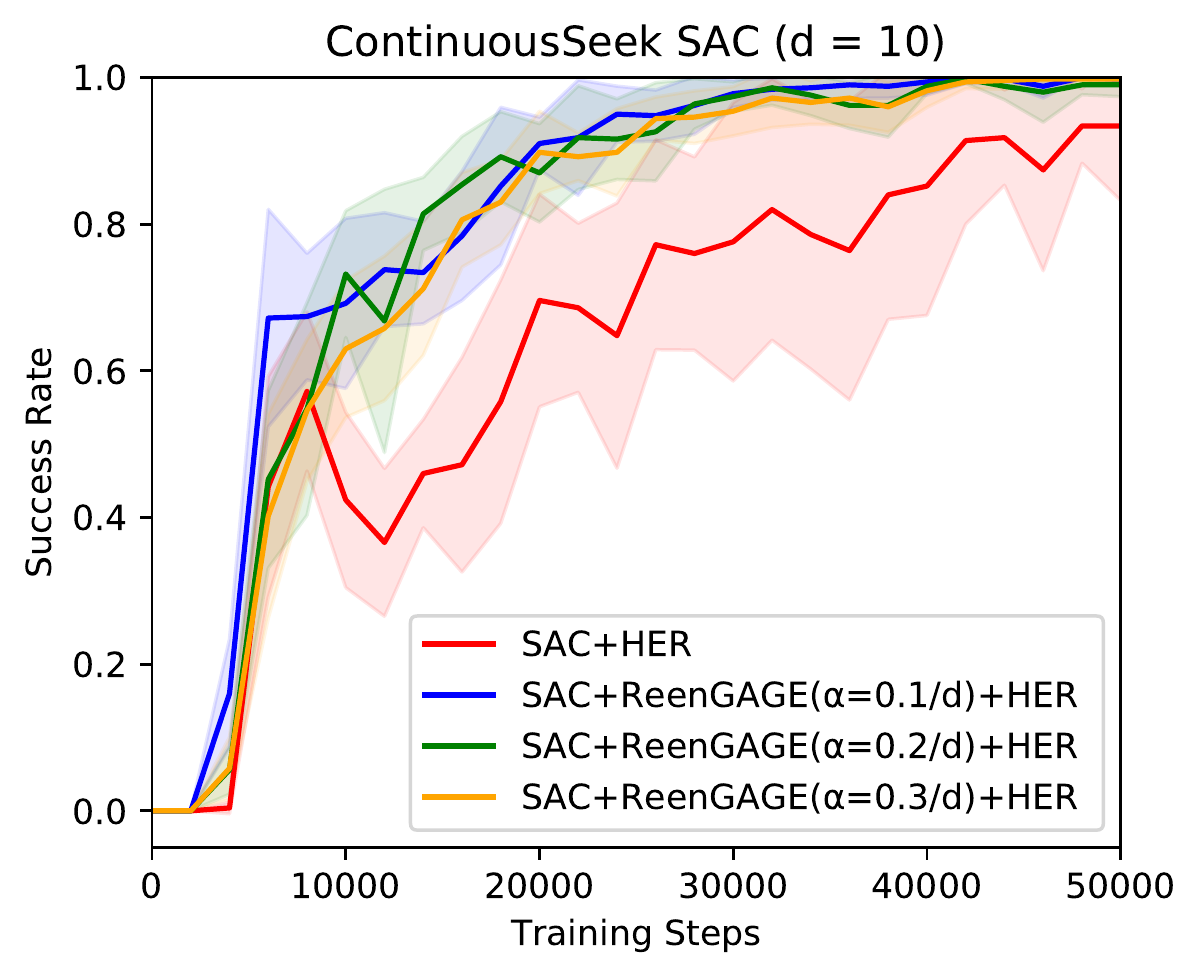}
     \end{subfigure}
     \hfill
     \begin{subfigure}[b]{0.33\textwidth}
         \centering
         \includegraphics[width=\textwidth]{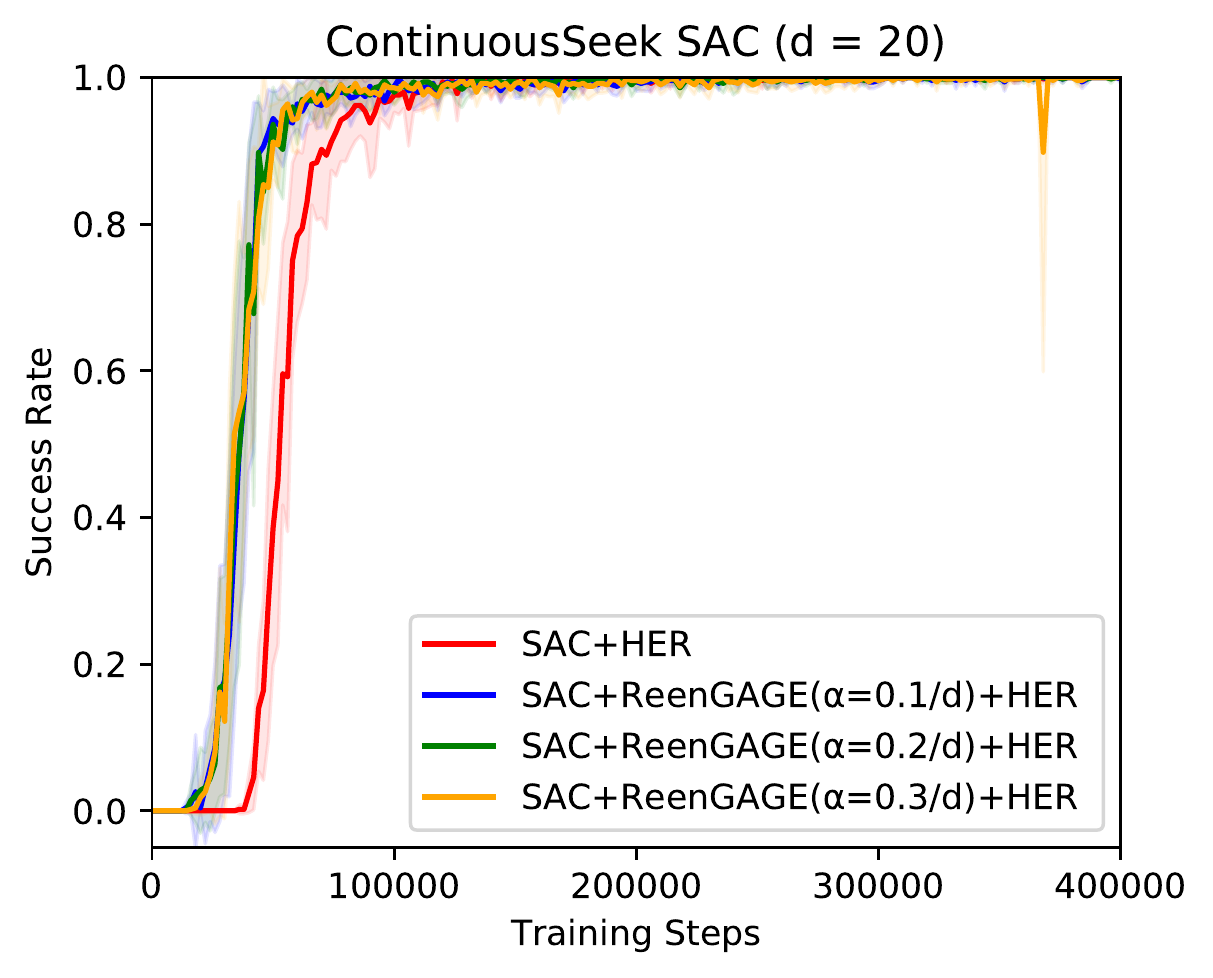}
     \end{subfigure}
        \caption{ContinuousSeek results with SAC. Lines show the mean and standard deviation over 10 random seeds (kept the same for all experiments.) The Y-axis represents the success rate, defined as the fraction of test episodes for which the goal is ever reached.}
        \label{fig:continuous_seek_sac}
\end{figure*}
\begin{figure*}
    \centering
    \includegraphics[width=\textwidth]{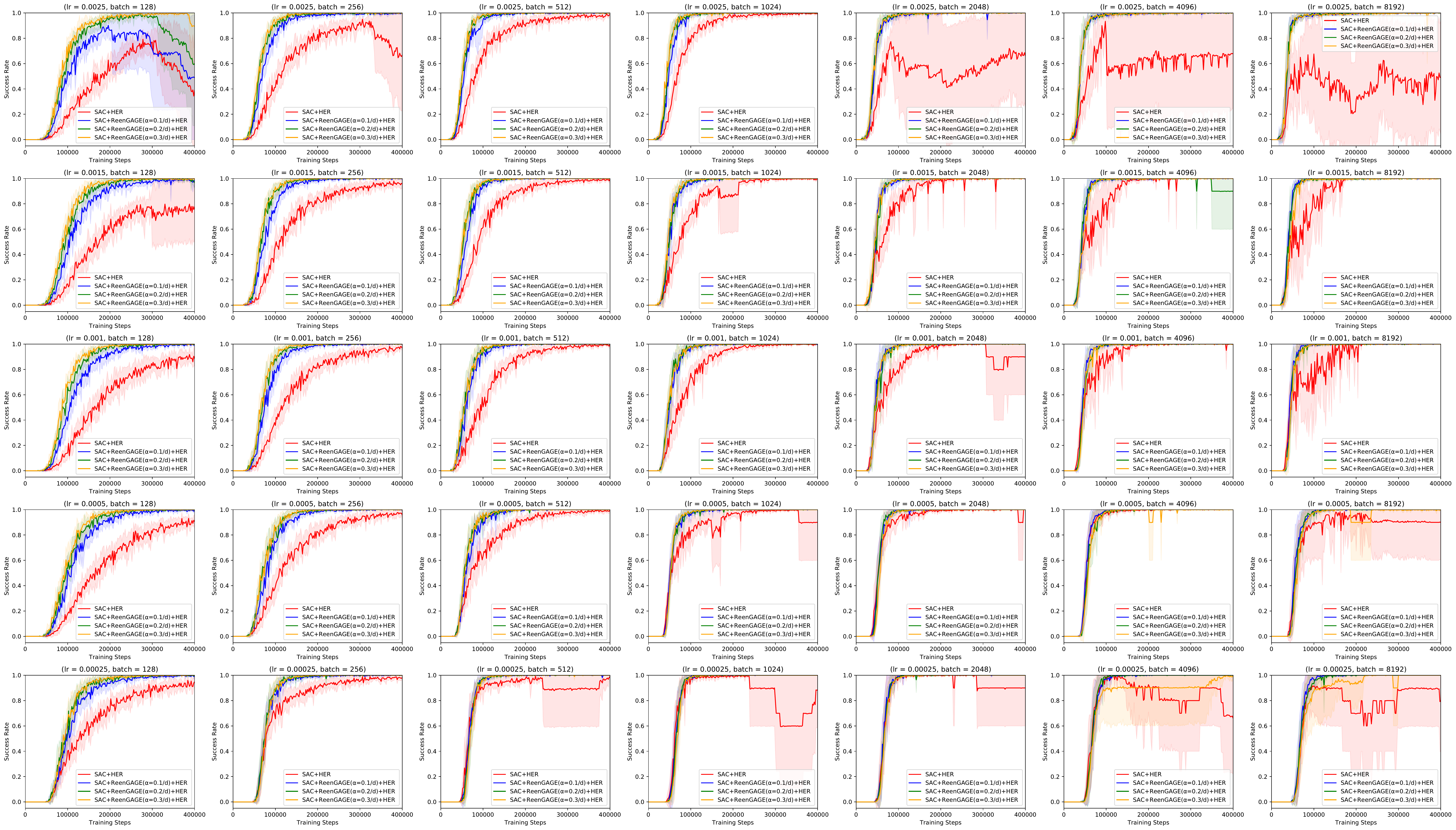}
    \caption{Complete SAC ContinuousSeek results for $d=20$.}
    \label{fig:continuous_seek_sac_all_20}
\end{figure*}
\begin{figure*}
    \centering
    \includegraphics[width=\textwidth]{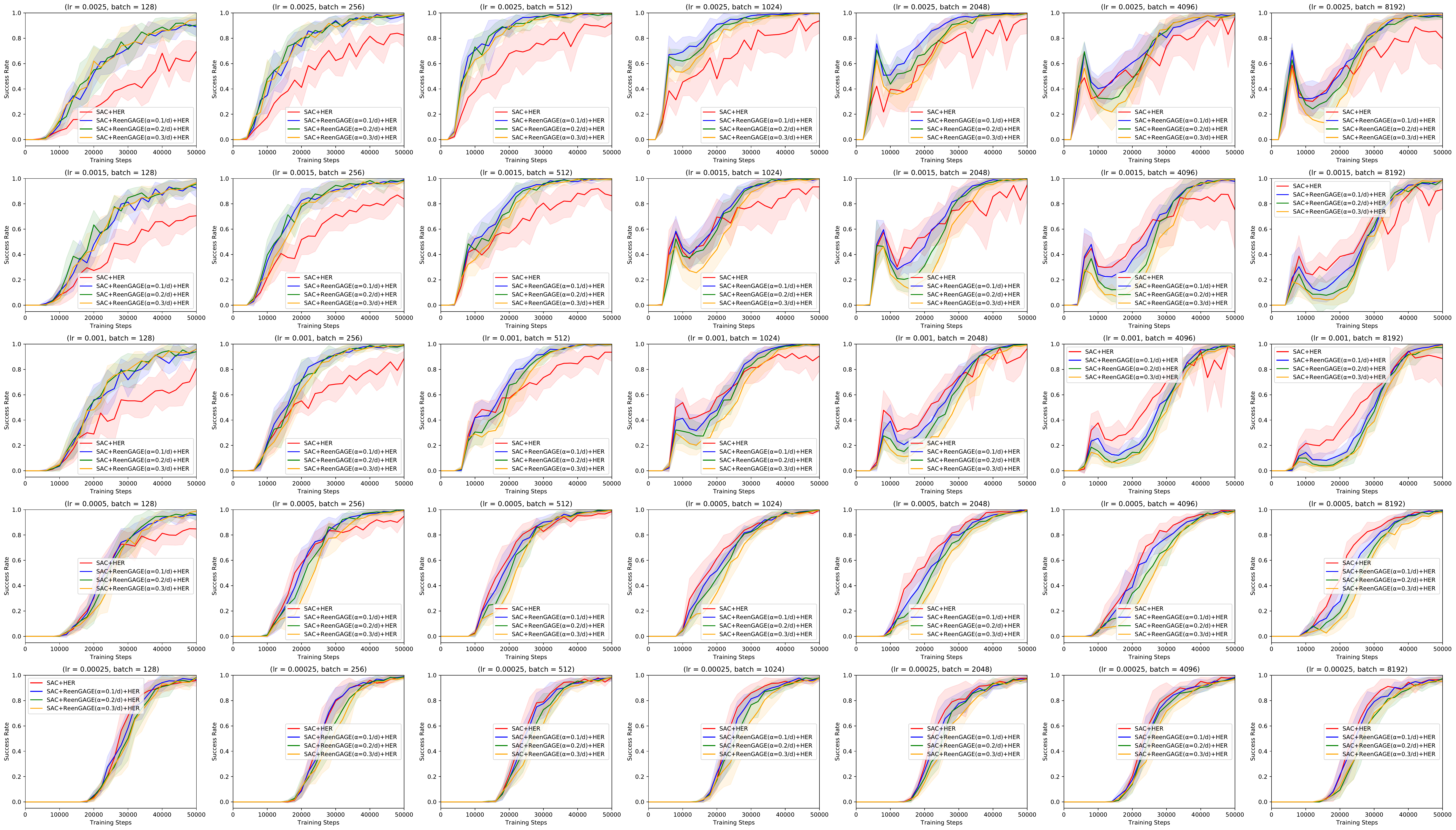}
    \caption{Complete SAC ContinuousSeek results for $d=10$.}
    \label{fig:continuous_seek_sac_all_10}
\end{figure*}
\begin{figure*}
    \centering
    \includegraphics[width=\textwidth]{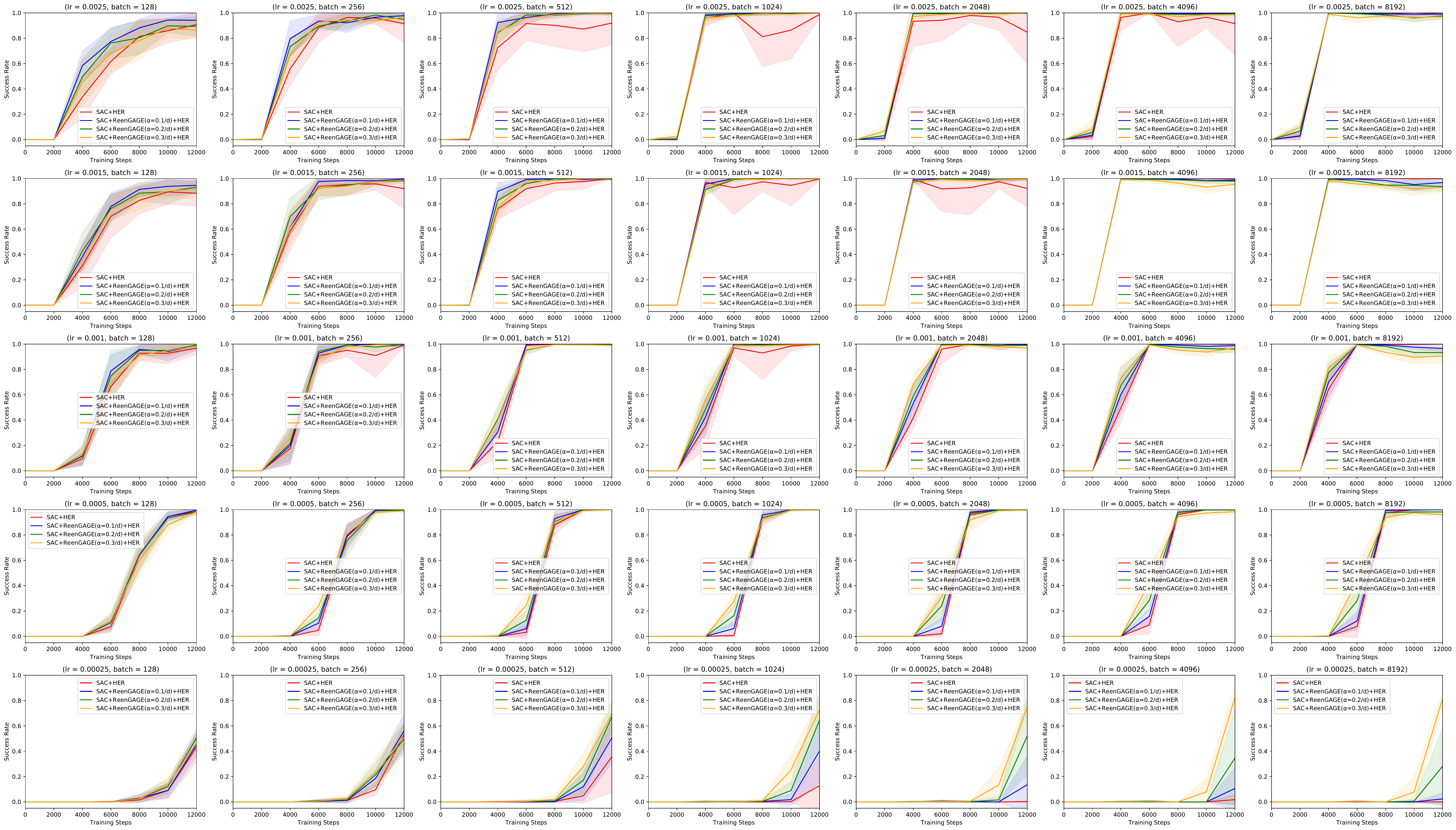}
    \caption{Complete SAC ContinuousSeek results for $d=5$.}
    \label{fig:continuous_seek_sac_all_5}
\end{figure*}
In this section, we explore applying ReenGAGE on top of SAC \cite{haarnoja2018soft}, a variant of DDPG which uses a stochastic policy and rewards for high-entropy polices, uses the current policy to compute targets, and also uses an ensemble critic architecture. The application of ReenGAGE to SAC is straightforward: when computing estimates for $\nabla_g \text{Targ.}(g;s,a)$, we use the well-known ``reparameterization trick'' to differentiate through the stochastic policy (which depends on the goal) and we also differentiate through the entropy reward (which depends on policy's action distribution, and therefore on the goal).

In our experiments, we combine ReenGAGE with SAC and HER, test on our ContinuousSeek environment, and compare to an SAC+HER baseline.  Hyperparameter-optimized ``best'' results for each value of $\alpha$ are shown in Figure \ref{fig:continuous_seek_sac}. As in our DDPG experiments, we performed a grid search over batch size and learning rate hyperparameters. However, using the range of these parameters we used for DDPG led to all best baselines and ReenGAGE models having values of the hyperparameters lying at the ``edges'' and ``corners'' of the hyperparameter grid search space. We therefore increased the search space size, testing all learning rates in $\{0.00025,0.0005,0.001,0.0015,0.0025\}$ and batch sizes in $\{128,256,512,1024,2048,4096,8192\}$. This led to optimal hyperparameters for the baseline in the interior of the search space for the larger scale experiments ($d = 10$ and $d = 20$); see Table \ref{tab:sac_batch_lr}. As with the DDPG ContinuousSeek experiments, the ReenGAGE models still lie on edges/corners of the hyperparameter search space, so it may be possible to get even better ReenGAGE performance by increasing the search space further. Full results are presented in Figures \ref{fig:continuous_seek_sac_all_20}, \ref{fig:continuous_seek_sac_all_10}, and \ref{fig:continuous_seek_sac_all_5}. Note that due to computational limits, we only use 10 random seeds in these experiments, as opposed to 20 for the DDPG experiments. 

Other, non-optimized hyperparameters were fixed, and are generally the same as those for DDPG listed in Table \ref{tab:continuous_seek_hyperparams}. As with DDPG, we use the implementation of SAC from Stable-Baselines3 \cite{JMLR:v22:20-1364} as our baseline; any unlisted hyperparameters are the default from this package.

\begin{table}
    \centering
    \small
    \begin{tabular}{|c|c|c|c|c|c|c|c|c|}
    \hline
         & \multicolumn{2}{|c|}{SAC+HER}   & \multicolumn{2}{|c|}{SAC+ReenGAGE($\alpha$=.1)+HER} & \multicolumn{2}{|c|}{SAC+ReenGAGE($\alpha$=.2)+HER} & \multicolumn{2}{|c|}{SAC+ReenGAGE($\alpha$=.3)+HER}  \\
    \hline
         & Batch& LR & Batch& LR & Batch& LR & Batch& LR\\
    \hline     
        d=5 &8192&0.0025&4096&0.0025&2048&0.0025&2048&0.0025\\
    \hline
        d=10  &1024&0.0015&1024&0.0025&512&0.0025&512&0.0025\\
    \hline
        d=20  &4096&0.0005&8192&0.0025&8192&0.0025&8192&0.0025\\
    \hline
    \end{tabular}
    \caption{``Best'' batch sizes and learning rates for ContinuousSeek for SAC and SAC+ReenGAGE. }
    \label{tab:sac_batch_lr}
\end{table}

\newpage

\section{Multi-ReenGAGE Implementation Details}
\subsection{Batch Implementation}
To efficiently implement the multi-goal Q-function in Equation \ref{eq:multi_q_func} as a batch equation, we use a constant-size representation of the goal set $g=\{g_1,...,g_n\}$. Specifically, we let $g=\{g_1,...,g_{n_{\text{max}}}\}$, where $\{g_{n+1},...,g_{n_{\text{max}}}\}$ are set to a dummy value (practically, $\bm{0}$), and the gate variables $\{b_{n+1},...,b_{n_{\text{max}}}\}$ are all set to zero. Then:

\begin{equation}
Q_\theta(s,a,g) := Q^{\text{head}}_{\theta^\text{h.}}(s,a,\sum_{i= 1}^n[b_i Q_{\theta^\text{e.}}^{\text{encoder}}(s,g_i)]) = Q^{\text{head}}_{\theta^\text{h.}}(s,a,\sum_{i= 1}^{n_{\max}}[b_i Q_{\theta^\text{e.}}^{\text{encoder}}(s,g_i)]).
\end{equation}

However, while this is equal to the intended form of the Q-function without the dummy inputs, the gradient with respect to the full vector $b$ differs from the intended form. Specifically, note that:

\begin{equation}
\frac{\partial Q_\theta(s,a,g)}{\partial b_i} = Q_{\theta^\text{e.}}^{\text{encoder}}(s,g_i) \cdot(\nabla Q^{\text{head}}_{\theta^\text{h.}})(s,a,\sum_{i= 1}^{n_{\max}}[b_i Q_{\theta^\text{e.}}^{\text{encoder}}(s,g_i)])
\end{equation}

In particular, this is nonzero even when $b_i$ is zero, so the gradient will depend on the \textbf{dummy} value $Q_{\theta^\text{e.}}^{\text{encoder}}(s,\bm{0})$. This is clearly not intended. To prevent this, we instead use the form:

\begin{equation}
 Q^{\text{head}}_{\theta^\text{h.}}(s,a,\sum_{i= 1}^{n_{\max}}[b_i^2 Q_{\theta^\text{e.}}^{\text{encoder}}(s,g_i)]).
\end{equation}

(For the target, we construct the policy network similarly using $b_i^2$'s). Note that the above form of the Q-function is equal to the intended form of the Q-function, because $0^2=0$ and $1^2= 1$. However, in this case: 

\begin{equation}
\begin{split}
&\frac{\partial Q_\theta(s,a,g)}{\partial b_i} =\\ &2b_iQ_{\theta^\text{e.}}^{\text{encoder}}(s,g_i) \cdot(\nabla Q^{\text{head}}_{\theta^\text{h.}})(s,a,\sum_{i= 1}^{n_{\max}}[b_i Q_{\theta^\text{e.}}^{\text{encoder}}(s,g_i)]) =\\ &\begin{cases}
2Q_{\theta^\text{e.}}^{\text{encoder}}(s,g_i) \cdot(\nabla Q^{\text{head}}_{\theta^\text{h.}})(s,a,\sum_{i= 1}^{n_{\max}}[b_i Q_{\theta^\text{e.}}^{\text{encoder}}(s,g_i)]) &\text{   if } b_i = 1\\
0 &\text{   if } b_i = 0\\ \end{cases}
\end{split}
\end{equation}

Which is the intended gradient of the Q-function without the dummy inputs, times a constant factor of two.

For the reward, we also use:

\begin{equation}
    R(s',g)   = \sum_{g_i\in g} b_i^2 R_\text{item}(s',g_i) 
\end{equation}
So that the gradient is:
\begin{equation}
    \frac{\partial R}{\partial b_i}  =  b_i R_\text{item}(s',g_i) = \begin{cases} 2R_\text{item}(s',g_i) &\text{   if } b_i = 1\\
    0 &\text{   if } b_i = 0\\ \end{cases}
\end{equation}
Which again the intended gradient of the Q-function without the dummy inputs, times a constant factor of two. Putting everything together, the final gradient loss term is in the form:
\begin{equation}
        \mathcal{L_\textbf{Multi-ReenGAGE}} =  
         \mathcal{L_\textbf{DDPG-Critic}} + \begin{cases} \alpha  \mathcal{L_{\text{mse}}} \Big[2\nabla_{b} Q_\theta(s,a,g), 
         2R_\text{item}(s',g)  + 2\gamma \nabla_b Q_{\theta'}  (s',\pi_{\phi'}(s', g), g)\Big] &\text{   if   }b_i = 1 \\
         \alpha  \mathcal{L_{\text{mse}}} \Big[0,0\Big]  \text{ }(=0) &\text{   if   }b_i = 0 \\
         \end{cases}
\end{equation}
Which is the desired loss, up to a constant factor.
\subsection{Shared Encoder Ablation Study}
\begin{figure*}
     \centering
     \begin{subfigure}[b]{0.44\textwidth}
         \centering
         \includegraphics[width=\textwidth]{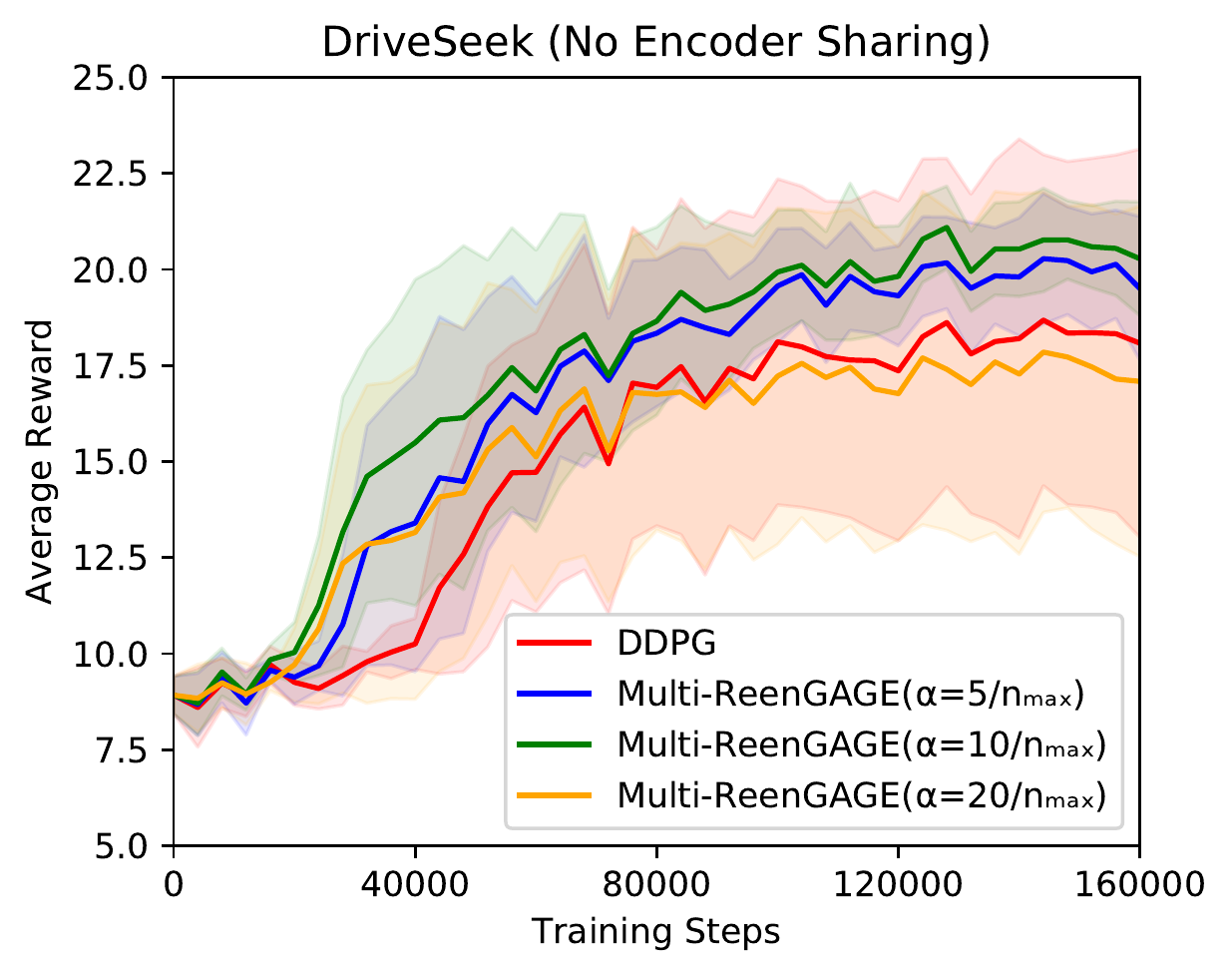}
         \caption{}
     \end{subfigure}
     \hfill
     \begin{subfigure}[b]{0.44\textwidth}
         \centering
         \includegraphics[width=\textwidth]{DriveSeek.pdf}
         \caption{}
     \end{subfigure}
        \caption{Ablation study of encoder sharing for Multi-ReenGAGE on DriveSeek. (a) Without Encoder Sharing; (b)  With Encoder Sharing.}
        \label{fig:multi_reengage_shared_ablation}
\end{figure*}

We performed an ablation study on sharing the encoder between the Q-value function and the policy on DriveSeek when using Multi-ReenGAGE. Results are presented in Figure \ref{fig:multi_reengage_shared_ablation}. We see that sharing the encoder improves the performance of both Multi-ReenGAGE and the DDPG baseline.
\subsection{Training Hyperparameters for Experiments}
The hyperparameters used for the Multi-ReenGAGE experiments are presented in Table \ref{tab:multi_reengage_seek_hyperparams}.
\begin{table}
    \centering
    \begin{tabular}{|c|c|}
    \hline
        Replay Buffer Size &  1000000\\
        Frequency of Training & Every 1 environment step \\
        Gradient Descent Steps per Training & 1 \\
        Initial Steps before Training & 1000\\
        Discount $\gamma$ & 0.95\\
        Batch Size & 256 \\
        Learning Rate & 0.001 \\
        Polyak Update $\tau$ & 0.005 \\
        Normal action noise for training $\sigma$ & 0.05 for DriveSeek; 0.1 for NoisySeek\\
        Embedding Dimension & 20 \\
        Extractor Architecture & Fully Connected; 2 hidden layers of width 400; ReLU activations \\
        Head Architecture (both actor and critic) & Fully Connected; 1 hidden layer of width 400; ReLU activation \\
        Evaluation episodes & 100\\
        Evaluation Frequency & Every 4000 environment steps\\
    \hline
    \end{tabular}
    \caption{Hyperparameters for Multi-ReenGAGE Experiments}
    \label{tab:multi_reengage_seek_hyperparams}
\end{table}
\subsection{Additional Details about Environments}
In this section, we provide additional details about the DriveSeek and NoisySeek environments not included in the main text, in order to more completely describe them.
\subsubsection{DriveSeek}
In the DriveSeek environment, the initial position is always fixed at $(0,0)$, and the initial velocity vector is always at $0$ radians. Episodes last 40 time steps. Goals are sampled by the following procedure: first, a number of goals $n$ is chosen uniformly at random from $\{1,...,200\}$; then $n$ goals are chosen uniformly without replacement from the integer coordinates in $[-10,10]^2$.
In addition to the goals, the observation $s$ that the agent and policy receives is 6 dimensional: it consists of the current position $s_\text{pos.}$, the \textit{rounded} version of $s_\text{pos.}$ (this is the ``achieved goal'': a reward is obtained if this matches one of the input goals), and the sine and cosine of the velocity vector.
\subsubsection{NoisySeek}
In the NoisySeek Environment, the initial position is fixed a $(0,0)$. Episodes last 40 time steps. Goals are sampled by the following procedure: first, a number of clusters $n_c$ is chosen from a geometric distribution with parameter $p=0.15$, and a maximum number of goals $n'$ is chosen uniformly at random from $\{1,...,200\}$. Then, cluster centers are chosen from a Gaussian distribution with mean 0 and standard deviation 10 in both dimensions. Next, a Dirichlet distribution of order $K=n_c$, with $\alpha_1,...,\alpha_K = 1$, is used to assign a probability $p_j$ to each cluster. Next, each of the $n'$ goals are assigned to a cluster, with probability $p_j$  of being assigned to cluster $j$. Then, each goal is determined by adding Gaussian noise with mean 0 and standard deviation 2  in both dimensions to the cluster center assigned to that goal, and then rounding to the nearest integer coordinates. Finally, goals are de-duplicated.

In addition to the goals, the observation $s$ that the agent and policy receives is 4 dimensional: it consists of the current position $s$, and the \textit{rounded} version of $s$ (this is the ``achieved goal'': a reward is obtained if this matches one of the input goals).
\subsection{NoisySeek results for additional $\alpha$ values}
\begin{figure}
    \centering
    \includegraphics[width=0.5\textwidth]{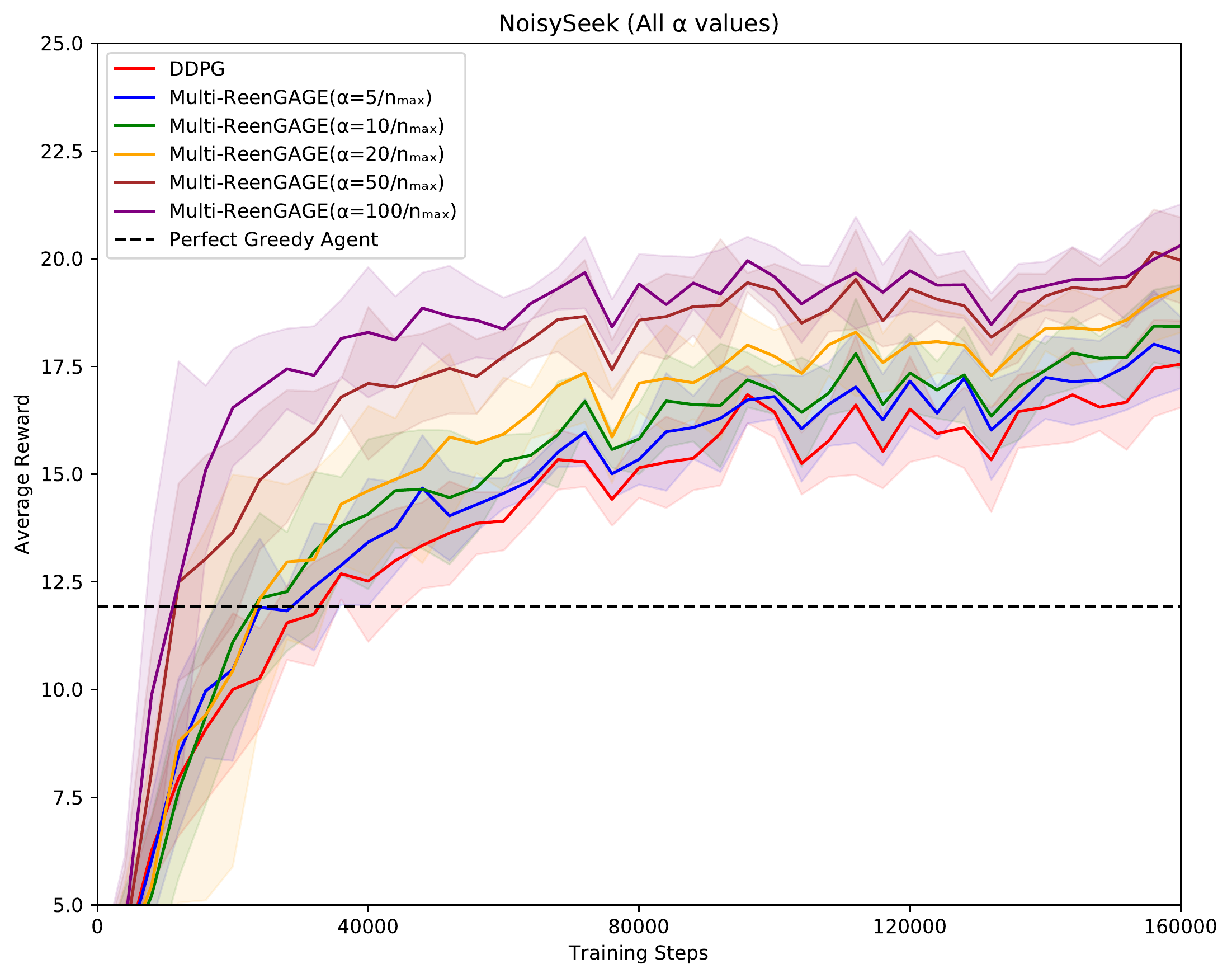}
    \caption{Results for NoisySeek Environment including additional values of $\alpha$.}
    \label{fig:noisyseek_alll}
\end{figure}

We tested NoisySeek  with additional values of $\alpha$, which we did not include in the main text to avoid cluttered presentation; the trend is generally the same as shown in the main text. Results are shown in Figure \ref{fig:noisyseek_alll}.
\subsection{DriveSeek with CNN Architecture}
\begin{figure}
   \centering 
   \includegraphics[width=0.5\textwidth]{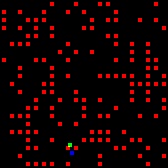} 
    \caption{Example input to CNN architecture for DriveSeek. We use one channel (red) to represent the goals (note that these pixels directly correspond to the differentiable indicator variables $b_i^2$ for each possible goal), another channel (green) to represent the current position, and a third channel (blue) to represent the next position if the action $a$ is zero: in other words, it indicates $\mathbf{s}_\text{vel.}$. Goals are spaced out so that (approximately, up to single-pixel rounding) a goal is achieved if the current position indicator (green) overlaps with the (red) goal. The exact position and velocity vectors are also provided as a separate input, apart from the CNN.}
    \label{fig:driveseek_cnn_frame}
\end{figure}
\begin{figure}
    \centering
    \includegraphics[width=0.5\textwidth]{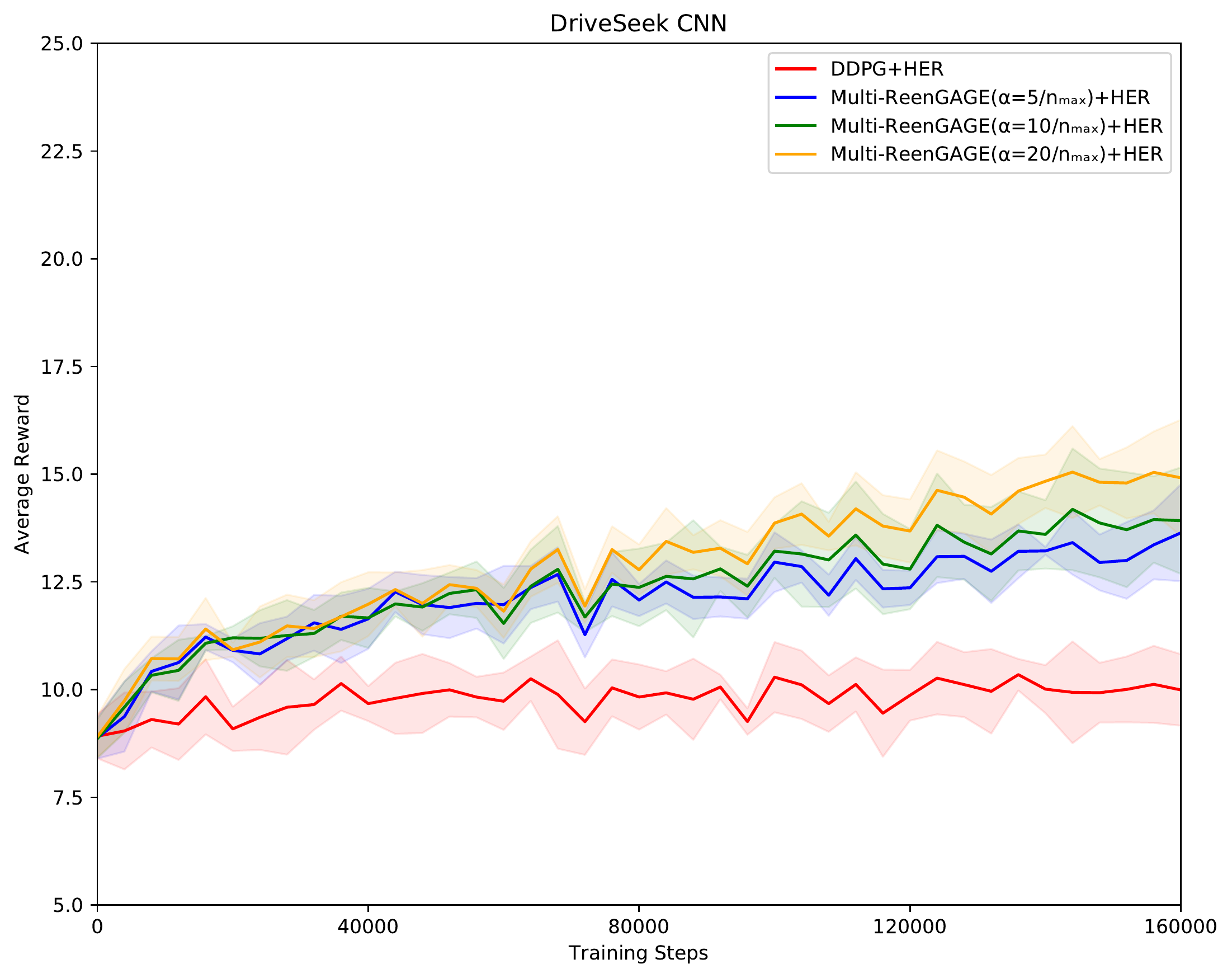}
    \caption{Results for DriveSeek with CNN architecture.}
    \label{fig:driveseek_cnn_main_results}
\end{figure}
Because the positions in the DriveSeek environment are bounded, we attempted using a CNN architecture to interpret the state and goals. In particular, because all possible goals appear at unique locations in two-dimensional space, rather than using a DeepSets \cite{NIPS2017_f22e4747}--style architecture, we can directly surface the goal ``gates" $b_i$ as part of the input image: the location in the image corresponding to $g_i$ is blank if $b_i$ is zero (the goal is absent) and colored in if $b_i$ is one (the goal is present).  See Figure \ref{fig:driveseek_cnn_frame}. Note that, as in the DeepSet implementation, we use $b_i^2$ in the representation, so that goals which are absent have zero associated attention. We used the standard ``Nature CNN'' architecture from \cite{mnih2015human} with otherwise the same training hyperparameters as used in the main-text experiment. The CNN was shared between the actor and the critic networks, and trained only using the critic loss; its output was then fed into separate actor and critic heads, consisting of a single hidden layer of width 400, as in the DeepSets-based architecture. Results are shown in Figure \ref{fig:driveseek_cnn_main_results}.

\begin{figure*}
     \centering
     \begin{subfigure}[b]{0.49\textwidth}
         \centering
         \includegraphics[width=\textwidth]{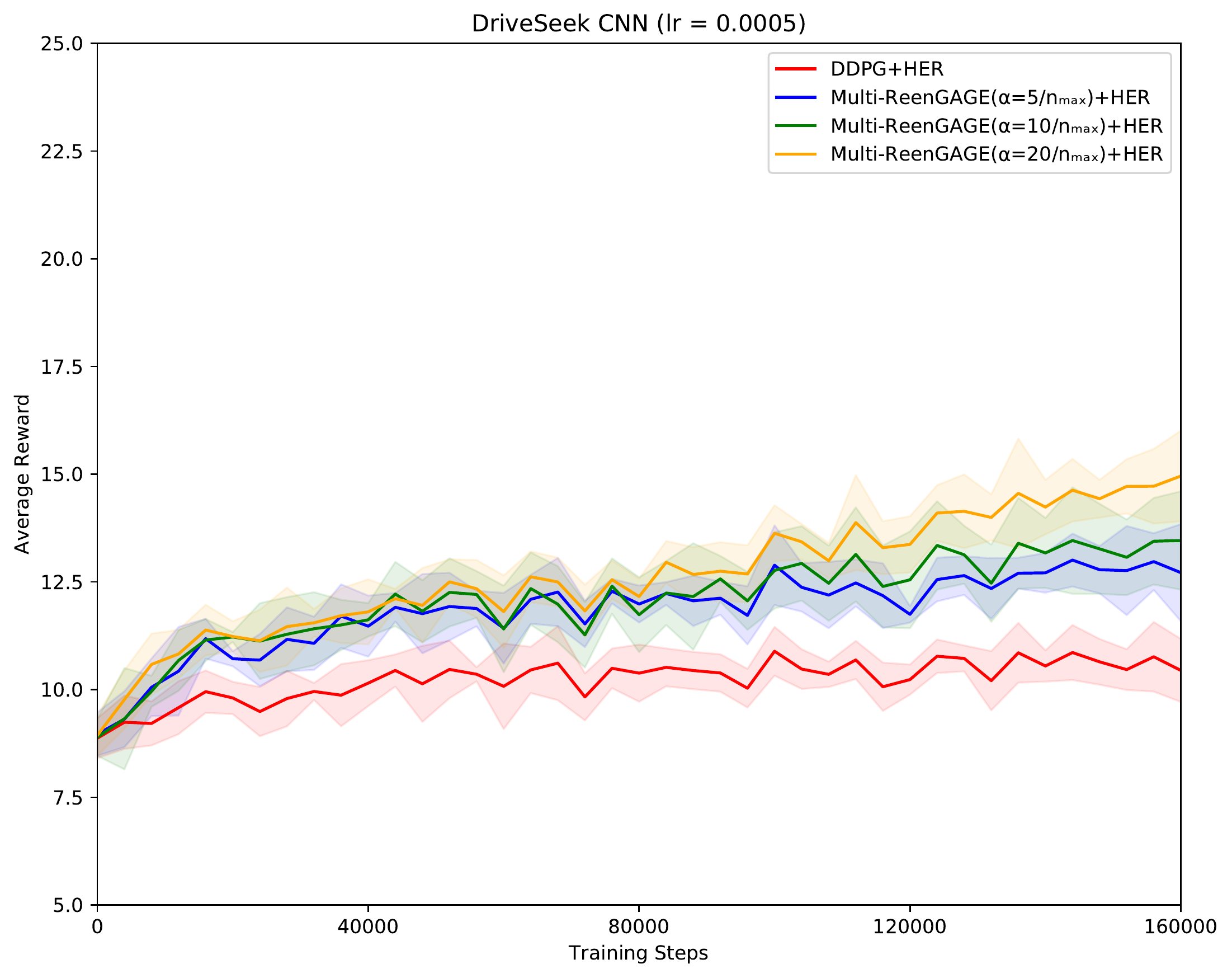}
         \caption{}
     \end{subfigure}
     \hfill
     \begin{subfigure}[b]{0.49\textwidth}
         \centering
         \includegraphics[width=\textwidth]{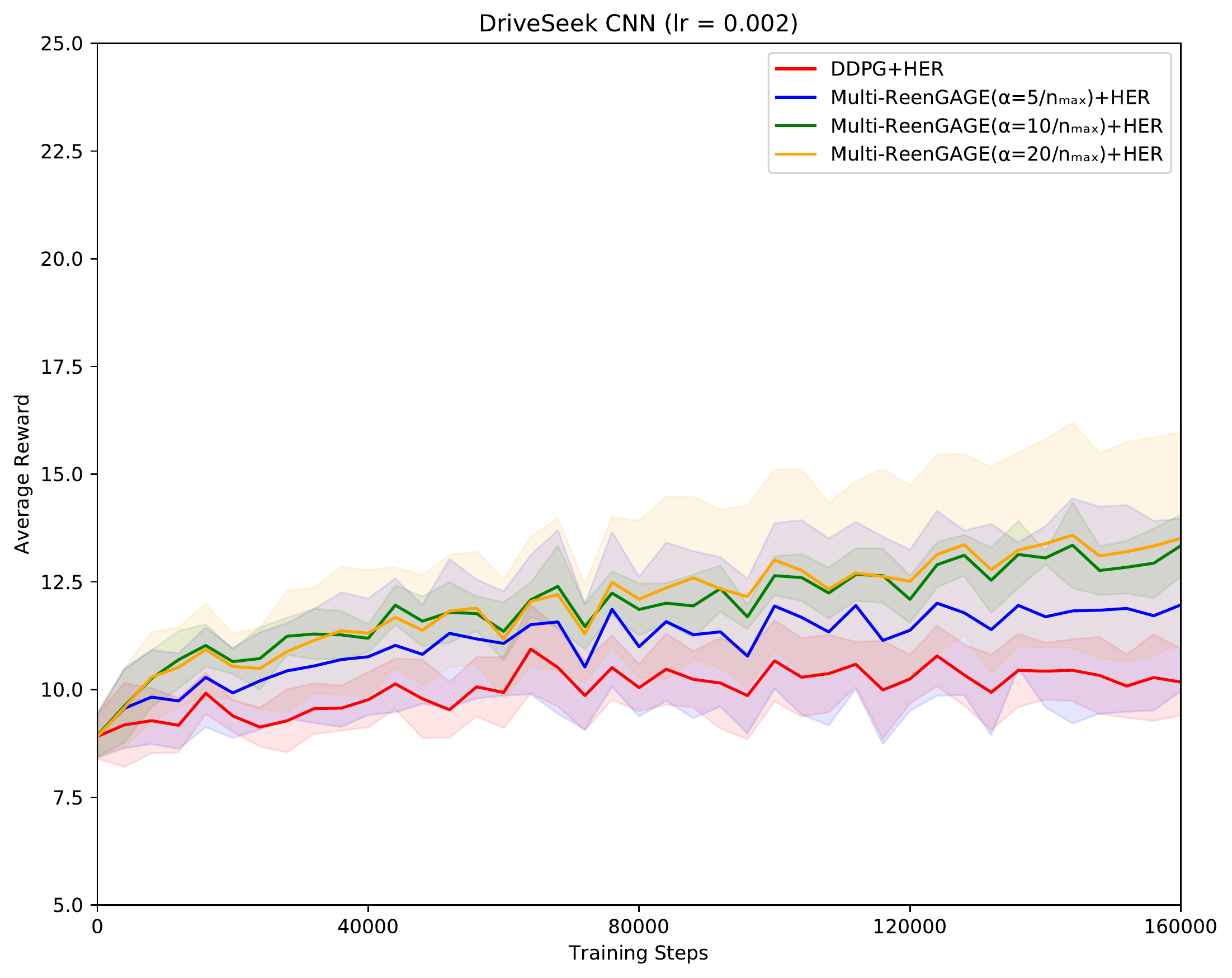}
         \caption{}
     \end{subfigure}
        \caption{DriveSeek with CNN architecture with varied values of the learning rate.}
        \label{fig:additional_driveseek_cnn}
\end{figure*}

In general, the performance was worse than using the DeepSets-style architecture; however, the models using Multi-ReenGAGE still outperform the standard DDPG models. One possible explanation for this is that the hyperparameters are poorly-tuned for the CNN architecture. Doubling and halving the learning rate (to 0.002 and 0.0005, respectively) did not seem to affect the results much (Figure \ref{fig:additional_driveseek_cnn}), but it is possible that other hyperparameter adjustment may lead to better performance. Another possible explanation for the poor performance is that the pixel-resolution of the images (each pixel has width of 0.125 units) was not sufficient to capture the real-valued dynamics of the environment (although we did also include the real-valued state position and velocity vectors [and rounded state position] as inputs concatenated to the CNN output).

\section{Runtime Discussion and Empirical Comparisons}
As noted in \cite{Etmann2019ACL}, loss terms involving gradients with respect to inputs, such as the ReenGAGE loss, should only scale the computational cost of computing the parameter gradient update by a constant factor.

For an intuition as to why this is the case, recall the well-known fact that for a differentiable scalar-output function $f(\mathbf{x}, \theta)$ represented by some computational graph which requires $n$ operations to compute, standard backpropagation can compute the gradient $\nabla_\mathbf{x} f(\mathbf{x}, \theta)$ (or the gradient  $\nabla_\theta f(\mathbf{x}, \theta)$) using only $cn$ operations, for a constant $c$. However, now note that $\nabla_x f(\mathbf{x},\theta)$ can \textit{itself} be though of as a $cn$-operation component of a larger computational graph. Thus, if we consider the scalar-valued function $h( \nabla_\mathbf{x}  f(\mathbf{x},\theta))$, where $h$ itself takes $k$ additional operations to compute, then we can upper-bound the total number of operations required to compute the gradient $\nabla_\theta h( \nabla_\mathbf{x} f(\mathbf{x},\theta))$ by  $c(cn+k) = c^2n + ck$; in other words, a constant factor $c^2$ of $n$, plus some overhead. In the particular case of the ReenGAGE loss term (in the sparse case, for simplicity), $x$ corresponds to the goal $g$ and $f$ corresponds to $Q_\theta(s,a,g)$, while $h$ corresponds to $\|\nabla_g f(g,\theta) - \nabla_g \gamma Q_{\theta'} (s',\pi_{\phi'}(s', g), g) \|_2^2$.  If the $Q$ function requires $n$ operations to evaluate and the policy $\pi$ requires $m$ operations, then the overall computational cost can therefore be upper-bounded by $c(cn + c(n+m) +k) = 2c^2n + c^2m + ck$, where $k$ is the (trivial) amount of computation required to compute the norm. This is therefore a constant factor of the time needed to compute the value of the Q function and its target (with some trivial overhead due to the norm computation).

 \cite{Etmann2019ACL} provides explicit algorithms for computations of gradients of forms similar to the form $\nabla_\theta h( \nabla_\mathbf{x}  f(\mathbf{x},\theta))$ discussed above and confirms the constant-factor increase in computational complexity; in particular, $h(\cdot)$ corresponds to $p(\cdot)$ in Equation 10 of \cite{Etmann2019ACL}. (Note that \cite{Etmann2019ACL}'s analysis is somewhat more general, allowing for a vector-valued $f$, with $p(\cdot)$ a function of a Jacobian-vector product of $f$ rather than simply the gradient).

To provide empirical support for this, we provide runtime comparisons for training with and without the ReenGAGE loss term, for the experiments in the main text. Note that we are comparing total runtimes, so these times include the environment simulation; however this should be relatively minor for all experiments (because the environments themselves are relatively simple) except possibly the robotics experiments. 

For ContinuousSeek and Multi-ReenGAGE experiments, all tests were run on a single GPU. We used a pool of shared computational resource, so the GPU models may have varied between runs; GPUs possibly used were NVIDIA RTX A4000, RTX A5000, and RTX A6000 models. (This adds some uncertainty to our runtime comparisons.) Robotics experiments were run on 20 CPUs each, as described by \cite{plappert2018}.

Runtime comparison results are given in Table \ref{tab:runtime_comparisons}. For ContinuousSeek, we consider only experiments with $d=20$, batch size $= 256$; for others, we include all experiments shown in the main text. For runtimes with ReenGAGE, we average over all values of $\alpha$ included in the main text. In general, runtime increases ranged from 34\%-60\%.

\begin{table}
    \centering
    \begin{tabular}{|c|c|c|c|}
    \hline
    Environment &  Runtime without ReenGAGE (s) & Runtime with ReenGAGE (s)& Mean Percent Increase\\
    \hline
        ContinuousSeek &2549 $\pm$ 203&3591 $\pm$ 248& 40.9\% \\
        HandReach &2533 $\pm$ 2 &3395 $\pm$ 208 & 34.0\%  \\
        DriveSeek &8931 $\pm$2037 &12467 $\pm$ 1354& 39.6\% \\
        NoisySeek &7385 $\pm$ 503&11814 $\pm$ 1519&  60.0\% \\
    \hline
    \end{tabular}
    \caption{Effect of ReenGAGE on runtimes. Error values shown are standard deviations over all runs.}
    \label{tab:runtime_comparisons}
\end{table}
\end{document}